\documentclass[12pt,a4paper,oneside]{report}
\usepackage[round]{natbib}
\usepackage{multirow}
\usepackage{setspace}
\usepackage{amsmath}
\usepackage{amssymb}
\usepackage{subfigure}
\usepackage{graphicx}
\usepackage[raggedright]{titlesec}
\usepackage{rotating}
\usepackage{xcolor}
\usepackage[hidelinks]{hyperref}

\begin{document}

\pagenumbering{gobble}











\begin{center}
\large \textbf{\uppercase{Deep Neural Networks for \\Relation Extraction}}

\vspace{25 mm}

\large \textbf{\uppercase{Tapas Nayak}} \\
\large \textit{(B.E. \& M.E., Jadavpur University, India)}

\vspace{25 mm}

\large \textbf{\uppercase{A THESIS SUBMITTED FOR THE DEGREE OF}} \\
\vspace{2 mm}
\textbf{\uppercase{DOCTOR OF PHILOSOPHY}} \\
\vspace{8 mm}
\textbf{\uppercase{ Department of Computer Science\\ School of Computing}} \\
\vspace{2 mm}
\textbf{\uppercase{NATIONAL UNIVERSITY OF SINGAPORE}}

\vspace{15 mm}

\textbf{2020}

\vspace{20 mm}

\textbf{Supervisor:}\\
Professor Ng Hwee Tou

\vspace{6 mm}

\textbf{Examiners:}\\
Associate Professor Kan Min Yen \\
Associate Professor Ng Teck Khim \\
Assistant Professor Huang Ruihong (Texas A\&M University)

\end{center}

\clearpage


\doublespacing

\pagenumbering{roman}

\noindent \huge \textbf{Declaration}
\vspace{18 mm}
\normalsize

I hereby declare that this thesis is my original work and it has been written by me in its entirety. I have duly acknowledged all the sources of information that have been used in this thesis.

This thesis has also not been submitted for any degree in any university previously.

\vspace{24 mm}

\begin{center}
\singlespacing
\includegraphics[scale=0.7]{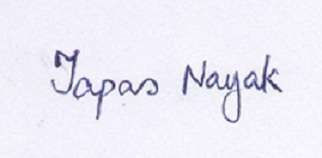}\\
\rule{60 mm}{1pt}\\
\doublespacing
Tapas Nayak\\
\today
\end{center}

\clearpage


\doublespacing

\vspace{22 mm}

\noindent \huge \textbf{Acknowledgements}

\vspace{25mm}

\normalsize

I would like to express my deepest gratitude to my advisor, Professor Ng Hwee Tou, for his valuable support. His knowledge and critical analysis have helped me to shape my research. Discussions with him have helped me a lot to improve my research capabilities. I have enjoyed working in a research environment that gives me ample opportunities for original thinking and initiatives. 

Also, I would like to thank the board of examiners, Associate Professor Kan Min Yen, Associate Professor Ng Teck Khin, and Assistant Professor Huang Ruihong for their valuable comments and insights.

I want to thank my friends and colleagues in the NUS NLP group for their help and support. It was a great pleasure to work with them.

Finally, I want to dedicate this thesis to my parents and my sisters for their support and encouragement. 

\clearpage
     

\tableofcontents

\chapter*{Summary}
\label{chapt:summary}

A knowledge base (KB) is a useful resource for many natural language processing tasks. KBs contain real-world entities and relations among them which can help downstream tasks such as question answering. A triplet of two entities and a relation between them is called a relation tuple. Existing KBs such as Freebase, Wikidata, and DBpedia contain a large number of relation tuples. But these KBs are built by crowd workers and it takes much time and effort to build them. The automatic extraction of relation tuples from natural language texts is referred to as relation extraction. In this thesis, we tackle this task using novel deep neural network models.

First, we use a pipeline approach for this task, where we assume that the entities have already been identified by an external named entity recognition system. We propose a syntax-focused multi-factor attention model to find the relation between two entities. We use the syntactic distance of words from the entities to determine their importance in establishing the relation between the two given entities. We also use multi-factor attention to focus on multiple pieces of evidence present in a text to support the relation. Our proposed model achieves significant improvements over prior works on widely used relation extraction datasets.

Second, we tackle the task of joint entity and relation extraction, where entities are not identified beforehand. There may be multiple relation tuples present in a sentence, and these relations may share one or both entities among them. Extracting such relation tuples with full entity names from sentences is a difficult task. We propose two approaches to use encoder-decoder networks for joint extraction of entities and relations. In the first approach, we propose a representation scheme for relation tuples that enables the decoder to generate one token at a time (like machine translation models) and still extract all the tuples present in a sentence, with full entity names of different lengths and with overlapping entities. Next, we propose a pointer network-based decoding approach where an entire tuple is generated at every time step. Our proposed models outperform prior works on widely used relation extraction datasets.

Finally, we extend our work to multi-hop relation extraction. Distantly supervised relation extraction models mostly focus on sentence-level relation extraction, where the two entities (subject and object entity) of a relation tuple must appear in the same sentence. This assumption is overly strict and for a large number of relations, we may not find sentences containing the two entities. To solve this problem, we propose multi-hop relation extraction, where the two entities of a relation tuple may appear in two different documents but these documents are connected via some common entities. We can find a chain of entities from the subject entity to the object entity via the common entities. The relation between the two entities can be established using this entity chain. Following this multi-hop approach, we create a dataset for 2-hop relation extraction, where each chain contains exactly two documents. This dataset covers a higher number of relations than previous sentence-level or document-level extraction datasets that are available in the public domain. To solve this task, we propose a hierarchical entity graph convolutional network (HEGCN) model that consists of a two-level hierarchy of graph convolutional networks (GCNs). The first-level GCN of the hierarchy captures the relations among the entity mentions within the documents, and the second-level GCN of the hierarchy captures the relations among the entity mentions across the documents in a chain. Our proposed HEGCN model improves the performance on our 2-hop relation extraction dataset and it can be readily extended to N-hop datasets. 

\listoftables
\listoffigures

%
\clearpage
\pagenumbering{arabic}
\setcounter{page}{1}

%
\chapter{Introduction}
\label{chapt:intro}

The Web is a huge source of unstructured texts. Humans can read and extract important information from the Web. However, a machine cannot handle this unstructured data very well. To extract important information from unstructured text automatically, we need to convert this unstructured data to some structured form which machines can understand easily. This is one of the major goals of information extraction in computer science.

Information extraction (IE) research goes back to the late 1970s. JASPER was the very first information extraction system built by Carnegie Group\footnote{\url{https://en.wikipedia.org/wiki/Carnegie_Group}} in the mid-1980s. In the beginning of 1987, IE research was spurred by a series of Message Understanding Conferences\footnote{\url{https://en.wikipedia.org/wiki/Message_Understanding_Conference}} and most of the funding for this research came from the U.S. Defense Advanced Research Projects Agency (DARPA\footnote{\url{https://en.wikipedia.org/wiki/DARPA}}). 

In recent years, with the growing amount of unstructured texts on the Web, the field of information extraction has attracted much attention. Currently, the Web is considered as a collection of documents. Users have to extract relevant information from these documents on their own. Success in IE research can convert unstructured documents to structured data and an automated system can give users all relevant information rather than just some relevant documents.

An information extraction system is an automated system that takes a sentence (mainly in the English language, but can be extended to other languages too) and extracts important information and presents it in a machine-readable structured format. Most IE systems extract binary relation tuples containing two entities and a relation between them. Some IE systems keep additional information like time, direction, and context too. They are called n-ary tuples. Table \ref{tab:binary_nary} gives an example of a binary and an n-ary tuple.

\begin{table}[ht]
\centering
\begin{tabular}{|l|l|l|l|}
\hline
Tuple       & Sentence               & Tuple     & Sentence             \\ \hline
arg1        & Eli Whitney            & arg1      & Eli Whitney          \\ 
rel         & created                & rel       & created    \\ 
arg2        & the cotton gin         & arg2      & the cotton gin       \\ 
            &                        & arg3      & in 1793                 \\ \hline
\multicolumn{2}{|l|}{binary relation}   & \multicolumn{2}{l|}{n-ary relation} \\ \hline
\end{tabular}
\caption{Binary vs n-ary relation tuple.}
\label{tab:binary_nary}
\end{table}

\begin{figure}[ht]
\centering
\includegraphics[scale=0.7]{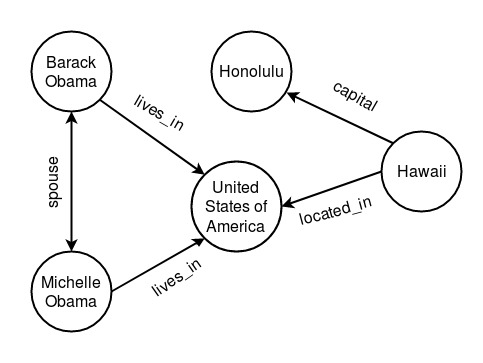}
\caption{An example of a tiny knowledge base.}
\label{fig:kb}
\end{figure}

A knowledge base (KB) is a good example of a large database that stores binary relation tuples about real world entities. Freebase \citep{bollacker2008freebase}, Wikidata \citep{wikidata}, and DBpedia \citep{bizer2009dbpedia} are examples of large KBs. Figure \ref{fig:kb} gives some idea about the structure of a KB. This example KB has five entities: {\em Barack Obama} and {\em Michele Obama} of PERSON type, and {\em United States of America}, {\em Hawaii}, and {\em Honolulu} of LOCATION type. In addition, there exist a few relations among them, such as {\em spouse}, {\em lives\_in}, {\em located\_in}, and {\em capital}. 

\begin{figure}[ht]
\centering
\includegraphics[scale=0.7]{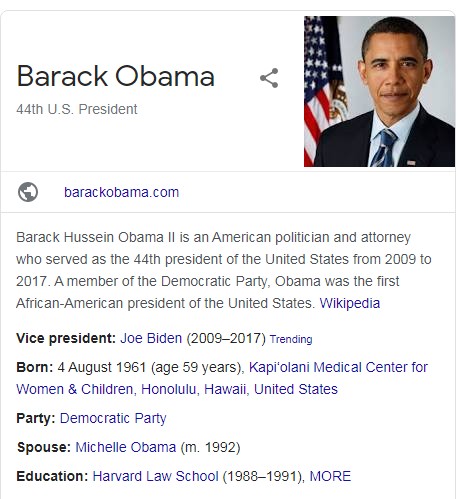}
\caption{The use of knowledge base in search engines. Source: \url{https://www.google.co.in/search?q=Barack+Obama}. Taken on 30 Oct 2020.}
\label{fig:se}
\end{figure}

These structured KBs are very useful for many downstream natural language understanding tasks such as question answering. Factoid question answering is an important task in natural language processing. With the help of KBs, factoid questions can be answered easily as shown in Table \ref{tab:kb_qa}. Many search engines use knowledge bases to populate important information automatically as infoboxes in search results as shown in Figure \ref{fig:se}. The user can obtain information about the entity from its infobox without going through the text. This will save much time for the users. 

\begin{table}[ht]
\centering
\begin{tabular}{|l|c|}
\hline
\multicolumn{1}{|c|}{Question}    & Answer        \\ \hline
What is the capital of Hawaii ?    & Honolulu      \\ 
Who is the wife of Barack Obama ? & Michelle Obama \\ \hline
\end{tabular}
\caption{The use of a knowledge base for question answering.}
\label{tab:kb_qa}
\end{table}

A graph is the most commonly used data structure to store the information of a KB. The nodes of a graph represent the entities, and the directed edges between the nodes represent the relations. This graph structure is shown in our example KB in Figure \ref{fig:kb}. If a KB only includes binary relations where a relation can have two entities, then such a KB can be represented as a set of relation tuples. Each relation tuple consists of two entities and a relation between them. We have shown in Table \ref{tab:kb_tuples} how our example knowledge base in Figure \ref{fig:kb} can be represented as a set of tuples. This set contains the same number of tuples (rows) as the number of directed edges in the graph.

\begin{table}[ht]
\centering
\begin{tabular}{|c|c|c|}
\hline
Entity 1       & Entity 2                & Relation    \\ \hline
Barack Obama   & Michelle Obama          & spouse      \\ 
Michelle Obama & Barack Obama            & spouse      \\ 
Barack Obama   & United State of America & lives\_in   \\ 
Michelle Obama & United State of America & lives\_in   \\ 
Hawaii         & United State of America & located\_in \\ 
Hawaii         & Honolulu                & capital     \\ \hline
\end{tabular}
\caption{A knowledge base represented as set of relation tuples.}
\label{tab:kb_tuples}
\end{table}

\section{Distant Supervision}

Existing KBs such as Freebase, Wikidata, and DBpedia are built manually and it takes much effort and time to do so. However, these KBs still have a large number of missing links. On the other hand, we can find evidence of a large number of relation tuples in free texts. We have included some examples of such tuples and texts in Table \ref{tab:text_tuple}. If we can extract relation tuples automatically from text, we can build a KB from scratch or add new tuples to the existing KBs without any manual effort. But to achieve this goal, we need a large number of texts annotated with relations, each relating two entities, and creating such a corpus manually is a daunting task. 

\begin{table}[ht]
\centering
\begin{tabular}{|c|c|c|l|}
\hline
\multicolumn{1}{|c|}{Relation} & Entity 1 & Entity 2                                    & \multicolumn{1}{c|}{Text}                                                                                                                                                                \\ \hline
acted\_in                        & \textcolor{red}{Meera Jasmine} & \textcolor{blue}{Sootradharan} & \begin{tabular}[c]{@{}l@{}}\textcolor{red}{Meera Jasmine} made her \\debut in the Malayalam \\film ``\textcolor{blue}{Soothradharan}" .\end{tabular}                                                                         \\ \hline
located\_in                    & \textcolor{red}{Chakkarakadavu} & \textcolor{blue}{Kerala}     & \begin{tabular}[c]{@{}l@{}}\textcolor{red}{Chakkarakadavu} is a \\small village to the \\east of the town of \\Cherai, on Vypin \\Island in Ernakulam \\district,  \textcolor{blue}{Kerala}, India .\end{tabular}             \\ \hline
birth\_place                      & \textcolor{red}{Barack Obama} & \textcolor{blue}{Hawaii}  & \begin{tabular}[c]{@{}l@{}}\textcolor{red}{Barack Obama} was born \\in \textcolor{blue}{Hawaii} . \end{tabular} \\ \hline
plays\_for                       & \textcolor{red}{Moussa Sylla} & \textcolor{blue}{Horoya AC}   & \begin{tabular}[c]{@{}l@{}}Fodé \textcolor{red}{Moussa Sylla} is a \\Guinean football player, \\who currently plays for \\\textcolor{blue}{Horoya AC} .\end{tabular}                      \\ \hline
owns                           & \textcolor{red}{MTV Channel} & \textcolor{blue}{Shakthi TV}    & \begin{tabular}[c]{@{}l@{}}\textcolor{red}{MTV Channel} (Pvt) Ltd \\is a Sri Lankan media \\company which owns \\three national television \\channels - \textcolor{blue}{Shakthi TV}, \\Sirasa TV and TV 1 .\end{tabular} \\ \hline
\end{tabular}
\caption{Examples of relation tuples found in free texts.}
\label{tab:text_tuple}
\end{table}

\cite{mintz2009distant}, \cite{riedel2010modeling}, and \cite{hoffmann2011knowledge} proposed the idea of distant supervision to automatically create such text-tuple mapping without any human effort. In distant supervision, the tuples from an existing KB are mapped to a free text corpus such as Wikipedia articles or news articles (e.g., New York Times). The idea of distant supervision is that if a sentence contains two entities of a tuple from a KB, that sentence can be considered as the source of this KB tuple. On the other hand, if a sentence contains two entities from a KB and there is no relation between these two entities in the KB, that sentence is considered as a source of {\em None} tuple between the two entities. These {\em None} samples are useful as distantly supervised models consider only a limited set of positive relations. Any relation outside this set is considered as {\em None} relation. This method can give us a large number of tuple-to-text mappings which can be used to build supervised models for this task. This idea of distant supervision can be extended easily to single-document or multi-document relation extraction.

But the distantly supervised data may contain many noisy samples. Sometimes, a sentence may contain the two entities of a positive tuple, but the sentence may not express any relation between them. These kinds of sentences and entity pairs are considered as noisy positive samples. Another set of noisy samples comes from the way samples for {\em None} relation are created. If a sentence contains two entities from the KB and there is no relation between these two entities in the KB, this sentence and entity pair are considered as a sample for {\em None} relation. But knowledge bases are often not complete and many valid relations between entities in a KB are missing. So it may be possible that the sentence contains information about some positive relation between the two entities, but since that relation is not present in the KB, this sentence and entity pair are incorrectly considered as a sample for {\em None} relation. These kinds of sentences and entity pairs are considered as noisy negative samples.

We include examples of clean and noisy samples generated using distant supervision in Table \ref{tab:noisy_examples}. The KB contains many entities out of which four entities are {\em Barack Obama}, {\em Hawaii}, {\em Karkuli}, and {\em West Bengal}. {\em Barack Obama} and {\em Hawaii} have a {\em birth\_place} relation between them. Karkuli and West Bengal are not connected by any relations in the KB. So we assume that there is no valid relation between these two entities. The sentence in the first sample contains the two entities {\em Barack Obama} and {\em Hawaii}, and it also contains information about {\em Obama} being born in {\em Hawaii}. So this sentence is a correct source for the tuple ({\em Barack Obama}, {\em Hawaii}, {\em birth\_place}). So this is a clean positive sample. The sentence in the second sample contains the two entities, but it does not contain the information about {\em Barack Obama} being born in {\em Hawaii}. So it is a noisy positive sample. In the case of the third and fourth sample, according to distant supervision, they are considered as samples for {\em None} relation. But the sentence in the third sample contains the information for the actual relation {\em located\_in} between {\em Karkuli} and {\em West Bengal}, even though the KB happens not to contain the {\em located\_in} relation relating {\em Karkuli} and {\em West Bengal}. So the third sample is a noisy negative sample. The fourth sample is an example of a clean negative sample.

\begin{table}[ht]
\small
\centering
\begin{tabular}{|l|c|c|c|c|c|}
\hline
\multicolumn{1}{|c|}{Sentence}                                                                                           & Entity 1                                                & Entity 2                                               & \begin{tabular}[c]{@{}c@{}}Distantly \\ Supervised \\ Relation\end{tabular} & \begin{tabular}[c]{@{}c@{}}Actual\\ Relation\end{tabular} & Status \\ \hline
\begin{tabular}[c]{@{}l@{}}\textcolor{red}{Barack Obama} \\was born in\\ \textcolor{blue}{Hawaii} .\end{tabular}                                             & \begin{tabular}[c]{@{}c@{}}\textcolor{red}{Barack} \\ \textcolor{red}{Obama}\end{tabular} & \textcolor{blue}{Hawaii}                                                 & birth\_place                                                          & birth\_place                                          & Clean         \\ \hline
\begin{tabular}[c]{@{}l@{}}\textcolor{red}{Barack Obama} \\ visited \textcolor{blue}{Hawaii} .\end{tabular}                                                 & \begin{tabular}[c]{@{}c@{}}\textcolor{red}{Barack} \\ \textcolor{red}{Obama}\end{tabular} & \textcolor{blue}{Hawaii}                                                 & birth\_place                                                          & None                                                      & Noisy        \\ \hline
\begin{tabular}[c]{@{}l@{}}Suvendu Adhikari \\was born at\\ \textcolor{red}{Karkuli} in Purba \\Medinipur in\\ \textcolor{blue}{West Bengal} .\end{tabular} & \textcolor{red}{Karkuli}                                                 & \begin{tabular}[c]{@{}c@{}}\textcolor{blue}{West} \\ \textcolor{blue}{Bengal}\end{tabular} & None                                                                      & located\_in                                           & Noisy        \\ \hline
\begin{tabular}[c]{@{}l@{}}Suvendu Adhikari, \\ transport minister \\ of \textcolor{blue}{West Bengal}, \\ visited \textcolor{red}{Karkuli} .\end{tabular}  & \textcolor{red}{Karkuli}                                                 & \begin{tabular}[c]{@{}c@{}}\textcolor{blue}{West} \\ \textcolor{blue}{Bengal}\end{tabular} & None                                                                      & None                                                      & Clean         \\ \hline
\end{tabular}
\caption{Examples of distantly supervised clean and noisy samples.}
\label{tab:noisy_examples}
\end{table}

Despite the presence of noisy samples, relation extraction models trained on distantly supervised data have proven to be successful for relation extraction. These models can be used to fill the missing facts of a KB by automatically finding tuples from free texts. It can save much manual effort towards completing an existing KB. 

\section{Task Formalization}

The task of relation extraction is to find relation tuples from free texts automatically. A relation extraction system takes a sentence and a set of relations as input and outputs a set of relation tuples present in the sentence. This task consists of two sub-tasks: (i) entity recognition and (ii) relation classification. In the first sub-task, entities are identified in a sentence. In the second sub-task, for each pair of entities, we classify the relation between the two entities, or that no relation exists between the two entities. This is a pipeline approach to solve this task. Another approach attempts to find the entities and relations jointly (i.e., not in a pipeline approach). In this thesis, we explore deep neural network models for relation extraction at the sentence-level and beyond, in both pipeline and joint extraction approaches.

\section{Scope of the Thesis}

Our goal is to use deep neural network models to find relation tuples from free texts. In our first work, we explore a pipeline approach where we assume that two entities are given and we need to find the relation between them, or that no relation exists between them. In our second work, we explore a joint extraction approach for entities and relations. We have proposed deep neural models for this task and achieve significantly improved performance with both approaches when evaluated on publicly available relation extraction datasets. In our third work, we explore a new multi-hop relation extraction task, where we use multiple documents to find relation tuples. This can help to extract a higher number of relations from knowledge bases than sentence-level relation extraction.

\section{Contributions of the Thesis}

The contributions of this thesis are three-fold as described below.

(1) We find that sentences found using distant supervision can be very long and two entities can be located far from each other in a sentence. The pieces of evidence supporting the presence of a relation between two entities may not be very direct, since the entities may be connected via some indirect links such as a third entity or via co-reference. Relation extraction in such scenarios becomes more challenging, as we need to capture the long-distance interaction among the entities and other words in the sentence. Also, the words in a sentence do not contribute equally in identifying the relation between the two entities. To address this issue, we propose a novel and effective attention model which incorporates syntactic information of the sentence and a multi-factor attention mechanism. Experiments on the New York Times corpus show that our proposed model outperforms prior state-of-the-art models. This work has been published as a full paper in {\em CoNLL} 2019.

(2) In the above-mentioned approach, we have to consider all possible pairing of entities and it will give a large number for {\em None} relations. Most prior work adopted such a pipeline approach, where entities were identified first followed by finding the relations among them, thus missing the interaction among the relation tuples in a sentence. In our next work, we explore how to extract entities and relations jointly. There may be multiple relation tuples present in a text and they may share one or both entities among them. We propose two approaches to use encoder-decoder architecture for jointly extracting entities and relations. In the first approach, we propose a representation scheme for relation tuples which enables the decoder to generate one word at a time like machine translation models. This approach still finds all the tuples present in a sentence with full entity names of different lengths and with overlapping entities. In the second approach, we propose a pointer network-based decoding approach where an entire tuple is generated at every time step. Experiments on the publicly available New York Times corpus show that our proposed approaches outperform previous work and achieve significantly higher F1 scores. This work has been published as a full paper in {\em AAAI} 2020.

(3) Distantly supervised relation extraction models mostly focus on sentence-level relation extraction, where the two entities (subject and object entity) of a relation tuple must appear in the same sentence. This assumption is overly strict and for a large number of relations, we may not find sentences containing the two entities. To solve this problem, we propose multi-hop relation extraction, where the two entities of a relation tuple may appear in two different documents but these documents are connected through some common entities. We can find a chain of entities from the subject entity to the object entity via the common entities. The relation between the subject and object entity can be established using this entity chain. Following this multi-hop approach, we create a dataset for 2-hop relation extraction, where each chain contains exactly two documents. This 2-hop dataset covers a higher number of relations than the previous sentence-level or document-level datasets. We also propose a hierarchical graph convolutional network (HEGCN) model consisting of a two-level hierarchy of graph convolutional networks to solve this task. The first-level GCN of the hierarchy captures the relation among the entity mentions within a document, and the second-level GCN of the hierarchy captures the relation among the entities across the documents in a chain. Our proposed HEGCN model improves the performance on our 2-hop relation extraction dataset and it can be readily extended to N-hop datasets. 

\section{Organization of the Thesis}

This thesis is organized as follows. We give a brief introduction to neural networks in Chapter \ref{chapt:neural}. In Chapter \ref{chapt:relatedwork}, we discuss related work on knowledge bases, named entity recognition, open information extraction, relation extraction, and multi-hop processing. In Chapter \ref{chapt:mfa4re}, we describe our proposed syntax-focused multi-factor attention model for relation extraction. In Chapter \ref{chapt:ed4jere}, we describe how encoder-decoder models can be used effectively to jointly extract entities and relations. We describe our work on the multi-hop relation extraction task in Chapter \ref{chap:mhre}. Finally, we conclude the thesis in Chapter \ref{chapt:conclusion}.

%
\chapter{Neural Networks}
\label{chapt:neural}

In this chapter, I briefly describe the neural networks that we have used in our models and the algorithms that we have used to train these models.

\section{Feed-Forward Neural Networks}

Feed-forward neural networks are a class of neural networks where information flows only in one direction. These networks consist of an input layer, an output layer, and one or more hidden layers. They are often referred to as multi-layer perceptron. Each hidden layer applies a function to its input and forwards the output to the next layer. This function is implemented using a linear transformation followed by a non-linear transformation. This non-linearity helps the feed-forward network to approximate more complex functions. The most widely used non-linear functions are sigmoid ($\sigma$), tanh, and ReLU. Another important non-linear function is softmax which is often used at the output layer for normalization. These non-linear functions are often called activation functions and are also used across other neural networks described in later sections. The following are the definitions of these non-linear functions with input $\mathbf{x}$.
\begin{align}
    &\sigma(\mathbf{x}) = \frac{1}{1+e^{-\mathbf{x}}}\\
    &\text{tanh}(\mathbf{x}) = \frac{e^{\mathbf{x}}-e^{-\mathbf{x}}}{e^{\mathbf{x}}+e^{-\mathbf{x}}}\\
    &\text{ReLU}(\mathbf{x}) = \text{max}(0, \mathbf{x})\\
    &\text{softmax}(\mathbf{x}) = \frac{e^\mathbf{x}}{\sum e^{x}}
\end{align}
A feed-forward network consisting of a single hidden layer implements the following function (Eq. (2.5)) with input $\mathbf{x}$.
\begin{align}
    &\text{FFN}(\mathbf{x}) = \rho(\mathbf{W}_2(\rho(\mathbf{W}_1 \mathbf{x} + \mathbf{b}_1)) + \mathbf{b}_2)
\end{align}
\noindent Here, $\mathbf{W}_1$ and $\mathbf{b}_1$ are the trainable parameters of the hidden layer, whereas $\mathbf{W}_2$ and $\mathbf{b}_2$ are the trainable parameters for the output layer. $\rho(.)$ is any non-linear activation function described above.

\section{Convolutional Neural Networks}

Convolutional neural networks (CNN) \citep{LeCun1989Backpropagation} are a special kind of neural networks that work on grid-like topology such as image data (2D grid of pixels) and text data (1D grid of word vectors). CNN is used to extract important features from data automatically. Here, I will describe the CNN in the context of text data. The tokens in the text can be represented as a sequence of vectors $\{\mathbf{x}_1, \mathbf{x}_2,....., \mathbf{x}_n\}$ where $\mathbf{x} \in \mathbb{R}^{d}$ and $n$ is the sequence length. Eq. (2.6) defines the convolution function for text data. $\Vert$ is the concatenation operation.
\begin{align}
    &c_i = \mathbf{f}^T (\mathbf{x}_{i} \Vert \mathbf{x}_{i + 1} \Vert .... \Vert \mathbf{x}_{i+k-1})\\
    & c_{max} = \text{max}(c_1,c_2,....,c_{n})\\
    & \mathbf{v}_{max} = [c_{max}^1, c_{max}^2, ...., c_{max}^{f_k}]\\
    & c_{avg} = \frac{c_1 + c_2 + .... + c_{n}}{n}\\
    & \mathbf{v}_{avg} = [c_{avg}^1, c_{avg}^2, ...., c_{avg}^{f_k}]
\end{align} 
$\mathbf{f}$ is a convolutional filter vector of length $kd$ where $k$ is the filter width and superscript $T$ represents the transpose operation. The index $i$ moves from $1$ to $n$ and produces a set of scalar values $\{c_1, c_2, .....,c_{n}\}$. These scalar values represent the local features of the data. Convolutional operations are followed by a pooling operation such as max-pooling (Eq. (2.7)) or average pooling (Eq. (2.9)). The pooled values across multiple filters are concatenated to obtain the feature vector. With $f_k$ number of filters, we get a feature vector $\mathbf{v}_{max}$ (Eq. (2.8)) or $\mathbf{v}_{avg}$ (Eq. (2.10)) of length $f_k$.

\section{Recurrent Neural Networks}

\begin{figure*}[ht]
    \centering
    \includegraphics[scale=0.4]{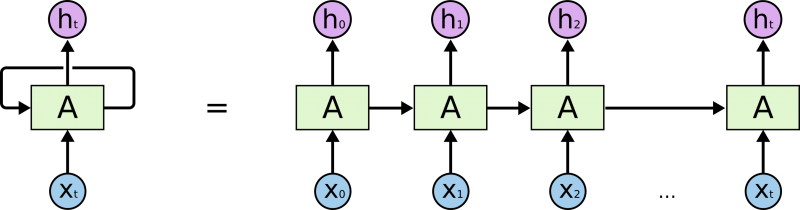}
    \caption{Recurrent Neural Network. Source: \url{http://cstwiki.wtb.tue.nl/index.php?title=File:RNN-unrolled.png}}
    \label{fig:rnn}
\end{figure*}

Recurrent neural networks (RNN) are a special kind of neural networks to process sequential data such as text. These networks contain a feedback loop that helps to remember past information. Figure \ref{fig:rnn} shows the unrolled version of a recurrent network. The unrolled version looks like a multi-layer feed-forward network except that parameters are shared across the time steps in RNN.

Though there are many different versions of RNN, the following is the widely used one \citep{Elman1990FindingSI}.
\begin{align}
    &\mathbf{h}_t = \rho(\mathbf{W}_h \mathbf{x}_t + \mathbf{U}_h \mathbf{h}_{t-1} + \mathbf{b}_h) \\
    &\mathbf{y}_t = \rho(\mathbf{W}_y \mathbf{h}_t + \mathbf{b}_y)
\end{align}
\noindent where $\rho(.)$ is any non-linear activation function, and $\mathbf{x}_t$ and $\mathbf{h}_t$ are the input and hidden state at time step $t$, respectively. $\mathbf{W}_h$, $\mathbf{W}_y$, $\mathbf{U}_h$, $\mathbf{b}_h$, and $\mathbf{b}_y$ are the network parameters and are learned during training. The hidden state ($\mathbf{h}_t$) is responsible for remembering past information for this network.

Due to the successive use of the activation function on the hidden state of the RNN (Eq. (2.11)), RNN suffers from the vanishing or exploding gradient problem for long sequences. The exploding gradient problem can be solved by clipping the gradient up to a certain threshold, but the vanishing gradient problem cannot be solved that easily. \citet{hochreiter1997long} proposed long short-term memory (LSTM) and \citet{cho2014properties} proposed gated recurrent unit (GRU) to address the vanishing gradient problem for long sequences. We have used the LSTM network extensively in this thesis and will discuss it briefly here.

\subsection{Long Short-Term Memory}

\citet{hochreiter1997long} solved the vanishing gradient problem in RNNs by introducing control gates in the network. They used three control gates as forget gate, input gate, and output gate. The forget gate ($\mathbf{f}_t$) is used to control how much past information to forget. The input gate ($\mathbf{i}_t$) is used to control how much current information has to be kept. The output gate ($\mathbf{o}_t$) is used to control the exposure of the output. They also added a cell state ($\mathbf{c}_t$) in RNN to remember the past information effectively. The following equations summarize the functionality of an LSTM network.
\begin{align}
    & \mathbf{i}_t = \sigma(\mathbf{W}_i \mathbf{x}_t + \mathbf{U}_i \mathbf{h}_{t-1} + \mathbf{b}_i) \nonumber\\
    & \mathbf{f}_t = \sigma(\mathbf{W}_f \mathbf{x}_t + \mathbf{U}_f \mathbf{h}_{t-1} + \mathbf{b}_f) \nonumber\\
    & \tilde{\mathbf{c}}_t = \tanh(\mathbf{W}_c \mathbf{x}_t + \mathbf{U}_c \mathbf{h}_{t-1} + \mathbf{b}_c) \nonumber\\
    & \mathbf{c}_t = \mathbf{i}_t \circ \tilde{\mathbf{c}}_{t} + \mathbf{f}_t \circ \mathbf{c}_{t-1} \nonumber\\
    & \mathbf{o}_t = \sigma(\mathbf{W}_o \mathbf{x}_t + \mathbf{U}_o \mathbf{h}_{t-1} + \mathbf{b}_o) \nonumber\\
    & \mathbf{h}_t = \mathbf{o}_t \circ \tanh(\mathbf{c}_t)
\end{align}
\noindent The input gate ($\mathbf{i}_t$), forget gate ($\mathbf{f}_t$), and output gate ($\mathbf{o}_t$) use sigmoid ($\sigma$) activation function to control the information flow. A $0$ output of this activation means no information will pass through and a $1$ output of this activation means full information will pass through. The cell state ($\mathbf{c}_t$) is updated without any activation function, thus the problem of gradient becoming very low (vanishing gradient) is eliminated. 

\section{Neural Attention Networks}

Simple CNN and LSTM networks treat all the words that are present in a sentence equally. But this is not appropriate for most NLP tasks, where some words carry more information than others specific to the task. We need neural models that can automatically learn which words are more important than others. These kinds of networks are called attention networks. An attention network includes one or more attention layers where each layer contains trainable parameters. We learn the weights of these parameters during the training process. An attention layer assigns normalized attention scores to all the words, where some words get higher scores, and some get lower scores. In this way, important words contribute more to the current prediction than the others. This idea leads to better performance on the task. \cite{Bahdanau2014NeuralMT} used such attention networks for neural machine translation. \cite{vaswani2017attention} showed that simple feed-forward neural networks with attention achieve performance similar to LSTM networks.

\section{Graph Convolutional Networks}

\begin{figure*}[ht]
    \centering
    \includegraphics[scale=0.5]{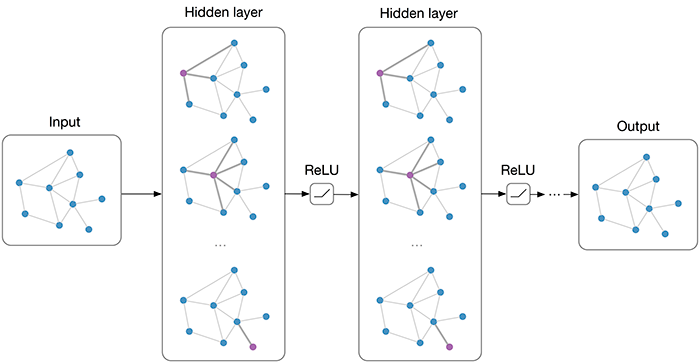}
    \caption{A Multi-layer Graph Convolutional Network. Source: \url{https://tkipf.github.io/graph-convolutional-networks}}
    \label{fig:gcn}
\end{figure*}

Convolutional neural networks (CNN) and recurrent neural networks (RNN) mostly work on linear data. But many real-world datasets come in the form of a graph structure. It is very challenging to use CNN or RNN on an arbitrary graph structure. Graph convolutional networks (GCN) are generalized neural architectures that can work on any arbitrary graph structure. Figure \ref{fig:gcn} shows a general architecture of a multi-layer graph convolutional network.

A graph $\mathcal{G}=\{\mathcal{V}, \mathcal{E}\}$ consists of a set of nodes $\mathcal{V}$ and a set of edges $\mathcal{E}$ that connect the nodes. A graph convolutional network takes the following two inputs:

\begin{enumerate}
    \item A feature vector $\mathbf{x}_i \in \mathbb{R}^d$ for every node $i$ in $\mathcal{G}$, where $d$ is the dimension of the input feature vector. The vectors of $n$ nodes in $\mathcal{G}$ can be summarized as a matrix $\mathbf{X} \in \mathbb{R}^{n \times d}$.
    \item An adjacency matrix $\mathbf{A}$ of size $n \times n$ which describes the graph structure. Generally, self-loops are added in $\mathbf{A}$ and it is normalized. 
\end{enumerate}

A graph convolutional network produces a node-level output $\mathbf{Z} \in \mathbb{R}^{n \times f}$ where $f$ is the dimension of the output feature vector. A pooling mechanism can be used to model the graph-level output from the node-level output depending on the task.

Every layer of a GCN can be described by the following function:
\begin{align}
    &\mathbf{H}^{l+1} = f(\mathbf{H}^l, \mathbf{A})
\end{align}
\noindent where $\mathbf{H}^0=\mathbf{X}$ and $\mathbf{H}^L=\mathbf{Z}$. $L$ is the number of layers in the GCN and $f(.,.)$ is a parameterized function. As an example, the following simple function can be used as $f(.,.)$ in GCN.
\begin{align}
    &f(\mathbf{H}^l, \mathbf{A}) = \rho(\mathbf{A}\mathbf{W}^l\mathbf{H}^l)
\end{align}
\noindent where $\mathbf{W}^l$ is the weight matrix of the $l$th layer and $\rho(.)$ is any non-linear activation function such as ReLU.

\section{Neural Network Training}

Neural network models use a lot of parameters and we need to approximate the optimal values of these parameters during training. The parameters are initialized randomly and then during training, they are updated. A loss function $C$ is used to estimate the difference between the network output and target output. The goal of the training process is to minimize this loss function. Network parameters $\theta$ are optimized using a gradient descent algorithm so that the loss is minimized.
\begin{align}
&\theta_{t+1} = \theta_t - \eta \dfrac{\partial {C}}{\partial \theta_t} 
\end{align}
\noindent Here, $\eta$ is called the step size or learning rate which is a hyperparameter. The loss function $C$ can be calculated over the entire training data, known as batch training. But this is computationally very expensive if the size of the training data is large. Another way of measuring the loss function $C$ is by each instance of the training data. So we need to update the network parameters $\theta$ after each instance. This is the stochastic gradient descent algorithm (SGD). However, this will make the training process very slow for large training data. A middle path is to update the parameters after each mini-batch of smaller sizes like 32/40/50. This is called mini-batch training. Samples in each mini-batch are chosen randomly from the training data without replacement. We have used this mini-batch training to train our models.

The SGD algorithm has two major problems. First, the value of $\eta$ has to be chosen manually. If it is set too high, then the network may not converge and if it is set too low, convergence will be very slow. Second, all parameters are updated with the same learning rate. However, different parameters in a neural network might need different learning rates to be optimized properly. To solve these problems, several adaptive optimization algorithms are proposed such as Adagrad \citep{duchi2011adaptive}, RMSprop (unpublished work by Geoffrey Hinton, 2012), and Adam \citep{kingma2014adam}. These optimization algorithms adapt the learning rate $\eta$ for each parameter differently during training. We have used Adagrad and Adam optimization algorithms for our models.

Another important aspect of neural network training is back-propagation. Neural networks consist of multiple layers and each layer has its parameters. We need to update the parameters of all the layers based on the loss function. This is achieved by the chain rule of differentiation. First, the parameters of the output layer are updated, then the parameters of the layer below it are updated, and so on. This process is repeated until the parameters of all the layers are updated. This process of updating network parameters starting from the output layer towards the input layer in backward fashion is referred to as back-propagation.

%
\chapter{Related Work}
\label{chapt:relatedwork}

In this chapter, we describe the different works that are connected to relation extraction. We first describe the different knowledge bases available that can be used for distant supervision. Next, we discuss the named entity recognition task which can be used in pipeline approaches of relation extraction. Then, we include a brief introduction of open information extraction and its limitations. We then move to discuss the datasets and the prior works on relation extraction. Finally, we finish the chapter by introducing multi-hop natural language processing.  

\section{Knowledge Bases}
In recent years, a number of research teams have created knowledge bases (KB) which cover entities across multiple domains. Here, we briefly describe a few such KBs: YAGO \citep{suchanek2007yago}, Freebase \citep{bollacker2008freebase}, Wikidata \citep{wikidata}, DBpedia \citep{bizer2009dbpedia}, and NELL \citep{mitchell2014never}.

\subsubsection{YAGO}
Yet Another Great Ontology (YAGO) is built by extracting facts from the structured data that is present in Wikipedia and WordNet. YAGO uses Wikipedia category pages to extract classes, entities, and relations between them. For example, `Zidane' is present in the category of `French football players', so YAGO extracts relations like (Zidane, is\_citizen\_of, France) and (Zidane, is\_a, football player) using handcrafted rules. However, Wikipedia category pages are not useful to build an ontology. As an example, `Zidane' is also present in the category of `French football', but `Zidane' is a football player, not a football. WordNet, on the other hand, provides a clean class hierarchy, which is used by YAGO to build an ontology. YAGO efficiently combines Wikipedia and WordNet to build the ontology and extract facts. As of 31 Oct 2020, YAGO contains more than 50 million entities and 2 billion facts. 

\subsubsection{Freebase}

Freebase is a scalable database for relation tuples built by Metaweb Technologies\footnote{\url{https://en.wikipedia.org/wiki/Metaweb}} Inc. in 2007. Google acquired the company in 2010 and shut down Freebase in 2016 after deciding to transfer all its data to Wikidata. At the time of its release in 2007, Freebase had 125 million entities and around 7,000 relations. 

\subsubsection{Wikidata}

Wikidata is another large, free, and collaborative knowledge base built by Wikimedia\footnote{\url{https://www.wikimedia.org}}. Along with facts, Wikidata stores the corresponding Wikipedia source page so that users can verify its validity. It also provides the aliases for every relation and its brief description. It stores facts in the form of {\it items} and {\it statements}. An item in Wikidata represents an entity, and each item has multiple statements. Each statement has a claim which consists of a property and its value. These properties are like relations. As of 31 Oct 2020, Wikidata contains more than 90 million items and more than 1.2 billion statements for over 8,060 properties.

\subsubsection{DBpedia}

DBpedia is another crowd-sourced project which extracts structured data from multi-lingual Wikidata. It extracts information from Wikidata in two formats: raw infobox extraction and mapping-based infobox extraction. In raw infobox extraction, DBpedia extracts Wikidata infobox information and does not map them to any ontology. In mapping-based extraction, an ontology is created with a community effort, and Wikidata infobox extractions are mapped to that ontology. As of 31 Oct 2020, the English version of DBpedia contains 4.58 million entities and 580 million facts about them. 

\subsubsection{NELL}

Never-Ending Language Learner (NELL) is another system that automatically extracts facts from the Web. NELL is a semi-supervised system which started with an initial ontology of hundreds of categories and relations. It uses around 10-15 seed examples for each category and relation. Along with this, it takes 500 million web pages and accesses the remaining web pages through Google search API as input. The goal of this system is to learn new instances of categories and relations. This system runs 24/7 to learn new facts and uses these facts to improve its learning algorithms for better extraction. As of 31 Oct 2020, NELL has accumulated over 50 million facts by reading the Web.

\section{Named Entity Recognition}

Named entity recognition (NER) is an important task in NLP and it helps many other NLP applications such as information retrieval, question answering, and relation extraction. As we discussed before, NER is the first step for pipeline relation extraction models. The goal of the NER task is to identify named entities such as person names, location names, and organization names in a text. A sequence tagging approach, called the `BIESO' tagging scheme, is used to identify the entities in a text. In this tagging scheme, every token of a text is assigned a tag out of the 5 tags `B', `I', `E', `S', and `O'. `B' refers to the beginning of an entity, `I' refers to the inside of an entity, and `E' refers to the end of an entity. `S' is used to tag single-token entities and `O' is used to tag the non-entity tokens. An entity type is appended as the suffix to a `B/I/E/S/O' tag to indicate the type of the entity such as PERSON, LOCATION, ORGANIZATION, GEOPOLITICAL ENTITY, etc.  Several named entity recognition shared tasks were organized with accompanying datasets such as MUC \citep{Grishman1996MessageUC}, CoNLL 2002 \citep{Sang2002IntroductionTT}, CoNLL 2003 \citep{tjong2003introduction}, and ACE04 \citep{doddington2004automatic}. 

Supervised learning models are popularly used to tackle the NER task. \citet{zhou2002named} used the Hidden Markov Model (HMM) for NER on MUC-6 datasets. They used orthographic features, trigger words, etc in their model. \citet{Malouf2002MarkovMF} used the Hidden Markov Model with maximum entropy for the same task. The winner of the CoNLL 2002 shared task \citep{Carreras2002NamedEE} used binary AdaBoost classifier with features like capitalization, trigger words, gazetteers, etc. \citet{Takeuchi2002UseOS} used SVM model for this task. They used part-of-speech tags, orthographic features, nearby words, and tags of previous tokens in their SVM model. \citet{chieu2002named} proposed a maximum entropy approach where they used document-level information along with sentence-level information for the NER task. They used MUC-6 and MUC-7 datasets for their experiments. \citet{chieu2003named} used a similar approach for the CoNLL 2003 NER shared task. 

\subsubsection{Neural NER Models}

\citet{Collobert2008AUA} proposed one of the first neural models for the NER task. They represented the different features used for NER as vectors in their model. Later, they replaced these manually extracted features with word vectors \citep{Collobert2011NaturalLP}. \citet{Huang2015BidirectionalLM} proposed an LSTM-based neural model with word embeddings for this task and achieved a high F1 score on the CoNLL 2003 dataset. \citet{Ma2016EndtoendSL} and \citet{chiu2016named} used character-based embeddings along with word embeddings for this task. Their experiments on the CoNLL 2003 dataset show that character-level embeddings help in improving the performance of the task. \citet{lample2016neural} also used an LSTM-CRF model for this task and used CoNLL 2002 and 2003 datasets for their experiments. Recently, contextualized word representations such as ELMo \citep{peters2018deep} and BERT \citep{devlin2019bert} have proven very effective for the NER task. Models with such representation \citep{strakova2019neural} achieved state-of-the-art performance on the CoNLL 2003 dataset.

\subsubsection{Zero-Shot or Few-Shot NER}

Zero-shot or few-shot named entity recognition is popular for low resource languages or domains. \cite{Ma2016LabelEF} proposed a label embedding method for the zero-shot NER task. They used prototypical and hierarchical information to learn the label embeddings and adapted a zero-shot framework for the NER task. \cite{Fritzler2019FewshotCI} proposed a semi-supervised learning technique to tackle the NER task using a prototypical network in a few-shot setting. It learns intermediate representations of words that cluster well into named entity classes. This property of the model allows classifying words with a limited number of training instances, and can potentially be used as a zero-shot learning method. \cite{BARI2020ZeroResourceCN} proposed an unsupervised cross-lingual NER model that can transfer knowledge from one language to another without using any annotated data or parallel corpora. 

\section{Open Information Extraction}

Open information extraction (OpenIE) is a generic form of relation extraction which can extract any kind of relations with its arguments from free text. Most OpenIE systems use hand-crafted rules or predefined sentence structures to extract entities and relations from sentences. The advantage of such a system is that it can work on texts from any domain. KnowItAll \citep{etzioni2004knowitall}, TEXTRUNNER \citep{yates2007textrunner}, REVERB \citep{etzioni2011reverb}, SRL-IE \citep{christensen2011srlie}, OLLIE \citep{schmitz2012ollie}, and RELNOUN \citep{pal2016relnoun} are some examples of OpenIE systems which can extract reliable facts from text. 

\subsubsection{KnowItAll}

This is a rule-based OpenIE system that uses hand-made patterns to extract facts from the Web. It also assigns a confidence score to the extracted tuples based on point-wise mutual information (PMI) between words associated with the identified entity or triple and pre-defined phrases for each entity or relation.

\subsubsection{TEXTRUNNER}

This system is built based on the idea of the KnowItAll system, but there is no need to give hand-crafted patterns to the system. It includes a self-supervised learner that uses dependency parse trees to label its training data as positive or negative samples. It marks the noun phrases in a dependency parse tree as arguments and then finds the relation words between those arguments. It groups a tuple as positive or negative based on certain syntactic structures like the length of the dependency chain between two arguments, the path between two arguments (that it should not cross sentence boundary, etc. Then it uses these samples to train a naive Bayes classifier to determine the trustworthiness of future tuples.

\subsubsection{REVERB}

This is an improved version of the TEXTRUNNER system. It uses additional syntactic and lexical constraints to remove incoherent and uninformative extractions.

\subsubsection{SRL-IE}

This is an open information extraction system based on a semantic role labeling (SRL) system built at UIUC \citep{punyakanok2008importance}. SRL is a common NLP task of detecting semantic arguments associated with a predicate in a sentence, and the classification of the arguments into different semantic roles like agent, patient, instrument, etc. The predicates and arguments extracted by an SRL system can be considered as relation tuples.

\subsubsection{OLLIE}

Open Language Learning for Information Extraction (OLLIE) attempts to fix the major issues of the REVERB system. REVERB only extracts a relation based on a verb. But some relations are based on noun phrases and not verbs. OLLIE starts with tuples extracted by REVERB. It collects sentences from the Web and then maps these tuples to the sentences. Using the dependency parse tree of these sentences, OLLIE tries to generate some generic patterns for each relation. OLLIE applies these learned patterns to extract more tuples from texts. 

\subsubsection{RELNOUN}

RELNOUN is an open relation extraction module which is used to extract relation tuples based on nouns instead of verbs. This system extracts tuples from titles and entity attributes. RELNOUN uses part-of-speech (POS) tags and noun phrase (NP) chunk patterns to extract noun-based tuples.

\subsubsection{Limitations of OpenIE}

Though open information extraction systems can extract a large number of tuples from free texts, they have two major limitations:
\begin{enumerate}
\item They extract a large number of uninformative tuples. Since every verb is considered a potential relation, the number of uninformative tuples will be very large. Removing uninformative tuples is a challenging task.
\item They do not normalize the relations. Every verb is considered a separate relation by open IE systems. However, different verbs can express the same relation. Open IE systems do not group different verbs with the same meaning into a single relation.
\end{enumerate}

\section{Relation Extraction}

The limitations of open IE systems can be solved using closed domain relation extraction (RE). We use supervised learning algorithms for this task. These models work with a set of pre-defined relations. Thus the issues of uninformative relations and normalization of relations do not arise. We describe prior research and the available datasets on relation extraction below.

\subsection{Pipeline Extraction Approaches}

In the beginning of IE research, pipeline approaches were quite popular. A pipeline approach has two steps: (i) First, a named entity recognizer is used to identify the named entities in a text. (ii) Next, a classification model is used to find the relation between a pair of entities. The named entities identified in the first step are mapped to the KB entities. Using distant supervision, a large number of text-tuple pairs can be generated and they can be used for creating supervised learning models for this task.

\subsubsection{Feature-Based Models}

\citet{mintz2009distant} proposed a feature-based model for this task. They used lexical features such as the sequence of words between two entities and their part-of-speech (POS) tags, a flag indicating which entity appears first, $k$ tokens to the left of entity 1 and $k$ tokens to the right of entity 2, syntactic features such as dependency path between two entities, and named entity types of the two entities in their model. \citet{riedel2010modeling} proposed multi-instance learning for this task to mitigate the problem of noisy sentences obtained using the distant supervision method. They used a factor graph to explicitly model the decision of whether two entities are related and whether this relation is mentioned in a given sentence. Also, they applied constraint-driven semi-supervision to train their model without any knowledge about which sentences express the relations. Their multi-instance learning model significantly improves the performance over the model proposed by \citet{mintz2009distant}. 

\citet{hoffmann2011knowledge} and \citet{mimlre} proposed the idea of multi-instance multi-relations to solve the problem of overlapping relations. They used probabilistic graphical models that take a bag of sentences containing two entities as input and find all possible relations between them. Similarly, \citet{ren2017cotype} used a feature-based model to jointly predict the relation between two entities and their fine-grained types. They used features like the head tokens of two entities, tokens of two entities, tokens between the two entities, their POS tags, ordering of the two entities, the distance between them, and the Brown cluster\footnote{\url{https://github.com/percyliang/brown-cluster}} of each token in their model. They proposed a joint optimization framework to learn the entity embeddings, relation embeddings, and fine-grained type embeddings of the entities together.

\subsubsection{CNN-Based Neural Models}

Distributed representations of words as word embeddings have transformed the way that natural language processing tasks like IE can be tackled. Word2Vec \citep{mikolov2013distributed} and GloVe \citep{pennington2014glove} are two sets of large and publicly available word embeddings that are used for many NLP tasks. Most neural network-based models for information extraction have used the distributed representation of words as their core component. The high dimensional distributed representation of words can encode important semantic information about words, which is very helpful for identifying the relations among the entities present in a sentence. Initially, neural models also follow the pipeline approach to solve this task.

\citet{zeng2014relation} used a convolutional neural network for relation extraction. They used the pre-trained word embeddings of \cite{turian2010word} to represent the tokens in a sentence and used two distance embedding vectors to represent the distance of each word from the two entities. They used a convolutional neural network (CNN) and max-pooling operation to extract a sentence-level feature vector. This sentence representation is passed to a feed-forward neural network with a softmax activation layer to classify the relation. 

\citet{zeng2015distant} introduced a piecewise convolutional neural network (PCNN) to improve relation extraction. \citet{zeng2014relation} applied the max-pooling operation across the entire sentence to get the single important feature from the entire sentence for a particular convolutional filter. In PCNN, the max-pooling operation is not performed for the entire sentence. Instead, the sentence is divided into three segments: from the beginning to the argument appearing first in the sentence, from the argument appearing first in the sentence to the argument appearing second in the sentence, and from the argument appearing second in the sentence to the end of the sentence. Max-pooling is performed in each of these three segments and for each convolutional filter to obtain three feature values. A sentence-level feature vector is obtained by concatenating all such feature values and is given to a feed-forward neural network with a softmax activation layer to classify the relation. 

\subsubsection{Attention-Based Neural Models}

Recently, attention networks have proven very useful for different NLP tasks. \citet{huang2016attention} and \citet{jat2018attention} used word-level attention model for single-instance sentence-level relation extraction. \citet{huang2016attention} proposed a combination of a convolutional neural network model and an attention network. First, a convolution operation with max-pooling is used to extract the global features of the sentence. Next, attention is applied to the words of the sentence based on the two entities separately. The word embedding of the last token of an entity is concatenated with the embedding of every word. This concatenated representation is passed to a feed-forward layer with tanh activation and then another feed-forward layer with softmax to get a scalar attention score for every word of that entity. The word embeddings are averaged based on the attention scores to get the attentive feature vectors. The global feature vector and two attentive feature vectors for the two entities are concatenated and passed to a feed-forward layer with softmax to determine the relation.

\citet{jat2018attention} used a bidirectional gated recurrent unit (Bi-GRU) \citep{cho2014properties} to capture the long-term dependency among the words in the sentence. The tokens vectors $\mathbf{x}_t$ are passed to a Bi-GRU layer. The hidden vectors of the Bi-GRU layer are passed to a bi-linear operator which is a combination of two feed-forward layers with softmax to compute a scalar attention score for each word. The hidden vectors of the Bi-GRU layer are multiplied by their corresponding attention scores for scaling up the hidden vectors. A piecewise convolution neural network \citep{zeng2015distant} is applied to the scaled hidden vectors to obtain the feature vector. This feature vector is passed to a feed-forward layer with softmax to determine the relation.

\citet{lin2016neural} have used attention model for multi-instance relation extraction. They applied attention over a bag of independent sentences containing two entities to extract the relation between them. First, CNN-based models are used to encode the sentences in a bag. Then a bi-linear attention layer is used to determine the importance of each sentence in the bag. This attention helps to mitigate the problem of noisy samples obtained by distant supervision to some extent. The idea is that clean sentences get higher attention scores over the noisy ones. The sentence vectors in the bag are merged in a weighted average fashion based on their attention scores. The weighted average vector of the sentences is passed to a feed-forward neural network with softmax to determine the relation. This bag-level attention is used only for positive relations and not used for {\em None} relation. The reason is that the representations of the bags that express no relations are always diverse and it is difficult to calculate suitable weights for them.

\citet{ye2019distant} used intra-bag and inter-bag attention networks in a multi-instance setting for relation extraction. Their intra-bag attention is similar to the attention used by \citet{lin2016neural}. Additionally, they used inter-bag attention to address the noisy bag problem. They divide the bags belonging to a relation into multiple groups. The attention score for each bag in a group is obtained based on the similarity of the bags to each other within the group. This inter-bag attention is used only during training as we do not know the relations during testing.

\subsubsection{Dependency-Based Neural Models}

Some previous works have incorporated the dependency structure information of sentences in their neural models for relation extraction. \citet{Xu2015ClassifyingRV} used a long short-term memory network (LSTM) \citep{hochreiter1997long} along the shortest dependency path (SDP) between two entities to find the relation between them. Each token along the SDP is represented using four embeddings -- pre-trained word vector, POS tag embedding, embedding for the dependency relation between the token and its child in the SDP, and embedding for its WordNet \citep{Fellbaum2000WordNetA} hypernym. They divide the SDP into two sub-paths: (i) The left SDP which goes from entity 1 to the common ancestor node (ii) The right SDP which goes from entity 2 to the common ancestor node. This common ancestor node is the lowest common ancestor between the two entities in the dependency tree. The token vectors along the left SDP and right SDP are passed to an LSTM layer separately. A pooling layer is applied to the hidden vectors to extract the feature vector from the left SDP and right SDP. These two vectors are concatenated and passed to a classifier to find the relation. 

\citet{liu2015dependency} exploited the shortest dependency path (SDP) between two entities and the sub-trees attached to that path (augmented dependency path) for relation extraction. Each token in the SDP is represented using its pre-trained embedding and its sub-tree representation. The sub-tree representation of a token is obtained from the sub-tree of the dependency tree where the token is the root node. The dependency relations are represented using trainable embeddings. Each node in the sub-tree of a token receives information from its children including the dependency relations. The sub-tree representation of the token is obtained by following the sub-tree rooted at the token from its leaf nodes to the root in a bottom-up fashion. Next, they use CNN with max-pooling on the vectors of the sequence of the tokens and dependency relations across the SDP. The output of the max-pooling operation is passed to a classifier to find the relation.

\citet{miwa2016end} used a tree LSTM network along the shortest dependency path (SDP) between two entities to find the relation between them. They used a bottom-up tree LSTM and top-down tree LSTM in their model. In the bottom-up tree LSTM, each node receives information from all of its children. The hidden representation of the root node of this bottom-up tree LSTM is used as the final output. In the top-down tree LSTM, each node receives the information from its parent node. The hidden representations of the head token of two entities are the final output of this tree LSTM. The representations of the bottom-up tree LSTM and top-down tree LSTM are concatenated and passed to a classifier to find the relation. They showed that using the SDP tree over the full dependency tree is helpful as unimportant tokens for the relation are ignored in the process.

\subsubsection{Graph-Based Neural Models}

Graph-based models are popular for many NLP tasks as they work on non-linear structures. \citet{Quirk2017DistantSF} proposed a graph-based model for cross-sentence relation extraction. They built a graph from the sentences where every word is considered as a node in the graph. Edges are created based on the adjacency of the words, dependency tree relations, and discourse relations. They extract all the paths from the graph starting from entity 1 to entity 2. Each path is represented by features such as lexical tokens, the lemma of the tokens, POS tags, etc. They use all the path features to find the relation between the two entities. 

\citet{peng2017cross} and \citet{Song2018NaryRE} used a similar graph for N-ary cross-sentence relation extraction. Rather than using explicit paths, they used an LSTM on a graph. A graph LSTM is a general structure for a linear LSTM or tree LSTM. If the graph contains only the word adjacency edges, then the graph LSTM becomes a linear LSTM. If the graph contains the edges from the dependency tree, it becomes a tree LSTM. A general graph structure may contain cycles. So \citet{peng2017cross} divides this graph into two directed acyclic graphs (DAG), where the forward DAG contains only the forward edges among the tokens and the backward DAG contains only the backward edges among the tokens. Each node has a separate forget gate for each of its neighbors. It receives information from the neighbors and updates its hidden states using LSTM equations \citep{hochreiter1997long}. If we only consider the word adjacency edges, this graph LSTM becomes a bi-directional linear LSTM. \citet{Song2018NaryRE} did not divide the graph into two DAGs, but directly used the graph structure to update the states of the nodes. At time step $t$, each node receives information from its neighbor from the previous time step and update its hidden states using LSTM equations. This process is repeated $k$ number of times where $k$ is a hyper-parameter.

\citet{Kipf2017SemiSupervisedCW} and \citet{velickovic2018graph} proposed a graph convolutional network (GCN) model which used simple linear transformations to update the node states, unlike the graph LSTMs used by \citet{peng2017cross} and \citet{Song2018NaryRE}. \citet{Kipf2017SemiSupervisedCW} gave equal weights to the edges, whereas \citet{velickovic2018graph} used an attention mechanism to assign different weights to the edges. \citet{vashishth2018reside}, \citet{zhang2018graph}, and \citet{guo2019aggcn} used graph convolutional networks for sentence-level relation extraction. They considered each token in a sentence as a node in the graph and used the syntactic dependency tree to create a graph structure among the nodes. \citet{vashishth2018reside} used the GCN in a multi-instance setting. They used a Bi-GRU layer and a GCN layer over the full dependency tree of the sentences to encode them. The sentence representations in a bag were aggregated and passed to a classifier to find the relation. Following \citet{miwa2016end}, \citet{zhang2018graph} used only the shortest dependency path (SDP) tree to build the adjacency matrix for the graph. Along with the SDP tree, they included the edges that are distance $K$ away from the SDP where $K$ is a hyper-parameter. \citet{guo2019aggcn} proposed a soft pruning strategy over the hard pruning strategy of \citet{zhang2018graph} in their GCN model. They considered the full dependency tree to build the adjacency matrix but using a multi-head self attention-based soft pruning strategy, they can identify the important and unimportant edges in the graph.

\citet{sahu2019inter}, \citet{christopoulou2019connecting}, and \citet{Nan2020ReasoningWL} used GCN for document-level relation extraction. \citet{sahu2019inter} considered each token in a document as a node in a graph. They used syntactic dependency tree edges, word adjacency edges, and coreference edges to create the connections among the nodes. \citet{christopoulou2019connecting} considered the entity mentions, entities, and sentences in a document as nodes of a graph. They used rule-based heuristics to create the edges among these nodes. In their graph, each node and each edge were represented by vectors. GCN was used to update the vectors of nodes and edges. Finally, the edge vector between the two concerned entities was passed to a classifier to find the relation. \citet{Nan2020ReasoningWL} considered the entity mentions, entities, and tokens on the shortest dependency path between entity mentions as nodes in a graph. They used a structure induction module to learn the latent structure of the document-level graph. A multi-hop reasoning module was used to perform inference on the induced latent structure, where representations of the nodes were updated based on an information aggregation scheme.

\subsubsection{Contextualized Embedding-Based Neural Models}

Contextualized word embeddings such as ELMo \citep{peters2018deep}, BERT \citep{devlin2019bert}, and SpanBERT \citep{Joshi2019SpanBERT} can be useful for relation extraction. These language models are trained on large corpora and can capture the contextual meaning of words in their vector representations. All neural models that are proposed for relation extraction use word representations such as Word2Vec \citep{mikolov2013distributed} or GloVe \citep{pennington2014glove} in their word embedding layer. Contextualized embeddings can be added in the embedding layer of the relation extraction models to improve their performance further. The SpanBERT model shows significant improvement in performance on the TACRED dataset. \citet{Joshi2019SpanBERT} replaced the entity 1 token with its type and SUBJ such as PER-SUBJ and entity 2 token with its type and OBJ such as LOC-OBJ in the sentences to train the model. Finally, they used a linear classifier on top of the CLS token vector to find the relation.

\citet{Wang2019FinetuneBF} proposed two-step fine-tuning of BERT for document-level relation extraction on the DocRED dataset. In the first step, they used BERT to identify whether or not there is a relation between two entities. In the second step, they used BERT to classify the relation. \citet{Nan2020ReasoningWL} also used BERT in their model to show that it significantly improved the performance on the DocRED dataset compared to GloVe vectors. \citet{Han2020AND} used BERT to identify all possible relations among the entity pairs in documents in a single pass. They used entity types and special tokens to mark all the entity mentions in documents. All entity mentions of an entity received the same special token. Documents were passed to a pre-trained BERT model. An entity mention vector was obtained by averaging the BERT outputs of the entity mention tokens. An entity vector was obtained by averaging all the entity mention vectors of that entity. A bilinear classifier was used to classify the relation between two entities. \citet{Tang2020HINHI} proposed a hierarchical inference network for document-level relation extraction. They also showed that using BERT in their model improved performance significantly.

\subsection{Noise Mitigation for Distantly Supervised Data}

The presence of noisy samples in distantly supervised data adversely affects the performance of models. Researchers have used different techniques in their models to mitigate the effects of noisy samples to make them more robust. Multi-instance relation extraction is one of the popular methods for noise mitigation. \citet{riedel2010modeling}, \citet{hoffmann2011knowledge}, \citet{mimlre}, \citet{lin2016neural}, \citet{yaghoobzadeh2017noise}, \citet{vashishth2018reside}, \citet{wu2018improving}, and \citet{ye2019distant} used this multi-instance learning concept in their proposed relation extraction models. For each entity pair, they used all the sentences that contained these two entities to find the relation between them. Their goal was to reduce the effect of noisy samples using this multi-instance setting. They used different types of sentence selection mechanisms to give importance to the sentences that contained relation-specific keywords and ignored the noisy sentences. \citet{ren2017cotype} and \citet{yaghoobzadeh2017noise} used the multi-task learning approach for mitigating the influence of noisy samples. They used fine-grained entity typing as an extra task in their model.

\citet{wu2017adversarial} used an adversarial training approach for the same purpose. They added noise to the word embeddings to make the model more robust for distantly supervised training. \citet{Qin2018DSGANGA} used a generative adversarial network (GAN) to address the issue of noisy samples in relation extraction. They used a separate binary classifier as a generator in their model for each positive relation class to identify the true positives for that relation and filter out the noisy ones. \citet{qin2018robust} used reinforcement learning to identify the noisy samples for the positive relation classes. \citet{He2020ImprovingNR} used reinforcement learning to identify the noisy samples for the positive relations and then used the identified noisy samples as unlabeled data in their model. \citet{Shang2020AreNS} used a clustering approach to identify the noisy samples. They assigned the correct relation label to these noisy samples and used them as additional training data in their model.

\subsection{Zero-Shot or Few-Shot Relation Extraction}

Distantly supervised datasets cover a small subset of relations from the KBs. Existing KBs such as Freebase, Wikidata, and DBpedia contain thousands of relations. Due to the mismatch of the surface form of entities in KBs and texts, distant supervision cannot find adequate training samples for most relations in KBs. It means that distantly supervised models cannot fill the missing links belonging to these uncovered relations. Zero-shot or few-shot relation extraction can address this problem. These models can be trained on a set of relations and can be used for inferring another set of relations. 

\citet{Levy2017ZeroShotRE} and \citet{Li2019EntityRelationEA} converted the relation extraction task to a question-answering task and used the reading comprehension approach for zero-shot relation extraction. In this approach, entity 1 and the relation are used as questions, and entity 2 is the answer to the question. If entity 2 does not exist, the answer is {\em NIL}. \citet{Levy2017ZeroShotRE} used the BiDAF model \citep{seo2016bidirectional} with an additional {\em NIL} node in the output layer for this task on the WikiReading \citep{hewlett2016wikireading} dataset with additional negative samples. They used a set of relations during training and another set of relations during testing. \citet{Li2019EntityRelationEA} used templates to create the question using entity 1 and the relation. They modified the machine-reading comprehension models to a sequence tagging model so that they can find multiple answers to a question. Although they did not experiment with the zero-shot scenario, this approach can be used for zero-shot relation extraction too. FewRel 2.0 \citep{gao2019fewrel} is a dataset for few-shot relation extraction. In few-shot relation extraction, training and test relations are different just like zero-shot extraction. But during testing, a few examples of the test relations are provided to the model for better prediction. 

\subsection{Joint Extraction Approaches}

All the previously mentioned works on relation extraction assume that entities are already identified by a named entity recognition system. They classify the relation between two given entities at the sentence level or the bag-of-sentences level. These models depend on an external named entity recognition system to identify the entities in a text. Recently, some researchers \citep{katiyar2016investigating,miwa2016end,bekoulis2018joint,dat2019end} tried to remove this dependency. They tried to bring the entity recognition and relation identification tasks closer by sharing their parameters and optimizing them together. They first identify all the entities in a sentence and then find the relation among all the pairs of identified entities. Although they identify the entities and relations in the same network, they still identify the entities first and then determine the relation among all possible pairs in the same network. So these models miss the interaction among the relation tuples present in a sentence. These approaches resemble the pipeline approach to some extent.

\citet{zheng2017joint} first proposed a truly joint extraction model for this task. They used a sequence tagging scheme to jointly extract the entities and relations. They created a set of tags derived from the Cartesian product of entity tags and relation tags. These new tags can encode the entity information and relation information together. But this strategy does not work when entities are shared among multiple tuples, as only one tag can be assigned to a token. \citet{zeng2018copyre} proposed an encoder-decoder model with a copy mechanism to extract relation tuples with overlapping entities. Their model has a copy network to copy the last token of two entities from the source sentence and a classification network to classify the relation between copied tokens. Their model cannot extract the full entity names of the tuples. Their best performing model uses a separate decoder to extract each tuple. During training, they need to fix the maximum number of decoders and during inference, their model can only extract up to that fixed number of tuples. Also, due to the use of separate decoders for each tuple, their model misses the interaction among the tuples. 

\citet{takanobu2019hrlre} proposed a hierarchical reinforcement learning-based (RL) deep neural model for joint entity and relation extraction. A high-level RL is used to identify the relation based on the relation-specific tokens in the sentences. After a relation is identified, a low-level RL is used to extract the two entities associated with the relation using a sequence labeling approach. This process is repeated multiple times to extract all the relation tuples present in the sentences. A special {\em None} relation is used to identify no relation situation in the sentences. Entities extracted associated with the {\em None} relations are ignored. \citet{fu2019graphrel} used a graph convolutional network (GCN) where they treated each token in a sentence as a node in a graph and edges were considered as relations. \citet{trisedya2019neural} used an N-gram attention mechanism with an encoder-decoder model for the completion of knowledge bases using distantly supervised data. \citet{Chen2019MrMepJE} used the encoder-decoder framework for this task where they used a CNN-based multi-label classifier to find all the relations first, then used multi-head attention \citep{vaswani2017attention} to extract the entities corresponding to each relation. 

\citet{Zeng2020CopyMTLCM} is an improved version of CopyR \citep{zeng2018copyre} model where they used a sequence tagging approach to extract multi-token entities. \citet{Bowen2020JointEO} decomposed the joint extraction task into two sub-tasks: (i) head entity extraction (ii) tail entity and relation extraction. They used a sequence tagging approach to solve these two sub-tasks. Similarly, \citet{Wei2020ANC} proposed a sequence tagging approach for this task. They first identified the head entities and then for each head entity and each relation, they identified the tail entities using a sequence tagging approach. They used pre-trained BERT \citep{devlin2019bert} in their model to improve the performance.

\subsection{Datasets}

Here, we give a brief description of the available datasets in the area of relation extraction. We describe the datasets that are used in our experiments in detail in the individual chapters of this thesis. \citet{hendrickx2010semeval} proposed a shared task on relation extraction in SemEval 2010 and released a dataset with 8,000 training sentences and 2,717 test instances across nine relations including {\em None}. The relations in this dataset are not taken from any knowledge base. They represent the relationship between two nominals in the sentences. Examples of such relations are {\em Cause-Effect, Component-Whole,} etc. ACE04 \citep{doddington2004automatic}, CoNLL04 \citep{roth2004linear}, and GDS \citep{jat2018attention} are three other datasets with $7$, $5$, and $4$ valid relations respectively. These datasets contain very few relations and few training samples which may not be suitable for building large-scale models. 

\citet{mintz2009distant} first proposed the idea of distant supervision or weak supervision to create a large text-tuple parallel training dataset for relation extraction. They mapped Freebase \citep{bollacker2008freebase} tuples to Wikipedia articles to obtain the dataset. \citet{riedel2010modeling} and \citet{hoffmann2011knowledge} mapped Freebase tuples to the New York Times (NYT) articles to obtain another dataset. These two datasets are used extensively by researchers for their experiments. FewRel 2.0 \citep{gao2019fewrel} is a few-shot relation extraction dataset. These datasets are created at the sentence level. 

WikiReading \citep{hewlett2016wikireading} and DocRED \citep{yao2019DocRED} are two document-level relation extraction datasets created using Wikipedia articles and Wikidata items. \citet{Quirk2017DistantSF} and \citet{peng2017cross} created two document-level relation extraction datasets for the biomedical domain. In the document-level datasets, if two entities appear in a document together, then that document is considered as a source of the tuples involving these two entities.

\section{Multi-Hop Natural Language Processing}

Multi-hop natural language processing (NLP) refers to processing natural language texts that involve multi-hop reasoning steps, possibly across multiple sentences within a single document or across multiple documents. Most NLP tasks such as NER, relation extraction, part-of-speech (POS) tagging, etc. focus on a single sentence. However, multi-hop NLP tasks have recently received more attention from the research community. Reading comprehension (RC) has become very popular with the advances made in deep neural network research. SQuAD \citep{rajpurkar2016squad} is a popular RC dataset that contains more than 100,000 questions based on Wikipedia articles. These questions are created by crowd workers in such a way that models need to focus on multiple sentences of an article to answer correctly. Multi-RC \citep{Khashabi2018LookingBT} is another multiple-choice reading comprehension dataset, where models need to focus on multiple sentences within the same paragraph to answer the question. SquAD and Multi-RC datasets are about multi-hop reasoning across sentences within a single passage, whereas the WikiHop \citep{welbl2018constructing} dataset is about multi-hop reasoning across multiple documents. WikiHop is created using Wikipedia articles and Wikidata tuples. The head entity of a tuple and the relation together form a question. The tail entity of the tuple is the answer to that question. They map these tuples to Wikipedia articles in such a way that the head entity and tail entity of a tuple do not appear in a single document. Thus, models need to consider more that one document to answer the question. OpenBookQA \citep{Mihaylov2018CanAS} is another RC dataset where questions come from elementary science facts. To answer these questions, models need to focus on multiple scientific facts and commonsense knowledge.

Attention-based neural models \citep{seo2016bidirectional,Wang2017GatedSN,kundu2018amanda,wei2018fast} have proven successful for solving the RC task. These models use attention mechanisms to focus on those parts of a passage which have higher similarity to the question. Recently, models based on contextualized word representations such as BERT \citep{devlin2019bert} also achieved human-level performance on the SQuAD RC task. However, these attention-based models do not perform well when multiple documents from different sources need to be used to find the answer. \citet{Dhingra2017GatedAttentionRF}, \citet{Dhingra2018NeuralMF}, \citet{Shen2017ReasoNetLT}, and \citet{Hu2017ReinforcedMR} proposed state-based reasoning models to solve the multi-hop tasks across multiple documents. These state-based models are closer to the attention-based RC models with an additional `state' representation that is updated iteratively. The `state' representations of a model allow it to focus on different parts of multiple documents during each iteration and then combine the information across multiple documents. \citet{Cao2018QuestionAB}, \citet{Song2018Exploring}, and \citet{cao2019bag} proposed graph-based models to solve this task. These models create an entity graph of the entities present in the documents and learn their representations using a convolutional neural network (CNN) or recurrent neural network (RNN). They use reasoning over this entity graph to find the answer to a given question. \citet{Fang2019HierarchicalGN} proposed a hierarchical graph network for multi-hop QA. They created three layers of graphs in their model: entity graph, sentence graph, and passage graph. Finally, a graph reasoning module was used to find the answer span from the passages. \citet{Tu2019MultihopRC} proposed a heterogeneous document-entity (HDE) graph model for multi-hop QA. Their graph contains different types of heterogeneous nodes such as document nodes, candidate answer nodes, and entity nodes. Edges connecting different types of nodes are treated differently in the graph. They also used graph-based reasoning over their HDE graph to find the correct answer. \citet{kundu2019exploiting} used an explicit path-based reasoning model across multiple documents for the multi-hop QA task. They constructed the chain of documents using common entities between documents where each chain led to a possible candidate answer. Each document chain may contain multiple entity-paths from the start document to the end document in the chain. They used an attention network to assign a score to each such entity path and finally assigned a score to each document chain. These scores were used to rank the candidate answers. This path-based reasoning can provide interpretable explanations for the multi-hop QA task. Similar to \citet{kundu2019exploiting}, \citet{Tang2020MultihopRC} proposed a path-based reasoning approach for this task. They created an entity graph using the entities present in documents and used graph convolutional network over this graph and the entity paths to find the answer.

%
\chapter{Syntax-Focused Multi-Factor Attention for Relation Extraction}
\label{chapt:mfa4re}

Relation extraction is the task of determining the relation between two entities in a sentence. Distantly-supervised models are popular for this task. However, we found that sentences collected using distant supervision can be very long, and two entities can be located far from each other in a sentence. The pieces of evidence supporting the presence of a relation between two entities may not be very direct, since the entities may be connected via some indirect links such as a third entity or via co-reference. Relation extraction in such scenarios becomes more challenging as we need to capture the long-distance interactions among the entities and other words in the sentence. Also, the words in a sentence do not contribute equally in identifying the relation between the two entities. To address this issue, we propose a novel and effective attention model which incorporates syntactic information of the sentence and a multi-factor attention mechanism. Experiments on the New York Times corpus show that our proposed model outperforms prior state-of-the-art models. Material from this chapter has been published in \citet{nayak2019effective}.

\section{Motivation}
\label{mfa4re-sec:background}

The sentences obtained for relation extraction using distant supervision are generally long, and two entities are often located far from each other in these sentences. \citet{zeng2014relation,zeng2015distant} used convolutional neural networks (CNN) with max-pooling to find the relation between two given entities in distantly supervised data. Though these models have performed reasonably well on distantly supervised data, they sometimes fail to predict the correct relation when sentences are long, and entities are far from each other. CNN models with max-pooling have limitations in understanding the semantic similarity of words with the given entities, and they also fail to capture the long-distance dependencies among the words and entities such as co-reference. Besides, all the words in a sentence may not be equally important in finding the relation, and this issue is more prominent in long sentences. Prior CNN-based models have limitations in identifying the multiple relevant factors to focus on in sentence-level relation extraction.

To address this issue, we propose a {\it novel} multi-factor attention model\footnote{The code and data of this work can be found at \url{https://github.com/nusnlp/MFA4RE}} focusing on the syntactic structure of a sentence for relation extraction. We use a dependency parser to obtain the syntactic structure of a sentence. We use a linear form of attention to measure the semantic similarity of words with the given entities and combine it with the dependency distance of words from the given entities to measure their influence in identifying the relation. Also, single attention may not be able to capture all pieces of evidence for identifying the relation due to normalization of attention scores. Thus we use multi-factor attention in the proposed model. Experiments on the New York Times (NYT) corpus show that the proposed model outperforms prior work in terms of F1 scores on sentence-level relation extraction.

\section{Problem Definition}
\label{mfa4re-sec:prob_def}
Sentence-level relation extraction is defined as follows: Given a sentence $S$ and two entities $\{E_1, E_2\}$ marked in the sentence, find the relation $r(E_1,E_2)$ between these two entities in $S$ from a pre-defined set of relations $R \cup \{\mathit{None}\}$. {\em None} indicates that none of the relations in $R$ holds between the two marked entities in the sentence. The relation between the entities is argument order-specific, i.e., $r(E_1,E_2)$ and $r(E_2,E_1)$ are not the same. Input to the system is a sentence $S$ and two entities $E_1$ and $E_2$, and output is the relation $r(E_1,E_2) \in R \cup \{\mathit{None}\}$. 

\section{Model Description}
\label{mfa4re-sec:model}

We use four types of embedding vectors in our model: (1) word embedding vector $\mathbf{w} \in \mathbb{R}^{d_w}$ (2) entity token indicator embedding vector $\mathbf{z} \in \mathbb{R}^{d_z}$, which indicates if a word belongs to entity $1$, entity $2$, or does not belong to any entity (3) a positional embedding vector $\mathbf{u}^1 \in \mathbb{R}^{d_u}$ which represents the linear distance of a word from the start token of entity $1$ (4) another positional embedding vector $\mathbf{u}^2 \in \mathbb{R}^{d_u}$ which represents the linear distance of a word from the start token of entity $2$.

We use a bi-directional long short-term memory (Bi-LSTM) \citep{hochreiter1997long} layer to capture the interaction among words in a sentence $S=\{w_1, w_2, ....., w_n\}$, where $n$ is the sentence length. The input to this layer is the concatenated vector $\mathbf{x} \in \mathbb{R}^{d_w + d_z}$ of word embedding vector $\mathbf{w}$ and entity token indicator embedding vector $\mathbf{z}$. 
\begin{align}
&\mathbf{x}_t=\mathbf{w}_t ~||~ \mathbf{z}_t \nonumber\\
&\overrightarrow{\mathbf{h}_t} = \overrightarrow{\mathrm{LSTM}}(\mathbf{x}_t, \mathbf{h}_{t-1}) \nonumber\\
&\overleftarrow{\mathbf{h}_t} = \overleftarrow{\mathrm{LSTM}}(\mathbf{x}_t, \mathbf{h}_{t+1}) \nonumber\\
&\mathbf{h}_t = \overrightarrow{\mathbf{h}_t} || \overleftarrow{\mathbf{h}_t}
\end{align}
\noindent $\overrightarrow{\mathbf{h}_t} \in \mathbb{R}^{d_w+d_z}$ and $\overleftarrow{\mathbf{h}_t} \in \mathbb{R}^{d_w+d_z}$ are the output at the $t$th step of the forward LSTM and backward LSTM respectively. We concatenate them (Eq. (4.1)) to obtain the $t$th Bi-LSTM output $\mathbf{h}_t \in \mathbb{R}^{2(d_w+d_z)}$.

\subsection{Global Feature Extraction}
We use a convolutional neural network (CNN) to extract the sentence-level global features for relation extraction. We concatenate the positional embeddings $\mathbf{u}^1$ and $\mathbf{u}^2$ of words with the hidden representation of the Bi-LSTM layer. We use the convolution operation with max-pooling on these concatenated vectors to extract the global feature vector. 
\begin{align}
&\mathbf{q}_t = \mathbf{h}_t \Vert \mathbf{u}_t^1 \Vert \mathbf{u}_t^2 \nonumber\\
&c_i = \mathbf{f}^T (\mathbf{q}_{i} \Vert \mathbf{q}_{i + 1} \Vert .... \Vert \mathbf{q}_{i+k-1}) \\
&c_{max} = \mathrm{max}(c_1,c_2,....,c_{n}) \\
&\mathbf{v}_g = [c_{max}^1, c_{max}^2, ...., c_{max}^{f_g}]
\end{align} 
$\mathbf{q}_t \in \mathbb{R}^{2(d_w+d_z+d_u)}$ is the concatenated vector for the $t$th word. $\mathbf{f}$ is a convolutional filter vector of dimension $2k(d_w+d_z+d_u)$ where $k$ is the filter width. Superscript $T$ represents the transpose operation. The index $i$ moves from $1$ to $n$ and produces a set of scalar values $\{c_1, c_2, .....,c_{n}\}$ (Eq. (4.2)). The max-pooling operation (Eq. (4.3)) chooses the maximum $c_{max}$ from these values as a feature. With $f_g$ number of filters, we get a global feature vector $\mathbf{v}_g \in \mathbb{R}^{f_g}$ (Eq. (4.4)).

\begin{figure*}[ht]
\centering
\includegraphics[scale=0.5]{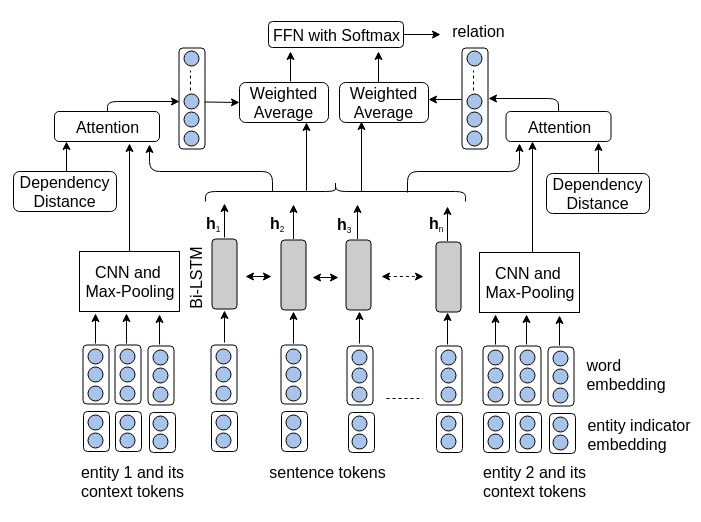}
\caption{Architecture of our attention model with $m=1$, where $m$ is the multi-factor count. We have not shown the CNN-based global feature extraction here. FFN = feed-forward network.}
\label{fig:mfa4re_model}
\end{figure*}

\subsection{Attention Modeling}

Figure~\ref{fig:mfa4re_model} shows the architecture of our attention model. We use a linear form of attention to find the semantically meaningful words in a sentence with respect to the entities which provide the pieces of evidence for the relation between them. Our attention mechanism uses the entities as attention queries and their vector representation is very important for our model. Named entities mostly consist of multiple tokens and many of them may not be present in the training data or their frequency may be low. The nearby words of an entity can give significant information about the entity. Thus we use the tokens of an entity and its nearby tokens to obtain its vector representation. We use the convolution operation with max-pooling in the context of an entity to get its vector representation.
\begin{align}
&c_i = \mathbf{f}^T (\mathbf{x}_{i} \Vert \mathbf{x}_{i + 1} \Vert .... \Vert \mathbf{x}_{i+k-1})\\
&c_{max} = \mathrm{max}(c_b,c_{b+1},....,c_{e})\\
&\mathbf{v}_e = [c_{max}^1, c_{max}^2, ...., c_{max}^{f_e}]
\end{align} 

\noindent $\mathbf{f}$ is a convolutional filter vector of size $k (d_w+d_z)$ where $k$ is the filter width. Superscript $T$ represents the transpose operation. $\mathbf{x}$ is the concatenated vector of word embedding vector ($\mathbf{w}$) and entity token indicator embedding vector ($\mathbf{z}$). $b$ and $e$ are the start and end index of the sequence of words comprising an entity and its neighboring context in the sentence, where $1 \leq b \leq e \leq n$. The index $i$ moves from $b$ to $e$ and produces a set of scalar values $\{c_b, c_{b+1}, .....,c_{e}\}$ (Eq. (4.5)). The max-pooling operation (Eq. (4.6)) chooses the maximum $c_{max}$ from these values as a feature. With $f_e$ number of filters, we get the entity vector $\mathbf{v}_e \in \mathbb{R}^{f_e}$ (Eq. (4.7)). We do this for both entities and get $\mathbf{v}_e^1 \in \mathbb{R}^{f_e}$ and $\mathbf{v}_e^2 \in \mathbb{R}^{f_e}$ as their vector representation. We adopt a simple linear function as follows (Eq. (4.8) and Eq. (4.9)) to measure the semantic similarity of words with the given entities:
\begin{align}
&\mathrm{f_{score}^1}(\mathbf{h}_i, \mathbf{v}_e^1)=\mathbf{h}_i^T \mathbf{W}_a^1 \mathbf{v}_e^1\\
&\mathrm{f_{score}^2}(\mathbf{h}_i, \mathbf{v}_e^2)=\mathbf{h}_i^T \mathbf{W}_a^2 \mathbf{v}_e^2
\end{align}
$\mathbf{h}_i$ is the Bi-LSTM hidden representation of the $i$th word. $\mathbf{W}_a^1 \in \mathbb{R}^{2(d_w+d_z) \times f_e}$ and $\mathbf{W}_a^2 \in \mathbb{R}^{2(d_w+d_z) \times f_e}$ are trainable weight matrices. $\mathrm{f_{score}^1}(\mathbf{h}_i, \mathbf{v}_e^1)$ and $\mathrm{f_{score}^2}(\mathbf{h}_i, \mathbf{v}_e^2)$ represent the semantic similarity score of the $i$th word and the two given entities.

\begin{figure}[t]
\centering
\includegraphics[scale=0.6]{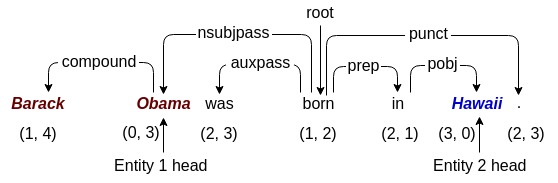}
\caption{An example dependency tree. The two numbers indicate the distance of the word from the head token of the two entities respectively along the dependency tree path.}
\label{fig:mfa4re_dep_dist}
\end{figure}

Not all words in a sentence are equally important in finding the relation between the two entities. The words which are closer to the entities are generally more important. To address this issue, we propose to incorporate the syntactic structure of a sentence in our attention mechanism. The syntactic structure is obtained from the dependency parse tree of the sentence. Words which are closer to the entities in the dependency parse tree are more relevant to finding the relation. In our model, we define the dependency distance to every word from the head token  (last token) of an entity as the number of edges along the dependency path (See Figure~\ref{fig:mfa4re_dep_dist} for an example). We use a distance window size $ws$ and words whose dependency distance is within this window receive attention and the other words are ignored. The details of our attention mechanism follow.
\begin{align}
&d_i^1 = \begin{cases}
    \frac{1}{2^{l_i^1-1}} \mathrm{exp(f_{score}^1}(\mathbf{h}_i, \mathbf{v}_e^1)) & \text{if } l_i^1 \in [1,ws]\\
    \frac{1}{2^{ws}} \mathrm{exp(f_{score}^1}(\mathbf{h}_i, \mathbf{v}_e^1)) & \text{otherwise}
\end{cases} \\
&d_i^2 = \begin{cases}
    \frac{1}{2^{l_i^2-1}} \mathrm{exp(f_{score}^2}(\mathbf{h}_i, \mathbf{v}_e^2)) & \text{if } l_i^2 \in [1,ws]\\
    \frac{1}{2^{ws}} \mathrm{exp(f_{score}^2}(\mathbf{h}_i, \mathbf{v}_e^2)) & \text{otherwise}
\end{cases}\\
&p_i^1 = \frac{d_i^1}{\sum_j{d_j^1}} \\
&p_i^2 = \frac{d_i^2}{\sum_j{d_j^2}}
\end{align}
$d_i^1$ (Eq. (4.10)) and $d_i^2$ (Eq. (4.11)) are un-normalized attention scores and $p_i^1$ (Eq. (4.12)) and $p_i^2$ (Eq. (4.13)) are the normalized attention scores for the $i$th word with respect to entity 1 and entity 2 respectively. $l_i^1$ and $l_i^2$ are the dependency distances of the $i$th word from the two entities. We mask those words whose average dependency distance from the two entities is larger than $ws$. We use the semantic meaning of the words and their dependency distance from the two entities together in our attention mechanism. The attention feature vectors $\mathbf{v}_a^1$ (Eq. (4.14)) and $\mathbf{v}_a^2$ (Eq. (4.15)) with respect to the two entities are determined as follows:
\begin{align}
    &\mathbf{v}_a^1 = \sum_{i=1}^n p_i^1 \mathbf{h}_i \\  &\mathbf{v}_a^2 = \sum_{i=1}^n p_i^2 \mathbf{h}_i 
\end{align}

\subsection{Multi-Factor Attention}

Two entities in a sentence, when located far from each other, can be linked via more than one co-reference chain or more than one important word. Due to the normalization of the attention scores as described above, single attention cannot capture all relevant information needed to find the relation between two entities. Thus we use a multi-factor attention mechanism, where the number of factors is a hyper-parameter, to gather all relevant information for identifying the relation. We replace the attention matrix $\mathbf{W}_a$ with an attention tensor $\mathbf{W}_a^{1:m} \in \mathbb{R}^{m \times 2(d_w+d_z) \times 2f_e}$ where $m$ is the factor count. This gives us $m$ attention vectors with respect to each entity. We concatenate all the feature vectors obtained using these attention vectors to get the multi-attentive feature vector $\mathbf{v}_{ma} \in \mathbb{R}^{4 m (d_w+d_z)}$ (Eq. (4.16)). 
\begin{align}
&\mathbf{v}_{ma} = \mathbf{v}_{a}^{11} \Vert .... \Vert \mathbf{v}_{a}^{1m} \Vert \mathbf{v}_{a}^{21} \Vert .... \Vert \mathbf{v}_{a}^{2m} 
\end{align}

\subsection{Relation Extraction}

We concatenate $\mathbf{v}_g$, $\mathbf{v}_{ma}$, $\mathbf{v}_e^1$, and $\mathbf{v}_e^2$, and this concatenated feature vector is given to a feed-forward layer with softmax activation to predict the normalized probabilities for the relation labels (Eq. (4.17)).
\begin{align}
&\mathbf{r} = \mathrm{softmax}(\mathbf{W}_r (\mathbf{v}_g~||~\mathbf{v}_{ma} ~||~ \mathbf{v}_e^1 ~||~ \mathbf{v}_e^2)+\mathbf{b}_r)
\end{align}
$\mathbf{W}_r \in \mathbb{R}^{(f_g + 2 f_e + 4 m (d_w+d_z)) \times (\vert R \vert+1)}$ is the weight matrix, $\mathbf{b}_r \in \mathbb{R}^{\vert R \vert+1}$ is the bias vector of the feed-forward layer for relation extraction, and $\mathbf{r}$ is the vector of normalized probabilities of relation labels.

\subsection{Loss Function}
We calculate the loss over each mini-batch of size $B$. We use the following negative log-likelihood (Eq. (4.18)) as our objective function for relation extraction:
\begin{equation}
\mathcal{L} = -\frac{1}{B} \sum_{i=1}^{B} \mathrm{log} (p(r_{i} \vert s_i, e_i^1, e_i^2, \theta))
\end{equation}
where $p(r_{i} \vert s_i, e_i^1, e_i^2, \theta)$ is the conditional probability of the true relation $r_i$ when the sentence $s_i$, two entities $e_i^1$ and $e_i^2$, and the model parameters $\theta$ are given.

\section{Experiments}
\label{mfa4re-sec:experiments}

\subsection{Datasets}

We use the New York Times (NYT) corpus \citep{riedel2010modeling} in our experiments. There are two versions of this corpus: (1) The original NYT corpus created by \citet{riedel2010modeling} which has $52$ valid relations and a {\em None} relation. We name this dataset NYT10. The training dataset has $455,412$ instances and $330,776$ of the instances belong to the {\em None} relation and the remaining $124,636$ instances have valid relations. The test dataset has $172,415$ instances and $165,974$ of the instances belong to the {\em None} relation and the remaining $6,441$ instances have valid relations. Both the training and test datasets have been created by aligning Freebase \citep{bollacker2008freebase} tuples to New York Times articles. (2) Another version created by \citet{hoffmann2011knowledge} which has $24$ valid relations and a {\em None} relation. We name this dataset NYT11. The corresponding statistics for NYT11 are given in Table~\ref{tab:dataset}. The training dataset is created by aligning Freebase tuples to NYT articles, but the test dataset is manually annotated.

\begin{table}[ht]
\centering
\begin{tabular}{l|lcc}
\hline
 &  & \multicolumn{1}{l}{NYT10} & \multicolumn{1}{l}{NYT11} \\ \hline
 & \#relations & 53 & 25 \\ \hline
\multirow{5}{*}{Train} & \# instances & 455,412 & 335,843 \\  
 & \#valid relation tuples & 124,636 & 100,671 \\ 
 & \#{\em None} relation tuples & 330,776 & 235,172 \\  
 & avg. sentence length & 41.1 & 37.2 \\  
 & avg. distance between entity pairs & 12.8 & 12.2 \\ \hline
\multirow{5}{*}{Test} & \# instances & 172,415 & 1,450 \\  
 & \#valid relation tuples & 6,441 & 520 \\ 
 & \#{\em None} relation tuples & 165,974 & 930 \\  
 & avg. sentence length & 41.7 & 39.7 \\  
 & avg. distance between entity pairs & 13.1 & 11.0 \\ \hline
\end{tabular}
\caption{Statistics of the NYT10 and NYT11 dataset.}
\label{tab:dataset}
\end{table}

\subsection{Evaluation Metrics}

We use precision, recall, and F1 scores to evaluate the performance of models on relation extraction after removing the {\em None} labels. We use a confidence threshold to decide if the relation of a test instance belongs to the set of relations $R$ or {\em None}. If the network predicts {\em None} for a test instance, then it is considered as {\em None} only. But if the network predicts a relation from the set $R$ and the corresponding softmax score is below the confidence threshold, then the final class is changed to {\em None}. This confidence threshold is the one that achieves the highest F1 score on the validation dataset. We also include the precision-recall curves for all the models.

\subsection{Parameter Settings}

We run word2vec \citep{mikolov2013distributed} on the NYT corpus to obtain the initial word embeddings with dimension $d_w=50$ and update the embeddings during training. We set the dimension of entity token indicator embedding vector $d_z=10$ and positional embedding vector $d_u=5$. The hidden layer dimension of the forward and backward LSTM is $60$, which is the same as the dimension of input word representation vector $\mathbf{x}$. The dimension of Bi-LSTM output is $120$. We use $f_g=f_e=230$ filters of width $k=3$ for feature extraction whenever we apply the convolution operation. The size of the word embeddings, number of convolution filters, and the filter size are taken from the literature. We use dropout in our network with a dropout rate of $0.5$, and in convolutional layers, we use the tanh activation function. We use the sequence of tokens starting from $5$ words before the entity to $5$ words after the entity as its context. We train our models using mini-batches of size $50$ and optimize the network parameters using the Adagrad optimizer \citep{duchi2011adaptive}. We use the dependency parser from spaCy\footnote{\url{https://spacy.io/}} to obtain the dependency distance of the words from the entities and use $ws=5$ as the window size for dependency distance-based attention.

\subsection{Baselines}

We compare our proposed model with the following state-of-the-art models.

\noindent (1) CNN \citep{zeng2014relation}: Words are represented using word embeddings and two positional embeddings. A convolutional neural network (CNN) with max-pooling is applied to extract the sentence-level feature vector. This feature vector is passed to a feed-forward layer with softmax to classify the relation.

\noindent (2) PCNN \citep{zeng2015distant}: Words are represented using word embeddings and two positional embeddings. A convolutional neural network (CNN) is applied to the word representations. Rather than applying a global max-pooling operation on the entire sentence, three max-pooling operations are applied on three segments of the sentence based on the location of the two entities (hence the name Piecewise Convolutional Neural Network (PCNN)). The first max-pooling operation is applied from the beginning of the sentence to the end of the entity appearing first in the sentence. The second max-pooling operation is applied from the beginning of the entity appearing first in the sentence to the end of the entity appearing second in the sentence. The third max-pooling operation is applied from the beginning of the entity appearing second in the sentence to the end of the sentence. Max-pooled features are concatenated and passed to a feed-forward layer with softmax to determine the relation. 

\noindent (3) Entity Attention (EA) \citep{huang2016attention}: This is the combination of a CNN model and an attention model. Words are represented using word embeddings and two positional embeddings. A CNN with max-pooling is used to extract global features. Attention is applied with respect to the two entities separately. The vector representation of every word is concatenated with the word embedding of the last token of the entity. This concatenated representation is passed to a feed-forward layer with tanh activation and then another feed-forward layer to get a scalar attention score for every word. The original word representations are averaged based on the attention scores to get the attentive feature vectors. The CNN-extracted global feature vector and two attentive feature vectors with respect to the two entities are concatenated and passed to a feed-forward layer with softmax to determine the relation.

\noindent (4) BiGRU Word Attention (BGWA) \citep{jat2018attention}: Words are represented using word embeddings and two positional embeddings. They are passed to a bidirectional gated recurrent unit (BiGRU) \citep{cho2014properties} layer. Hidden vectors of the BiGRU layer are passed to a bilinear operator (a combination of two feed-forward layers) to compute a scalar attention score for each word. Hidden vectors of the BiGRU layer are multiplied by their corresponding attention scores. A piece-wise CNN is applied on the weighted hidden vectors to obtain the feature vector. This feature vector is passed to a feed-forward layer with softmax to determine the relation.

\noindent (5) BiLSTM-CNN: This is our own baseline. Words are represented using word embeddings and entity indicator embeddings. They are passed to a bidirectional LSTM. Hidden representations of the LSTMs are concatenated with two positional embeddings. We use CNN and max-pooling on the concatenated representations to extract the feature vector. Also, we use CNN and max-pooling on the word embeddings and entity indicator embeddings of the context words of entities to obtain entity-specific features. These features are concatenated and passed to a feed-forward layer to determine the relation. This model does not have the attention module of our proposed model.

\begin{table*}[ht]
\centering
\begin{tabular}{llll|lll}
\hline \hline
 & \multicolumn{3}{c|}{NYT10} & \multicolumn{3}{c}{NYT11} \\ 
Model & Prec. & Rec. & F1 & \multicolumn{1}{c}{Prec.} & \multicolumn{1}{c}{Rec.} & \multicolumn{1}{c}{F1} \\ \hline
CNN  & 0.413 & 0.591 & 0.486 & 0.444 & 0.625 & 0.519  \\
PCNN & 0.380 & \textbf{0.642} & 0.477 & 0.446 & 0.679 & 0.538$^\dagger$  \\
EA & 0.443 & 0.638 & 0.523$^\dagger$ & 0.419 & 0.677 & 0.517  \\ 
BGWA & 0.364 & 0.632 & 0.462 & 0.417 & \textbf{0.692} & 0.521  \\ \hline
BiLSTM-CNN & 0.490 & 0.507 & 0.498 & 0.473 & 0.606 & 0.531  \\
Our model & \textbf{0.541} & 0.595 & \textbf{0.566}* & \textbf{0.507} & 0.652 & \textbf{0.571}*  \\ \hline \hline
\end{tabular}
\caption{Performance comparison of different models on the two datasets. * denotes a statistically significant improvement over the previous best state-of-the-art model with $p < 0.01$ under the bootstrap paired t-test. $^\dagger$ denotes the previous best state-of-the-art model.}
\label{tab:re}
\end{table*}

\begin{figure}[ht]
\hfill
\subfigure{\includegraphics[width=6cm]{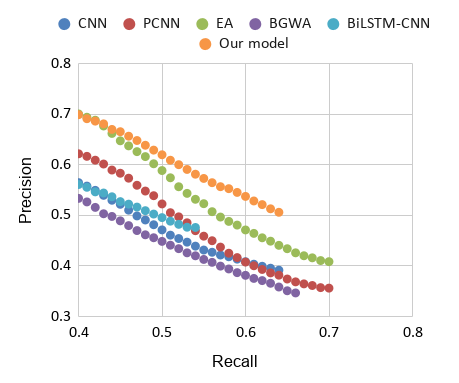}}
\hfill
\subfigure{\includegraphics[width=6cm]{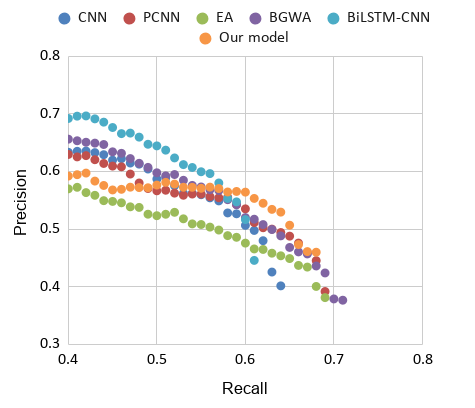}}
\hfill
\caption{Precision-Recall curve for the NYT10 (left) and NYT11 (right) datasets.}
\label{fig:nyt_pr}
\end{figure}

\subsection{Experimental Results}

We present the results of our final model on the relation extraction task on the two datasets in Table \ref{tab:re}. Our model outperforms the previous state-of-the-art models on both datasets in terms of F1 score. On the NYT10 dataset, it achieves $4.3\%$ higher F1 score compared to the previous best state-of-the-art model EA. Similarly, it achieves $3.3\%$ higher F1 score compared to the previous best state-of-the-model PCNN on the NYT11 dataset. Our model improves the precision scores on both datasets with good recall scores. This will help to build a cleaner knowledge base with fewer false positives. We also show the precision-recall curves for the NYT10 and NYT11 datasets in Figure \ref{fig:nyt_pr}. The goal of any relation extraction system is to extract as many relations as possible with minimal false positives. If the recall score becomes very low, the coverage of the KB will be poor. On the NYT10 dataset (left one in Figure \ref{fig:nyt_pr}), we observe that when the recall score is above $0.4$, our model achieves higher precision than all the competing models. On the NYT11 dataset (right one in Figure \ref{fig:nyt_pr}), when recall score is above $0.6$, our model achieves higher precision than the competing models. Achieving higher precision with high recall score helps to build a cleaner KB with good coverage.

\section{Analysis and Discussion}
\label{mfa4re-sec:analysis}

\subsection{Varying the Number of Factors}

We investigate the effects of the multi-factor count $(m)$ in our final model on the test datasets in Table \ref{tab:mfa}. We observe that for the NYT10 dataset, $m=\{1, 2, 3\}$ gives good performance with $m=1$ achieving the highest F1 score. On the NYT11 dataset, $m=4$ gives the best performance. These experiments show that the number of factors giving the best performance may vary depending on the underlying data distribution.

\begin{table}[ht]
\centering
\begin{tabular}{llll|lll}
\hline \hline
 & \multicolumn{3}{c|}{NYT10} & \multicolumn{3}{c}{NYT11} \\ 
$m$ & Prec. & Rec. & F1 & \multicolumn{1}{c}{Prec.} & \multicolumn{1}{c}{Rec.} & \multicolumn{1}{c}{F1} \\ \hline
${1}$ & 0.541 & 0.595 & \textbf{0.566} & 0.495 & 0.621 & 0.551  \\
${2}$ & 0.521 & 0.597 & 0.556 & 0.482 & 0.656 & 0.555  \\
${3}$ & 0.490 & 0.617 & 0.547 & 0.509 & 0.633 & 0.564  \\ 
${4}$ & 0.449 & 0.623 & 0.522 & 0.507 & 0.652 & \textbf{0.571}  \\
${5}$ & 0.467 & 0.609 & 0.529 & 0.488 & 0.677 & 0.567  \\\hline \hline
\end{tabular}
\caption{Performance comparison of our model with different values of $m$ on the two datasets. $m$ refers to the multi-factor count.}
\label{tab:mfa}
\end{table}

\subsection{Effectiveness of Model Components}

We include the ablation results on the NYT11 dataset in Table \ref{tab:nyt11_ablation}. When we add multi-factor attention to the baseline BiLSTM-CNN model without the dependency distance-based weight factor in the attention mechanism, we get $0.8\%$ F1 score improvement (A2$-$A1). Adding the dependency weight factor with a window size of $5$ improves the F1 score by $3.2\%$ (A3$-$A2). Increasing the window size to $10$ reduces the F1 score marginally (A3$-$A4). Replacing the attention normalizing function of Eq. (4.12) and Eq. (4.13) with softmax function in the final model also reduces the F1 score marginally (A3$-$A5). In our final model, we concatenate the features extracted by each attention layer. Rather than concatenating them, we can apply max-pooling operation across the multiple attention scores to compute the final attention scores. These max-pooled attention scores are used to obtain the weighted average vector of Bi-LSTM hidden vectors. This affects the model performance negatively and F1 score of the model decreases by $3.0\%$ (A3$-$A6).

\begin{table}
\centering
\begin{tabular}{lccc}
\hline
 & Prec. & Rec. & F1 \\ \hline
(A1) BiLSTM-CNN & 0.473 & 0.606 & 0.531 \\
(A2) \quad + Standard attention & 0.466 & 0.638 & 0.539 \\
(A3) \quad \quad + Dep. dist. weight factor$_{ws=5}$ & 0.507 & 0.652 & \textbf{0.571} \\
(A4) \quad \quad + Dep. dist. weight factor$_{ws=10}$ & 0.510 & 0.640 & 0.568 \\ \hline
(A5) Our model with softmax normalization & 0.490 & 0.658 & 0.562 \\ 
(A6) Our model with max-pool aggregation & 0.492 & 0.600 & 0.541 \\ \hline
\end{tabular}
\caption{Effectiveness of model components ($m=4$) on the NYT11 dataset. $m$ refers to the multi-factor count. $ws$ is the dependency window size used in Eq. (4.10) and Eq. (4.11).}
\label{tab:nyt11_ablation}
\end{table}

\begin{figure}
\hfill
\subfigure{\includegraphics[width=6cm]{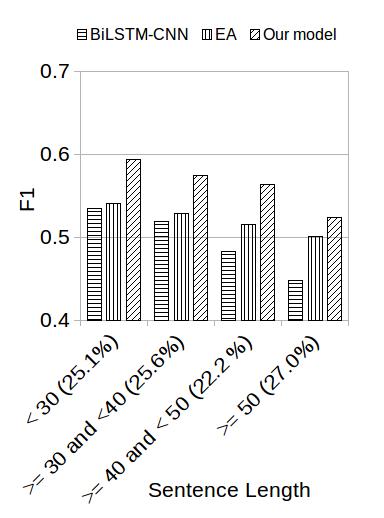}}
\hfill
\subfigure{\includegraphics[width=6cm]{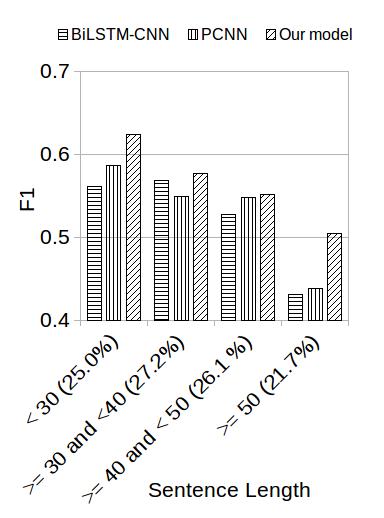}}
\hfill
\caption{Performance comparison across different sentence lengths on the NYT10 (left) and NYT11 (right) datasets.}
\label{fig:nyt_len}
\end{figure}

\begin{figure}
\hfill
\subfigure{\includegraphics[width=6cm]{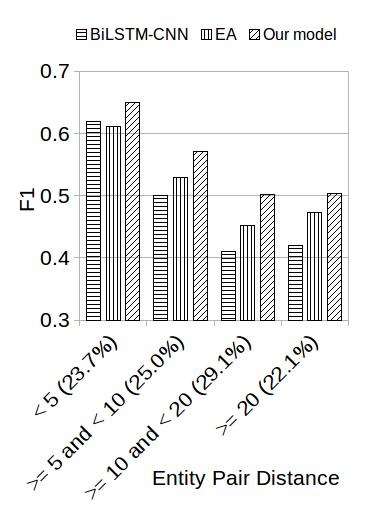}}
\hfill
\subfigure{\includegraphics[width=6cm]{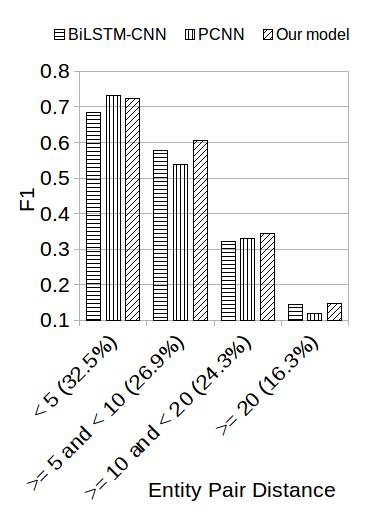}}
\hfill
\caption{Performance comparison across different distances between entities on the NYT10 (left) and NYT11 (right) datasets.}
\label{fig:nyt_dist}
\end{figure}

\subsection{Error Analysis}

In relation extraction, finding relations in long sentences where the two entities are far apart becomes more difficult, and models start to make mistakes in these scenarios. To understand the behavior of the models in such scenarios, we analyze their performance with varying sentence lengths and with varying distances between the two entities in the two datasets in Figure \ref{fig:nyt_len} and \ref{fig:nyt_dist} respectively. We compare the performance of our proposed model against the previous state-of-the-art baselines in the corresponding dataset. We also include our own BiLSTM-CNN model for comparison to show the effectiveness of our attention module. From Figure \ref{fig:nyt_len}, we see that models start making more mistakes when the sentence length increases in both datasets. The F1 scores of the models drop significantly with increasing distance between two entities. In comparison, our model performs better than the other two models we compare to across all categories of sentences, based on either length or entity pair distance on the NYT10 dataset. On the NYT11 dataset, our model either performs better than or similar to the two other models in similar categories of sentences.

\section{Summary}
\label{mfa4re-sec:summary}

In this chapter, we describe a multi-factor attention model utilizing the syntactic structure of sentences for relation extraction. The syntactic structure component of our model helps to identify important words in a sentence and the multi-factor component helps to gather different pieces of evidence present in a sentence. Together, these two components improve the performance of our model on this task, and our model outperforms previous state-of-the-art models when evaluated on the New York Times (NYT) corpus, achieving significantly higher F1 scores.

%
\chapter{Encoder-Decoder Models for Joint Entity and Relation Extraction}
\label{chapt:ed4jere}

The pipeline relation extraction approaches have an external dependency on an effective named entity recognition system. To remove that dependency, we explore joint entity and relation extraction in this work. A sentence may have multiple relation tuples and these tuples may share one or both entities among them. Extracting such relation tuples is a difficult task and sharing of entities or overlapping entities among the tuples makes it more challenging. Most prior work that adopted the pipeline approach cannot capture the interaction among the relation tuples in a sentence in an effective way. In this work, we propose two approaches to use encoder-decoder architecture for jointly extracting entities and relations, where relation tuples are generated in a sequence. In the first approach, we propose a representation scheme for relation tuples that enables the decoder to generate one word at a time, like machine translation models and still finds all the tuples present in a sentence with full entity names of different length and with overlapping entities. Next, we propose a pointer network-based decoding approach where an entire tuple is generated at every time step of the decoding process. Experiments on the publicly available New York Times corpus show that our proposed approaches outperform previous work and achieve significantly higher F1 scores. Material from this chapter has been published in \citet{nayak2019ptrnetdecoding}.

\section{Motivation}
\label{ed4jere-sec:background}
\noindent Distantly-supervised information extraction systems extract relation tuples with a set of pre-defined relations from text. Traditionally, researchers \citep{mintz2009distant,riedel2010modeling,hoffmann2011knowledge,zeng2014relation,zeng2015distant,huang2016attention,ren2017cotype,jat2018attention,vashishth2018reside} use pipeline approaches where a named entity recognition (NER) system is used to identify the entities in a sentence and then a classifier is used to find the relation (or no relation) between them. However, due to the complete separation of entity detection and relation classification, these models miss the interaction between multiple relation tuples present in a sentence. In this approach, models find the relation only between two given entities in a sentence and do not explicitly consider the other entities and relations present in the same sentence. Thus these models miss the interaction among multiple tuples while classifying the relation between two entities.

Recently, several neural network-based models \citep{katiyar2016investigating,miwa2016end} were proposed to jointly extract entities and relations from a sentence. These models used a parameter-sharing mechanism to extract the entities and relations in the same network. But they still find the relations after identifying all the entities and do not fully capture the interaction among multiple tuples. \citet{zheng2017joint} proposed a joint extraction model based on neural sequence tagging scheme. But their model could not extract tuples with overlapping entities in a sentence as it could not assign more than one tag to a word. \citet{zeng2018copyre} proposed a neural encoder-decoder model for extracting relation tuples with overlapping entities. However, they used a copy mechanism to copy only the last token of the entities, thus this model could not extract the full entity names. Also, their best performing model used a separate decoder to extract each tuple which limited the power of their model. This model was trained with a fixed number of decoders and could not extract tuples beyond that number during inference. Encoder-decoder models are powerful models and they are successful in many NLP tasks such as machine translation, sentence generation from structured data, and open information extraction.

\begin{table}[ht]
\centering
\resizebox{\columnwidth}{!}{
\begin{tabular}{l|l}
\hline
Sentence & Berlin is the capital of Germany . \\ \hline

Tuples  & \begin{tabular}[c]{@{}l@{}}\textless{}Germany, Berlin, capital\textgreater\\ \textless{}Germany, Berlin, contains\textgreater\\ \textless{}Berlin, Germany, country\textgreater{}\end{tabular} \\ \hline

Target: word-based decoding  & \multicolumn{1}{l}{\begin{tabular}[c]{@{}l@{}}Germany ; Berlin ; capital $\vert$ \\Germany ; Berlin ; contains $\vert$ \\Berlin ; Germany ; country \end{tabular}}  \\ \hline

Target: pointer network-based decoding & \multicolumn{1}{l}{\begin{tabular}[c]{@{}l@{}}$<$5 5 0 0 capital$>$ \\$<$5 5 0 0 contains$>$ \\$<$0 0 5 5 country$>$ \end{tabular}}                                                        \\ \hline
\end{tabular}
}
\caption{Relation tuple representation for encoder-decoder models.}
\label{tab:mt_schemes}
\end{table}

In this work, we explore how encoder-decoder models can be used effectively for extracting relation tuples from sentences. There are three major challenges in this task: (i) The model should be able to extract entities and relations together. (ii) It should be able to extract multiple tuples with overlapping entities. (iii) It should be able to extract exactly two entities of a tuple with their full names. To address these challenges, we propose two {\it novel} approaches using encoder-decoder architecture\footnote{The code and data of this work can be found at \url{https://github.com/nusnlp/PtrNetDecoding4JERE}}. We first propose a new representation scheme for relation tuples (Table \ref{tab:mt_schemes}) such that it can represent multiple tuples with overlapping entities and different lengths of entities in a simple way. We employ an encoder-decoder model where the decoder extracts one word at a time like machine translation models. At the end of sequence generation, due to the unique representation of the tuples, we can extract the tuples from the sequence of words. Although this model performs quite well, generating one word at a time is somewhat unnatural for this task. Each tuple has exactly two entities and one relation, and each entity appears as a continuous text span in a sentence. The most effective way to identify them is to find their start and end location in the sentence. Each relation tuple can then be represented using five items: start and end location of the two entities and the relation between them (see Table \ref{tab:mt_schemes}). Keeping this in mind, we propose a pointer network-based decoding framework. This decoder consists of two pointer networks which find the start and end location of the two entities in a sentence, and a classification network which identifies the relation between them. At every time step of the decoding, this decoder extracts an entire relation tuple, not just a word. Experiments on the New York Times (NYT) datasets show that our approaches work effectively for this task and achieve state-of-the-art performance. To summarize, the contributions of this work are as follows:

\noindent(1) We propose a new representation scheme for relation tuples such that an encoder-decoder model, which extracts one word at each time step, can still find multiple tuples with overlapping entities and tuples with multi-token entities from sentences. We also propose a masking-based copy mechanism to extract the entities from the source sentence only.\\
\noindent(2) We propose a modification in the decoding framework with pointer networks to make the encoder-decoder model more suitable for this task. At every time step, this decoder extracts an entire relation tuple, not just a word. This new decoding framework helps in speeding up the training process and uses less resources (GPU memory). This will be an important factor when we move from sentence-level tuple extraction to document-level extraction. \\
\noindent(3) Experiments on the NYT datasets show that our approaches outperform all the previous state-of-the-art models significantly and set a new benchmark on these datasets.

\section{Problem Definition}

A relation tuple consists of two entities and a relation. Such tuples can be found in sentences where an entity is a text span in a sentence and a relation comes from a pre-defined set $R$. These tuples may share one or both entities among them. Based on this, we divide the sentences into three classes: (i) {\em No Entity Overlap (NEO)}: A sentence in this class has one or more tuples, but they do not share any entities. (ii) {\em Entity Pair Overlap (EPO)}: A sentence in this class has more than one tuple, and at least two tuples share both the entities in the same or reverse order. (iii) {\em Single Entity Overlap (SEO)}: A sentence in this class has more than one tuple and at least two tuples share exactly one entity. It should be noted that a sentence can belong to both EPO and SEO classes. Our task is to extract all relation tuples present in a sentence.

\begin{table*}[ht]
\small
\centering
\begin{tabular}{l|l|l}
\hline
\multicolumn{1}{c|}{Class} & \multicolumn{1}{c|}{Sentence}                                                                                                                          & \multicolumn{1}{c}{Tuples}                                                                                                                                                                                 \\ \hline
NEO                                  & \begin{tabular}[c]{@{}l@{}}The original Joy of \\Cooking was \\published in 1931 \\by Irma Rombauer,\\ a St. Louis housewife.\end{tabular} & \textless{}Irma Rombauer, St. Louis, place\_lived\textgreater{}                                                                                                                                      \\ \hline
EPO                                     & \begin{tabular}[c]{@{}l@{}}Berlin is the capital\\ of Germany.\end{tabular}                                                                                                                     & \begin{tabular}[c]{@{}l@{}}\textless{}Germany, Berlin, capital\textgreater\\ \textless{}Germany, Berlin, contains\textgreater\\ \textless{}Berlin, Germany, country\textgreater{}\end{tabular} \\ \hline
SEO                                     & \begin{tabular}[c]{@{}l@{}}Dr. C. V. Raman who \\was born in Chennai\\ worked mostly in \\Kolkata.\end{tabular}                                      & \begin{tabular}[c]{@{}l@{}} \textless{}Dr. C. V. Raman, Chennai, birth\_place\textgreater\\  \textless{}Dr. C. V. Raman, Kolkata, place\_lived\textgreater{}\end{tabular}                         \\ \hline
\end{tabular}
\caption{Examples of different classes of overlapping relation tuples.}
\label{tab:ov_classes}
\end{table*}

\section{Model Description}
\label{ed4jere-sec:model}

In this task, the input to the system is a sequence of words, and the output is a set of relation tuples. In our first approach, we represent each tuple as {\em entity1 ; entity2 ; relation}. We use `;' as a separator token to separate the tuple components. Multiple tuples are separated using the `$\vert$' token. We have included one example of such representation in Table \ref{tab:mt_schemes}. Multiple relation tuples with overlapping entities and different lengths of entities can be represented in a simple way using these special tokens (; and $\vert$). During inference, after the end of sequence generation, relation tuples can be extracted easily using these special tokens. Due to this uniform representation scheme, where entity tokens, relation tokens, and special tokens are treated similarly, we use a shared vocabulary between the encoder and decoder which includes all of these tokens. The input sentence contains clue words for every relation which can help generate the relation tokens. We use two special tokens so that the model can distinguish between the beginning of a relation tuple and the beginning of a tuple component. To extract the relation tuples from a sentence using the encoder-decoder model, the model has to generate the entity tokens, find the clue words for the relations, map them to the relation tokens, and generate the special tokens at appropriate time. Our experiments show that the encoder-decoder models can achieve this quite effectively.

\subsection{Embedding Layer \& Encoder}

We create a single vocabulary $V$ consisting of the source sentence tokens, relation names from relation set $R$, special separator tokens (`;', `$\vert$'), start-of-target-sequence token ({\em SOS}), end-of-target-sequence token ({\em EOS}), and unknown word token ({\em UNK}). Word-level embeddings are formed by two components: (1) pre-trained word vectors (2) character embedding-based feature vectors. We use a word embedding layer $\mathbf{E}_w \in \mathbb{R}^{\vert V \vert \times d_w}$ and a character embedding layer $\mathbf{E}_c \in \mathbb{R}^{\vert A \vert \times d_c}$, where $d_w$ is the dimension of word vectors, $A$ is the character alphabet of input sentence tokens, and $d_c$ is the dimension of character embedding vectors. Following \citet{chiu2016named}, we use a convolutional neural network with max-pooling to extract a feature vector of size $d_f$ for every word. Word embeddings and character embedding-based feature vectors are concatenated ($\Vert$) to obtain the representation of the input tokens.

 A source sentence $\mathbf{S}$ is represented by vectors of its tokens $\mathbf{x}_1, \mathbf{x}_2,....,\mathbf{x}_n$, where $\mathbf{x}_i \in \mathbb{R}^{(d_w+d_f)}$ is the vector representation of the $i$th word and $n$ is the length of $\mathbf{S}$. These vectors $\mathbf{x}_i$ are passed to a bi-directional LSTM \citep{hochreiter1997long} (Bi-LSTM) to obtain the hidden representation $\mathbf{h}_i^E$. We set the hidden dimension of the forward and backward LSTM of the Bi-LSTM to be $d_h/2$ to obtain $\mathbf{h}_i^E \in \mathbb{R}^{d_h}$, where $d_h$ is the hidden dimension of the sequence generator LSTM of the decoder described below.

\subsection{Word-level Decoder \& Copy Mechanism}

A target sequence $\mathbf{T}$ is represented by only word embedding vectors of its tokens $\mathbf{y}_0, \mathbf{y}_1,....,\mathbf{y}_m$ where $\mathbf{y}_i \in \mathbb{R}^{d_w}$ is the embedding vector of the $i$th token and $m$ is the length of the target sequence. $\mathbf{y}_0$ and $\mathbf{y}_m$ represent the embedding vector of the {\em SOS} and {\em EOS} token respectively. The decoder generates one token at a time and stops when {\em EOS} is generated. We use an LSTM as the decoder and at time step $t$, the decoder takes the source sentence encoding, $\mathbf{e}_t \in \mathbb{R}^{d_h}$, and the previous target word embedding, $\mathbf{y}_{t-1}$, as the input and generates the hidden representation of the current token ($\mathbf{h}_t^D \in \mathbb{R}^{d_h}$). The sentence encoding vector $\mathbf{e}_t$ can be obtained using attention mechanism. $\mathbf{h}_t^D$ is projected to the vocabulary $V$ using a linear layer with weight matrix $\mathbf{W}_v \in \mathbb{R}^{\vert V \vert \times d_h}$ and bias vector $\mathbf{b}_v \in \mathbb{R}^{\vert V \vert}$ (projection layer). $\Vert$ refers to the concatenation operation.
\begin{align}
&\mathbf{h}_t^D = \mathrm{LSTM}(\mathbf{e}_t \Vert \mathbf{y}_{t-1}, \mathbf{h}_{t-1}^D) \nonumber\\
&\hat{\mathbf{o}}_t = \mathbf{W}_v \mathbf{h}_t^D + \mathbf{b}_v \nonumber\\
&\mathbf{o}_t = \mathrm{softmax}(\hat{\mathbf{o}}_t)
\end{align}
$\mathbf{o}_t$ represents the normalized scores of all the words in the embedding vocabulary at time step $t$. $\mathbf{h}_{t-1}^D$ is the previous hidden state of the LSTM.

The projection layer of the decoder maps the decoder output to the entire vocabulary. During training, we use the gold label target tokens directly. However, during inference, the decoder may predict a token from the vocabulary which is not present in the current sentence or the set of relations or the special tokens. To prevent this, we use a masking technique while applying the softmax operation at the projection layer. We mask (exclude) all words of the vocabulary except the current source sentence tokens, relation tokens, separator tokens (`;', `$\vert$'), {\em UNK}, and {\em EOS} tokens in the softmax operation. To mask (exclude) some word from softmax, we set the corresponding value in $\hat{\mathbf{o}}_t$ at $-\infty$ and the corresponding softmax score will be zero. This ensures the copying of entities from the source sentence only. We include the {\em UNK} token in the softmax operation to make sure that the model generates new entities during inference. If the decoder predicts an {\em UNK} token, we replace it with the corresponding source word which has the highest attention score. During inference, after decoding is finished, we extract all tuples based on the special tokens, remove duplicate tuples and tuples in which both entities are the same or tuples where the relation token is not from the relation set. This model is referred to as {\em WordDecoding} (WDec) henceforth.

\subsection{Pointer Network-Based Decoder}

In the second approach, we identify the entities in the sentence using their start and end locations. We remove the special tokens and relation names from the word vocabulary and word embeddings are used only at the encoder side along with character embeddings. We use an additional relation embedding matrix $\mathbf{E}_r \in \mathbb{R}^{\vert R \vert \times d_r}$ at the decoder side of our model, where  $R$ is the set of relations and $d_r$ is the dimension of relation vectors. The relation set $R$ includes a special relation token {\em EOS} which indicates the end of the sequence. Relation tuples are represented as a sequence $T=y_0, y_1,....,y_m$, where $y_t$ is a tuple consisting of four indexes in the source sentence indicating the start and end location of the two entities and a relation between them (see Table \ref{tab:mt_schemes}). $y_0$ is a dummy tuple that represents the start tuple of the sequence and $y_m$ functions as the end tuple of the sequence which has {\em EOS} as the relation (entities are ignored for this tuple). The decoder consists of an LSTM with hidden dimension $d_h$ to generate the sequence of tuples, two pointer networks to find the two entities, and a classification network to find the relation of a tuple. At time step $t$, the decoder takes the source sentence encoding, $\mathbf{e}_t \in \mathbb{R}^{d_h}$, and the representation of all previously generated tuples $\mathbf{y}_{prev}$ (Eq. (5.2)) as the input and generates the hidden representation of the current tuple, $\mathbf{h}_t^D \in \mathbb{R}^{d_h}$ (Eq. (5.3)). The sentence encoding vector $\mathbf{e}_t$ is obtained using an attention mechanism as explained later. Relation tuples are a set and to prevent the decoder from generating the same tuple again, we pass the information about all previously generated tuples at each time step of decoding. $\mathbf{y}_j$ is the vector representation of the tuple predicted at time step $j < t$ and we use the zero vector ($\mathbf{y}_0=\overrightarrow{0}$) to represent the dummy tuple $y_0$. $\mathbf{h}_{t-1}^D$ is the hidden state of the LSTM at time step $t-1$. $\Vert$ refers to the concatenation operation.
\begin{align}
&\mathbf{y}_{prev} = \sum_{j=0}^{t-1} \mathbf{y}_j \\
&\mathbf{h}_t^D = \mathrm{LSTM}(\mathbf{e}_t \Vert \mathbf{y}_{prev}, \mathbf{h}_{t-1}^D)
\end{align}

\begin{figure}[t]
\centering
\includegraphics[scale=0.5]{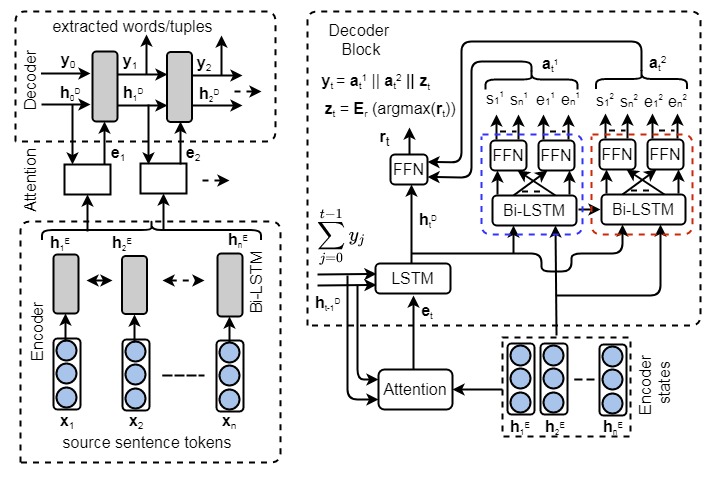}
\caption{The architecture of an encoder-decoder model (left) and a pointer network-based decoder block (right).}
\label{fig:ptrnet_model}
\end{figure}

\subsubsection{Relation Tuple Extraction}

After obtaining the hidden representation of the current tuple $\mathbf{h}_t^D$, we first find the start and end pointers of the two entities in the source sentence. We concatenate the vector $\mathbf{h}_t^D$ with the hidden vectors $\mathbf{h}_i^E$ of the encoder and pass them to a Bi-LSTM layer with hidden dimension $d_p$ for forward and backward LSTM. The hidden vectors of this Bi-LSTM layer $\mathbf{h}_i^k \in \mathbb{R}^{2d_p}$ are passed to two feed-forward networks (FFN) with softmax (Eq. (5.4) to Eq. (5.7)) to convert each hidden vector into two scalar values between $0$ and $1$. The softmax operation is applied across all the words in the input sentence. These two scalar values represent the probability of the corresponding source sentence token to be the start and end location of the first entity. This Bi-LSTM layer with the two feed-forward layers (the blue block in Figure \ref{fig:ptrnet_model}) is the first pointer network which identifies the first entity of the current relation tuple.
\begin{align}
&\hat{s}_i^1 = \mathbf{W}_s^1 \mathbf{h}_i^k + {b}_s^1\\
&\mathbf{s}^1 = \mathrm{softmax}(\hat{\mathbf{s}}^1)\\
&\hat{e}_i^1 = \mathbf{W}_e^1 \mathbf{h}_i^k + {b}_e^1\\
&\mathbf{e}^1 = \mathrm{softmax}(\hat{\mathbf{e}}^1)
\end{align}

\noindent where $\mathbf{W}_s^1 \in \mathbb{R}^{1 \times 2d_p}$, $\mathbf{W}_e^1 \in \mathbb{R}^{1 \times 2d_p}$, ${b}_s^1$, and ${b}_e^1$ are the weights and bias parameters of the feed-forward layers. ${s}_i^1$, ${e}_i^1$ represent the normalized probabilities of the $i$th source word being the start and end token of the first entity of the predicted tuple. We use another pointer network (the red block in Figure \ref{fig:ptrnet_model}) to extract the second entity of the tuple. We concatenate the hidden vectors $\mathbf{h}_i^k$ with $\mathbf{h}_t^D$ and $\mathbf{h}_i^E$ and pass them to the second pointer network to obtain ${s}_i^2$ and ${e}_i^2$, which represent the normalized probabilities of the $i$th source word being the start and end of the second entity. These normalized probabilities are used to find the vector representation of the two entities, $\mathbf{a}_t^1$ (Eq. (5.8)) and $\mathbf{a}_t^2$ (Eq. (5.9)).
\begin{align}
&\mathbf{a}_t^1 = \sum_{i=1}^n {s}_i^1 \mathbf{h}_i^k \Vert \sum_{i=1}^n {e}_i^1 \mathbf{h}_i^k \\ 
&\mathbf{a}_t^2 = \sum_{i=1}^n {s}_i^2 \mathbf{h}_i^l \Vert \sum_{i=1}^n {e}_i^2 \mathbf{h}_i^l
\end{align}
We concatenate the entity vector representations $\mathbf{a}_t^1$ and $\mathbf{a}_t^2$ with $\mathbf{h}_t^D$ and pass it to a feed-forward network (FFN) with softmax to find the relation (Eq. (5.10)). This feed-forward layer has a weight matrix $\mathbf{W}_r \in \mathbb{R}^{\vert R \vert \times (8d_p + d_h)}$ and a bias vector $\mathbf{b}_r \in \mathbb{R}^{\vert R \vert}$.
\begin{align}
&\mathbf{r}_t = \mathrm{softmax}(\mathbf{W}_r (\mathbf{a}_t^1 \Vert \mathbf{a}_t^2 \Vert  \mathbf{h}_t^D) + \mathbf{b}_r)\\
&\mathbf{z}_t=\mathbf{E}_r(\mathrm{argmax}(\mathbf{r}_t))\\
&\mathbf{y}_t=\mathbf{a}_t^1 \Vert \mathbf{a}_t^2 \Vert \mathbf{z}_t
\end{align}
$\mathbf{r}_t$ represents the normalized probabilities of the relation at time step $t$. The relation embedding vector $\mathbf{z}_t$ is obtained using $\mathrm{argmax}$ of $\mathbf{r}_t$ and $\mathbf{E}_r$ (Eq. (5.11)). $\mathbf{y}_t \in \mathbb{R}^{(8d_p + d_r)}$ (Eq. (5.12)) is the vector representation of the tuple predicted at time step $t$. During training, we pass the embedding vector of the gold label relation in place of the predicted relation. So the $\mathrm{argmax}$ function does not affect the back-propagation during training. The decoder stops the sequence generation process when the predicted relation is {\em EOS}. This is the classification network of the decoder.

During inference, we select the start and end location of the two entities such that the product of the four pointer probabilities is maximized keeping the constraints that the two entities do not overlap with each other and $1 \leq b \leq e \leq n$ where $b$ and $e$  are the start and end location of the corresponding entities. We first choose the start and end location of entity 1 based on the maximum product of the corresponding start and end pointer probabilities. Then we find entity 2 in a similar way excluding the span of entity 1 to avoid overlap. The same procedure is repeated but this time we first find entity 2 followed by entity 1. We choose that pair of entities which gives the higher product of four pointer probabilities between these two choices. This model is referred to as {\em PtrNetDecoding} (PNDec) henceforth.

\subsection{Attention Modeling}

We experimented with three different attention mechanisms for our word-level decoding model to obtain the source context vector $\mathbf{e}_t$:

\noindent(1) Avg.: The context vector is obtained by averaging the hidden vectors of the encoder:
\begin{align}
    &\mathbf{e}_t=\frac{1}{n}\sum_{i=1}^n \mathbf{h}_i^E
\end{align}
       
\noindent(2)  N-gram: The context vector is obtained by the N-gram attention mechanism of \citet{trisedya2019neural} with N=3. This attention mechanism can be helpful to identify the multi-token entities.
\begin{align}
    &\textnormal{a}_i^g=(\mathbf{h}_n^{E})^T \mathbf{V}^g \mathbf{w}_i^g \nonumber\\
    &\boldsymbol{\alpha}^g = \mathrm{softmax}(\mathbf{a}^g) \nonumber\\
    &\mathbf{e}_t=[\mathbf{h}_n^E \Vert \sum_{g=1}^N \mathbf{W}^g (\sum_{i=1}^{\vert G^g \vert} \alpha_i^g \mathbf{w}_i^g)]
\end{align}
\noindent Here, $\mathbf{h}_n^E$ is the last hidden state of the encoder, $g \in \{1, 2, 3\}$ refers to the word gram combination, $G^g$ is the sequence of g-gram word representations for the input sentence, $\mathbf{w}_i^g$ is the $i$th g-gram vector (2-gram and 3-gram representations are obtained by average pooling), $\alpha_i^g$ is the normalized attention score for the $i$th g-gram vector, $\mathbf{W} \in \mathbb{R}^{d_h \times d_h}$ and $\mathbf{V} \in \mathbb{R}^{d_h \times d_h}$ are trainable parameters. \\
\noindent(3)  Single: The context vector is obtained by the attention mechanism proposed by \citet{Bahdanau2014NeuralMT}. This attention mechanism gives the best performance with the word-level decoding model.
\begin{align}
    &\mathbf{u}_t^i = \mathbf{W}_{u} \mathbf{h}_i^E \nonumber\\
    &\mathbf{q}_t^i = \mathbf{W}_{q} \mathbf{h}_{t-1}^D  + \mathbf{b}_{q} \nonumber\\
    &\textnormal{a}_t^i = \mathbf{v}_a \tanh(\mathbf{q}_t^i + \mathbf{u}_t^i) \nonumber\\
    &\boldsymbol{\alpha}_t = \mathrm{softmax(\mathbf{a}_t)} \nonumber\\
    &\mathbf{e}_t = \sum_{i=1}^n \alpha_t^i \mathbf{h}_i^E
\end{align}
\noindent where $\mathbf{W}_u \in \mathbb{R}^{d_h \times d_h}$, $\mathbf{W}_q \in \mathbb{R}^{d_h \times d_h}$, and $\mathbf{v}_a \in \mathbb{R}^{d_h}$ are all trainable attention parameters and $\mathbf{b}_q \in \mathbb{R}^{d_h}$ is a bias vector. $\alpha_t^i$ is the normalized attention score of the $i$th source word at the decoding time step $t$.

For our pointer network-based decoding model, we use three variants of the single attention model. First, we use $\mathbf{h}_{t-1}^D$ to calculate $\mathbf{q}_t^i$ in the attention mechanism. Next, we use $\mathbf{y}_{prev}$ to calculate $\mathbf{q}_t^i$, where $\mathbf{W}_q \in \mathbb{R}^{(8d_p + d_r) \times d_h}$. In the final variant, we obtain the attentive context vector by concatenating the two attentive vectors obtained using $\mathbf{h}_{t-1}^D$ and $\mathbf{y}_{prev}$. This gives the best performance with the pointer network-based decoding model. These variants are referred to as $\mathrm{dec_{hid}}$, $\mathrm{tup_{prev}}$, and $\mathrm{combo}$ in Table \ref{tab:ablation}.

\subsection{Loss Function}

We minimize the negative log-likelihood loss of the generated words for word-level decoding ($\mathcal{L}_{word}$) (Eq. (5.16)) and minimize the sum of negative log-likelihood loss of relation classification and the four pointer locations for pointer network-based decoding ($\mathcal{L}_{ptr}$) (Eq. (5.17)).
\begin{align}
&\mathcal{L}_{word} = -\frac{1}{B \times T} \sum_{b=1}^{B} \sum_{t=1}^{T} \text{log} (v_{t}^b)\\
&\mathcal{L}_{ptr} = -\frac{1}{B \times T} \sum_{b=1}^{B} \sum_{t=1}^{T} [ \text{log} (r_{t}^b) + \sum_{c=1}^{2} \text{log} (s_{c,t}^{b} e_{c,t}^{b})]
\end{align}
 \noindent $v_t^b$ is the softmax score of the target word at time step $t$ for the word-level decoding model. $r$, $s$, and $e$ are the softmax score of the corresponding true relation label, true start and end pointer location of an entity. $b$, $t$, and $c$ refer to the $b$th training instance, $t$th time step of decoding, and the two entities of a tuple respectively. $B$ and $T$ are the batch size and maximum time step of the decoder respectively.

\section{Experiments}
\label{ed4jere-sec:experiments}

\subsection{Datasets}

We focus on the task of extracting multiple tuples with overlapping entities from sentences. We choose the New York Times (NYT) corpus for our experiments. This corpus has multiple versions, and we choose the following two versions as their test dataset has a significantly larger number of instances of multiple relation tuples with overlapping entities. (i) The first version is used by \citet{zeng2018copyre} (mentioned as NYT in their paper) and has $24$ relations. We name this version as NYT24. This dataset is derived from the NYT11 dataset of the previous chapter. The test data of NYT11 do not have any sentences with multiple tuples or overlapping tuples. But the training data of NYT11 have such sentences. So they split the training data of NYT11 to create the test data for the joint extraction task. (ii) The second version is used by \citet{takanobu2019hrlre} (mentioned as NYT10 in their paper) and has $29$ relations. We name this version as NYT29. This dataset is derived from the NYT10 dataset of the previous chapter. They remove those relations from the dataset which do not appear in the test dataset. We select 10\% of the original training data and use it as the validation dataset. The remaining 90\% is used for training. We include statistics of the training and test datasets in Table \ref{tab:data_stat}. Both NYT24 and NYT29 datasets do not contain any sentence with zero positive tuples.

\begin{table}[ht]
\centering
\resizebox{0.85\columnwidth}{!}{
\begin{tabular}{lcc|cc}
\hline
       & \multicolumn{2}{c|}{NYT29} & \multicolumn{2}{c}{NYT24} \\ 
       & Train            & Test           & Train               & Test               \\ \hline
\#relations  & 29           & 29          & 24              & 24              \\ 
\#sentences    & 63,306           & 4,006          & 56,196              & 5,000              \\ 
\#tuples     & 78,973           & 5,859          & 88,366              & 8,120 \\ \hline
\#Entity overlap type       &             &            & \\ 
NEO  & 53,444           & 2,963          & 37,371              & 3,289              \\ 
EPO     & 8,379           & 898          & 15,124              & 1,410              \\ 
SEO    & 9,862           & 1,043          & 18,825              & 1,711              \\ \hline
\#tuples in a sentence       &             &            & \\ 
1 & 53,001           & 2,950          & 36,835              & 3,240              \\ 
2    & 6,154           & 595          & 12,065              & 1,047              \\ 
3    & 3,394           & 187          & 3,672              & 314              \\ 
4    & 450           & 239          & 2,623              & 290              \\ 
$\geq 5$    & 307           & 35          & 1,001              & 109              \\ \hline
\end{tabular}
}
\caption{Statistics of train/test split of the two datasets.}
\label{tab:data_stat}
\end{table}

\subsection{Evaluation Metrics}

We use the same evaluation method used by \citet{takanobu2019hrlre} in their experiments. We consider the extracted tuples as a set and remove the duplicate tuples. An extracted tuple is considered as correct if the corresponding full entity names are correct and the relation is also correct. We report precision, recall, and F1 score for comparison. 

\subsection{Parameter Settings}

We run the Word2Vec \citep{mikolov2013distributed} tool on the NYT corpus to initialize the word embeddings. The character embeddings and relation embeddings are initialized randomly. All embeddings are updated during training. We set the word embedding dimension $d_w=300$, relation embedding dimension $d_r=300$, character embedding dimension $d_c=50$, and character-based word feature dimension $d_f=50$. To extract the character-based word feature vector, we set the CNN filter width at $3$ and the maximum length of a word at $10$. The hidden dimension $d_h$ of the decoder LSTM cell is set at $300$ and the hidden dimension of the forward and the backward LSTM of the encoder is set at $150$. The hidden dimension of the forward and backward LSTM of the pointer networks is set at $d_p=300$. The model is trained with mini-batch size of $32$ and the network parameters are optimized using Adam \citep{kingma2014adam}. Dropout layers with a dropout rate fixed at $0.3$ are used in our network to avoid overfitting. 

\subsection{Baselines}

We compare our model with the following state-of-the-art joint entity and relation extraction models:

\noindent(1) SPTree \citep{miwa2016end}: This model is an end-to-end neural entity and relation extraction model using a sequence LSTM and a tree LSTM. The sequence LSTM identifies all the entities first, and then the tree LSTM finds the relation between all pairs of entities. They bring these two tasks together using shared parameters. Although this is an end-to-end model, entity detection and relation classification are still performed separately.

\noindent(2) Tagging \citep{zheng2017joint}: This is a neural sequence tagging model which jointly extracts the entities and relations using an LSTM encoder and an LSTM decoder. They used a Cartesian product of entity tags and relation tags to encode the entity and relation information together. This model does not work when tuples have overlapping entities.

\noindent(3) CopyR \citep{zeng2018copyre}: This model uses an encoder-decoder approach for the joint extraction of entities and relations. It copies only the last token of an entity from the source sentence. Their best performing multi-decoder model is trained with a fixed number of decoders where each decoder extracts one tuple. Since there is a separate decoder for each tuple, interaction among the tuples is very limited in this model.

\noindent(4) HRL \citep{takanobu2019hrlre}: This model uses a reinforcement learning (RL) algorithm with two levels of hierarchy for tuple extraction. A high-level RL finds the relation and a low-level RL identifies the two entities using a sequence tagging approach. This sequence tagging approach cannot always ensure extraction of exactly two entities.

\noindent(5) GraphR \citep{fu2019graphrel}: This model represents each token in a sentence as a node in a graph. The edges connecting the nodes represent the relations between them. They use a graph convolutional network (GCN) to predict the relation of each edge and then filter out some of the relations. Since each token is a node in the graph, there will be too many edges in it that do not represent any relation.

\noindent(6) N-gram Attention \citep{trisedya2019neural}: This model uses an encoder-decoder approach with N-gram attention mechanism for knowledge-base completion using distantly supervised data. The encoder uses the source tokens as its vocabulary and the decoder uses the entire Wikidata \citep{wikidata} entity IDs and relation IDs as its vocabulary. The encoder takes the source sentence as input and the decoder outputs the two entity IDs and relation ID for every tuple. During training, it uses the mapping of entity names and their Wikidata IDs of the entire Wikidata for proper alignment. Our task of extracting relation tuples with the raw entity names from a sentence is more challenging since entity names are not of fixed length. Our more generic approach is also helpful for extracting new entities which are not present in the existing knowledge bases such as Wikidata. We use this N-gram attention mechanism in Eq. (5.14) to compare its performance with other attention models (Table \ref{tab:ablation}).

\noindent(7) Pipeline Models: We also compare the performance of our joint extraction models against the pipeline extraction approaches. We use a pre-trained NER model from spaCy and a neural NER model to identify the named entities and use CNN, PCNN, and our proposed syntax-focused multi-factor attention (SFMFA) model for classifying the relations. 

Our neural NER model is similar to the model proposed by \citet{chiu2016named} and we use the `BIESO' sequence tagging approach. We are not interested in the type of entities, so we use a total of 5 NER tags. We use features based on word embeddings and character embeddings to represent the sentence tokens. These token vectors are passed to a Bi-LSTM layer for encoding. We use separate tag embeddings to represent the NER tags. We concatenate the tag embedding of the previous token with the Bi-LSTM hidden vectors and pass the concatenated vector to a feed-forward layer with softmax to classify the tags. We train this NER model on the corresponding training data with the sentences and entity names present in them. 

Our training data set contains many entity pairs that do not have any valid relation in the knowledge base. We consider them as {\em None} relation samples to train the relation classification models. We get around 100,000 and 110,000 {\em None} instances for the NYT29 and NYT24 datasets respectively. On the test data, first, we identify the entities using the NER models and then use the relation classification models to find the relation between each pair of these entities (or determine that there is no relation).

\subsection{Experimental Results}

We present our experimental results in Table \ref{tab:comparison}. The pipeline approaches perform better when they are used with the neural NER module. Since the neural NER module is trained on the training data, it performs better than the NER module of spaCy. We can see a very high variance in the results of the pipeline approaches with two different NER modules. This shows that pipeline approaches have too much dependency on the NER module and joint models are more suitable for this task.

\begin{table}[ht]
\centering
\resizebox{0.9\columnwidth}{!}{
\begin{tabular}{lllllll}
\hline
 & \multicolumn{3}{c}{NYT29} & \multicolumn{3}{c}{NYT24} \\ 
Model & Prec. & Rec. & F1 & Prec. & Rec. & F1 \\ \hline
Pipeline \\
spaCy NER + CNN & 0.037 & 0.063 & 0.047  & 0.187    & 0.723    & 0.297  \\ 
spaCy NER + PCNN & 0.034    & 0.044    & 0.086  & 0.190    & 0.714    & 0.301  \\ 
spaCy NER + SFMFA & 0.026    & 0.054    & 0.035  & 0.319    & 0.628    & 0.423  \\ 
Neural NER + CNN & 0.287 & 0.738 & 0.414  & 0.435    & 0.820    & 0.568  \\ 
Neural NER + PCNN & 0.287    & 0.737    & 0.412  & 0.435    & 0.815    & 0.567  \\ 
Neural NER + SFMFA & 0.420    & 0.582    & 0.488  & 0.588    & 0.771    & 0.667  \\ \hline
Joint single   \\
tagging & 0.593 & 0.381 & 0.464  & 0.624    & 0.317    & 0.420  \\ 
CopyR & 0.569 & 0.452 & 0.504  & 0.610    & 0.566    & 0.587   \\ 
SPTree & 0.492 & 0.557 & 0.522  &- &- &-  \\ 
GraphR & - & - & -  & 0.639 & 0.600 & 0.619  \\ 
HRL* & 0.692 & 0.601 & 0.643  & 0.781    & 0.771    & 0.776  \\ 
WDec* & \textbf{0.777} & 0.608 & \textbf{0.682}  & \textbf{0.881} & 0.761 & \textbf{0.817}  \\ 
PNDec* & 0.732 & \textbf{0.624} & 0.673 & 0.806 & \textbf{0.773} & 0.789  \\ \hline
Joint ensemble  \\
HRL & 0.764 & 0.604 & 0.674 & 0.842 & 0.778 & 0.809  \\ 
WDec & \textbf{0.846} & 0.621 & \textbf{0.716} & \textbf{0.945} & 0.762 & \textbf{0.844}  \\ 
PNDec & 0.815 & \textbf{0.639} & \textbf{0.716} & 0.893 & \textbf{0.788} & 0.838    \\ 
WDec\_PNDec & \textbf{0.846} & 0.621 & \textbf{0.716} & \textbf{0.945} & 0.762 & \textbf{0.844} \\ \hline
\end{tabular}
}
\caption{Performance comparison of the models on the two datasets. The rows with * show the median of five runs.}
\label{tab:comparison}
\end{table}

Among the joint extraction approaches, HRL achieves significantly higher F1 scores on the two datasets. We run their model and our models five times and report the median results in Table \ref{tab:comparison}. Scores of other baselines in Table \ref{tab:comparison} are taken from previous published papers \citep{zeng2018copyre,takanobu2019hrlre,fu2019graphrel}. Our {\em WordDecoding} (WDec) model achieves F1 scores that are $3.9\%$ and $4.1\%$ higher than HRL on the NYT29 and NYT24 datasets respectively. Similarly, our {\em PtrNetDecoding} (PNDec) model achieves F1 scores that are $3.0\%$ and $1.3\%$ higher than HRL on the NYT29 and NYT24 datasets respectively. We perform a statistical significance test (t-test) under a bootstrap pairing between HRL and our models and see that the higher F1 scores achieved by our models are statistically significant ($p < 0.001$). Next, we combine the outputs of five runs of our models and five runs of HRL to build ensemble models. For a test instance, we include those tuples which are extracted in the majority ($\geq 3$) of the five runs. This ensemble mechanism increases the precision significantly on both datasets with a small improvement in recall as well. In the ensemble scenario, compared to HRL, WDec achieves $4.2\%$ and $3.5\%$ higher F1 scores and PNDec achieves $4.2\%$ and $2.9\%$ higher F1 scores on the NYT29 and NYT24 datasets respectively. When we ensemble the five runs of WDec and five runs of PNDec together, we see that it achieves the same performance as the ensemble version of WDec, since the tuples extracted by the ensemble version of the PNDec model are a proper subset of the tuples extracted by the ensemble version of the WDec model.

\section{Analysis and Discussion}
\label{ed4jere-sec:analysis}

\subsection{Comparison between the Two Decoding Frameworks}

From Table \ref{tab:comparison}, we see that our proposed word-level decoding framework and pointer network-based decoding framework perform comparably. But pointer network decoding is much more intuitive than word-level decoding. The word-level decoder depends on the generation of the special tokens at appropriate time steps. It generates the relation names in the same way as it generates the entity tokens. Sometimes, it may generate non-relation tokens at the time steps when it must generate relation tokens. So we will not be able to extract the tuples from the generated sequence of tokens. This problem does not arise for pointer network decoding as it always extracts two entities and a relation between them at every time step.

Finding new entities in the test data is much simpler in pointer network decoding as it directly points to the entities in the input sentence. There is no direct way to find new entities in the test data in word-level decoding. We use the {\em UNK} token for this purpose. Whenever the decoder generates the {\em UNK} token, we replace it with the corresponding source token with the highest attention score. In pointer network decoding, we obtain the entire tuple representation at every time step and use it in the attention mechanism. But in word-level decoding, it is not possible to find the tuple representation, so we apply attention at the token level.

\subsection{Ablation Studies}

We include the ablation of our masking mechanism (+ Masking) and replacement of {\em UNK} token (+ Rep\_UNK) for the {\em WordDecoding} model in Table \ref{tab:ablation}. We see that both components contribute to significant improvements in F1 score for three types of attention. From Table \ref{tab:ablation}, we also see that {\em PtrNetDecoding} achieves the highest F1 scores when we combine the two attention mechanisms with respect to the previous hidden vector of the decoder LSTM ($\mathbf{h}_{t-1}^D$) and representation of all previously extracted tuples ($\mathbf{y}_{prev}$). 

\begin{table}[ht]
\centering
\resizebox{0.9\columnwidth}{!}{
\begin{tabular}{lllllll}
\hline
 & \multicolumn{3}{c}{NYT29} & \multicolumn{3}{c}{NYT24} \\ 
Model & Prec. & Rec. & F1 & Prec. & Rec. & F1 \\ \hline
WDec \\
Avg. & 0.638 & 0.523 & 0.575 & 0.771 & 0.683 & 0.724  \\ 
\quad + Masking & 0.709 & 0.561 & 0.626 & 0.843 & 0.717 & 0.775  \\ \hline
N-gram & 0.640 & 0.498 & 0.560 & 0.783 & 0.698 & 0.738  \\ 
\quad + Masking & 0.699 & 0.519 & 0.596 & 0.825 & 0.715 & 0.766 \\
\quad \quad + Rep\_UNK & 0.739 & 0.519 & 0.610 & 0.847 & 0.716 & 0.776  \\ \hline
Single & 0.683 & 0.545 & 0.607 & 0.816 & 0.716 & 0.763  \\ 
\quad + Masking & 0.723 & 0.567 & 0.636 & 0.842 & 0.728 & 0.781  \\
\quad \quad + Rep\_UNK & \textbf{0.777} & \textbf{0.608} & \textbf{0.682} & \textbf{0.881} & \textbf{0.761} & \textbf{0.817}  \\ \hline \hline
PNDec \\ 
$\mathrm{dec_{hid}}$ & 0.720 & 0.615 & 0.663 & 0.798 & 0.772 & 0.785  \\ 
$\mathrm{tup_{prev}}$ & 0.726 & 0.614 & 0.665 & 0.805 & 0.764 & 0.784  \\ 
$\mathrm{combo}$ & \textbf{0.732} & \textbf{0.624} & \textbf{0.673} & \textbf{0.806} & \textbf{0.773} & \textbf{0.789}  \\ \hline
\end{tabular}
}
\caption{Ablation of attention mechanisms with {\em WordDecoding} (WDec) and {\em PtrNetDecoding} (PNDec) model. For the Avg. type of attention in the WDec model, we do not replace the generated {\em UNK} token with any of the sentence tokens as they have the same attention weight.}
\label{tab:ablation}
\end{table}

\subsection{Performance Analysis}

From Table \ref{tab:comparison}, we see that CopyR, HRL, and our models achieve significantly higher F1 scores on the NYT24 dataset than the NYT29 dataset. Both datasets have a similar set of relations and similar texts (NYT). So task-wise both datasets should pose a similar challenge. However, the F1 scores suggest that the NYT24 dataset is easier than NYT29. The reason is that NYT24 has around 72.0\% of overlapping tuples between the training and test data (\% of test tuples that appear in the training data with different source sentences). In contrast, NYT29 has only 41.7\% of overlapping tuples. Due to the memorization power of deep neural networks, it can achieve much higher F1 score on NYT24. The difference between the F1 scores of {\em WordDecoding} and {\em PtrNetDecoding} on NYT24 is marginally higher than NYT29, since {\em WordDecoding} has more trainable parameters (about 27 million) than {\em PtrNetDecoding} (about 24.5 million) and NYT24 has very high tuple overlap. However, their ensemble versions achieve closer F1 scores on both datasets.

Despite achieving marginally lower F1 scores, the pointer network-based model can be considered more intuitive and suitable for this task. {\em WordDecoding} may not extract the special tokens and relation tokens at the right time steps, which is critical for finding the tuples from the generated sequence of words. {\em PtrNetDecoding} always extracts two entities of varying length and a relation for every tuple. We also observe that {\em PtrNetDecoding} is more than two times faster and takes one-third of the GPU memory of {\em WordDecoding} during training and inference. This speedup and smaller memory consumption are achieved due to the fewer number of decoding steps of {\em PtrNetDecoding} compared to {\em WordDecoding}. {\em PtrNetDecoding} extracts an entire tuple at each time step, whereas {\em WordDecoding} extracts just one word at each time step and so requires eight time steps on average to extract a tuple (assuming that the average length of an entity is two). The softmax operation at the projection layer of {\em WordDecoding} is applied across the entire vocabulary and the vocabulary size can be large (more than 40,000 for our datasets). In case of {\em PtrNetDecoding}, the softmax operation is applied across the sentence length (maximum of 100 in our experiments) and across the relation set (24 and 29 for our datasets). The costly softmax operation and the higher number of decoding time steps significantly increase the training and inference time for {\em WordDecoding}. The encoder-decoder model proposed by \citet{trisedya2019neural} faces a similar softmax-related problem as their target vocabulary contains the entire Wikidata entity IDs and relation IDs which is in the millions. HRL, which uses a deep reinforcement learning algorithm, takes around 8x more time to train than {\em PtrNetDecoding} with a similar GPU configuration. The speedup and smaller memory consumption will be useful when we move from sentence-level extraction to document-level extraction, since document length is much higher than sentence length and a document contains a higher number of tuples.

\begin{table}[ht]
\centering
\resizebox{0.8\columnwidth}{!}{
\begin{tabular}{l|l|l|lll}
\hline
                       &                           & Model      & Prec. & Rec.  & F1    \\ \hline
\multirow{6}{*}{NYT29} & \multirow{3}{*}{Ent}   & HRL      & 0.833 & 0.827 & 0.830 \\
                       &                             & WDec      & \textbf{0.865} & 0.812 & 0.838 \\
                       &                           & PNDec & 0.858 & \textbf{0.851} & \textbf{0.855} \\ \cline{2-6} 
                       & \multirow{3}{*}{Rel} & HRL      & 0.846 & 0.745 & 0.793 \\  
                       &                         & WDec      & \textbf{0.895} & 0.729 & 0.803 \\
                       &                           & PNDec & 0.884 & \textbf{0.770} & \textbf{0.823} \\ \hline
\multirow{6}{*}{NYT24} & \multirow{3}{*}{Ent}   & HRL      & 0.887 & 0.892 & 0.890 \\
                       &                        & WDec      & \textbf{0.926} & 0.858 & 0.891 \\
                       &                           & PNDec & 0.906 & \textbf{0.901} & \textbf{0.903} \\ \cline{2-6} 
                       & \multirow{3}{*}{Rel} & HRL      & 0.906 & 0.896 & 0.901 \\ 
                       &                        & WDec      & \textbf{0.941} & 0.880 & 0.909 \\
                       &                           & PNDec & 0.930 & \textbf{0.921} & \textbf{0.925} \\ \hline
\end{tabular}
}
\caption{Comparison on entity and relation generation tasks.}
\label{tab:ent_rel}
\end{table}

\begin{table}[ht]
\centering
\resizebox{0.8\columnwidth}{!}{
\begin{tabular}{lcccccc}
\hline
      & \multicolumn{3}{c}{NYT29} & \multicolumn{3}{c}{NYT24} \\ 
Model & Order    & Ent1   & Ent2   & Order    & Ent1   & Ent2   \\ \hline
HRL  & 0.2      & 5.9    & 6.6    & 0.2      & 4.7    & 6.3    \\
WDec  & 0.0      & 4.2    & 4.7    & 0.0      & 2.4    & 2.4    \\ 
PNDec & 0.8      & 5.6    & 6.0    & 1.0      & 4.0    & 6.1    \\ \hline
\end{tabular}
}
\caption{\% errors for wrong ordering and entity mismatch.}
\label{tab:error}
\end{table}

\subsection{Error Analysis}

The relation tuples extracted by a joint model can be erroneous for multiple reasons such as: (i) extracted entities are wrong; (ii) extracted relations are wrong; (iii) pairings of entities with relations are wrong. To see the effects of the first two reasons, we analyze the performance of HRL and our models on entity generation and relation generation separately. For entity generation, we only consider those entities which are part of some tuple. For relation generation, we only consider the relations of the tuples. We include the performance of our two models and HRL on entity generation and relation generation in Table \ref{tab:ent_rel}. Our proposed models perform better than HRL on both tasks. Comparing our two models, {\em PtrNetDecoding} performs better than {\em WordDecoding} on both tasks, although {\em WordDecoding} achieves higher F1 scores in tuple extraction. This suggests that {\em PtrNetDecoding} makes more errors while pairing the entities with relations. We further analyze the outputs of our models and HRL to determine the errors due to ordering of entities (Order), mismatch of the first entity (Ent1), and mismatch of the second entity (Ent2) in Table \ref{tab:error}. {\em WordDecoding} generates fewer errors than the other two models in all the categories and thus achieves the highest F1 scores on both datasets.

\section{Summary}
\label{ed4jere-sec:summary}

Jointly extracting entities and relations from sentences is a challenging task due to different lengths of entities, the presence of multiple tuples, and overlapping of entities among tuples. In this chapter, we describe two {\em novel} approaches using encoder-decoder architecture to address this task. Experiments on the New York Times (NYT) corpus show that our proposed models achieve significantly improved new state-of-the-art F1 scores.

%
\chapter{A Hierarchical Entity Graph Convolutional Network for Relation Extraction across Documents}
\label{chap:mhre}

Distantly supervised relation extraction models mostly focus on sentence-level relation extraction, where the two entities (subject and the object entity) of a relation tuple must appear in the same sentence. This assumption is overly strict and for a large number of relations, we may not find sentences containing the two entities. To solve this problem, we propose multi-hop relation extraction, where the two entities of a relation tuple may appear in two different documents but these documents are connected via some common entities. We can find a chain of entities from the subject entity to the object entity via common entities. The relation between the subject and object entities can be established using this entity chain. Following this multi-hop approach, we create a dataset for 2-hop relation extraction, where each chain contains exactly two documents. This dataset covers a higher number of relations than sentence-level extraction. We also propose a hierarchical entity graph convolutional network (HEGCN) model to solve this task, consisting of a two-level hierarchy of graph convolutional networks (GCNs). The first-level GCN of the hierarchy captures the relations among the entity mentions within the documents, and the second-level GCN of the hierarchy captures the relations among the entity mentions across the documents in a chain. Our proposed HEGCN model improves the performance by 2.2\% F1 score on our 2-hop relation extraction dataset, and it can be readily extended to N-hop datasets.

\section{Motivation}

The task of relation extraction is to find relation tuples from free text. \cite{mintz2009distant}, \cite{riedel2010modeling}, and \cite{hoffmann2011knowledge} proposed the idea of distant supervision to automatically obtain a large amount of training data for this task. The idea is to map the relation tuples in existing knowledge bases (KBs) to text corpora such as Wikipedia or news articles. The assumption is that if the subject and object entities of a relation tuple appear in a sentence, then this sentence can be considered as providing evidence of the relation between the entities. Most relation extraction work focuses on such distantly supervised sentence-level extraction. This method can give a significantly large amount of training data to build supervised relation extraction models. But the assumption of distant supervision that the two entities of a tuple must appear in the same sentence is overly strict. We may not find an adequate number of evidence sentences for many relations if the two entities in a relation do not appear in the same sentence. The relation extraction models built on such data can find relations only for a small number of relations and most relations of the KBs will be out of the reach of such models.

\begin{table}[ht]
\small
\centering
\begin{tabular}{|l|l|}
\hline
\begin{tabular}[l]{@{}l@{}}D1: \textcolor{red}{Ghanshyam Tiwari} \\held the position of \\education minister in \\Government of \textcolor{orange}{Rajasthan}.\end{tabular}                 & \begin{tabular}[l]{@{}l@{}}D1: \textcolor{red}{Arnage} is a commune \\in the Sarthe department \\in the region of Pays-de-la\\-Loire in north-western \\\textcolor{orange}{France}.\end{tabular} \\ \hline
\begin{tabular}[l]{@{}l@{}}D2: Shekhawati is a \\semi-arid historical region \\located in the northeast \\part of \textcolor{orange}{Rajasthan}, \textcolor{blue}{India}.\end{tabular} & \begin{tabular}[l]{@{}l@{}}D2: \textcolor{blue}{Paris} is the capital \\and most populous city \\of \textcolor{orange}{France}.\end{tabular}                                                \\ \hline
\begin{tabular}[c]{@{}l@{}}Subj.: \textcolor{red}{Ghanshyam Tiwari} \\ Obj.: \textcolor{blue}{India}\\ Common: \textcolor{orange}{Rajasthan}\end{tabular}                                                                                & \begin{tabular}[c]{@{}l@{}}Subj.: \textcolor{red}{Arnage}\\ Obj.: \textcolor{blue}{Paris}\\ Common: \textcolor{orange}{France}\end{tabular}                                                                             \\ \hline
Rel: country\_of\_citizenship                                                                                                                                                            & Rel: None                                                                                                                                                               \\ \hline
\end{tabular}
\caption{Examples of 2-hop relations.}
\label{tab:examples}
\end{table}

To solve this problem, we propose a multi-hop relation extraction task where the subject and object entities of a tuple can appear in two different documents, and these two documents are connected via some common entities. We can create a chain of entities from the subject entity to the object entity of a tuple via the common entities across multiple documents. Each link in this chain represents a relation between the entities located at the endpoints of the link. We can determine the relation between the subject and object entities of a tuple by following this chain of relations. This approach can give training instances for more relations than sentence-level distant supervision. Following the proposed multi-hop approach, we create a 2-hop relation extraction dataset for the task. Each instance of this dataset has two documents, where the first document contains the subject entity and the second document contains the object entity of a tuple. These two documents are connected via at least one common entity. We have included one positive and one negative example of 2-hop relations in Table \ref{tab:examples}. This idea can be extended to create an N-hop dataset.

We also propose a hierarchical entity graph convolutional network (HEGCN) model for the task. Our proposed model has two levels of graph convolutional networks (GCNs). The first-level GCN of the hierarchy is applied to the entity mention level graph of every document to capture the relations among the entity mentions within a document. The second-level GCN of the hierarchy is applied on a unified entity level graph, which is built using all the unique entities present in the document chain. This entity level graph can be built on the document chain of any length and it can capture the relations among the entities across the multiple documents in the chain. Our proposed HEGCN model improves the performance on our 2-hop dataset. To summarize, the following are the contributions of this work:\\
(1) We propose a multi-hop relation extraction task and create a distantly supervised dataset for the task. Our dataset has more relations than the other popular distantly supervised sentence-level or document-level relation extraction datasets. \\
(2) We propose a novel hierarchical entity graph convolutional network (HEGCN) for multi-hop relation extraction. Our proposed model improves the F1 score by 2.2\% on our multi-hop dataset, compared to strong neural baselines.

\section{Problem Definition}

Multi-hop relation extraction can be defined as follows. Consider two entities, a subject entity $e_s$ and an object entity $e_o$, and a chain of documents $D=\{D_s \rightarrow D_1 \rightarrow D_2 \rightarrow ... \rightarrow D_n \rightarrow D_o\}$ where $e_s \in D_s$ and $e_o \in D_o$. There exists a chain of entities $e_s \rightarrow c_1 \rightarrow c_2 \rightarrow ... \rightarrow c_{n+1} \rightarrow e_o$ where $c_1 \in \{D_s, D_1\}$, $c_2 \in \{D_1, D_2\}, ... , $ $c_{n+1} \in \{D_{n}, D_o\}$. The task is to find the relation between $e_s$ and $e_o$ from a pre-defined set of relations $R \cup \{\mathit{None}\}$, where $R$ is the set of relations and {\em None} indicates that none of the relations in $R$ holds between $e_s$ and $e_o$. A simpler version of this task is 2-hop relation extraction where $D_s$ and $D_o$ are directly connected by at least one common entity. In this work, we focus on 2-hop relation extraction.

\section{Multi-Hop Dataset Construction}

We create a dataset for this multi-hop relation extraction from a multi-hop question-answering (QA) dataset WikiHop \citep{welbl2018constructing}. \citet{welbl2018constructing} defined the multi-hop QA task as follows: Given a set of supporting documents $D_s$ and a set of candidate answer $C_a$ which are mentioned in $D_s$, the goal is to find the correct answer $a^* \in C_a$ by drawing on the supporting documents. They used Wikipedia articles and Wikidata \citep{wikidata} tuples for creating this dataset. Each positive tuple $(e_s, e_o, r_p)$ in Wikidata has two entities, a subject entity $e_s$ and an object entity $e_o$, and a positive relation $r_p$ between the subject and object entity. The questions are created by combining the subject entity $e_s$ and the relation $r_p$, and the object entity $e_o$ is the correct answer $a^*$ for a given question. The other candidate answers are carefully chosen from Wikidata entities so that they have a similar type as the correct answer. The supporting documents are chosen in such a way that at least two documents are needed to find the correct answer. This means the subject entity $e_s$ and the object entity $e_o$ do not appear in the same document. They used a bipartite graph partition technique to create the dataset. In this bipartite graph, vertices on one side correspond to Wikidata entities, and vertices on the other side correspond to Wikipedia articles. An edge is created between an entity vertex and a document vertex if this document contains the entity. As we traverse the graph starting from vertex $e_s$, it visits many document vertices and entity vertices. This constitutes the supporting document set and candidate answer set. If the candidate answer set does not contain the object entity $e_o$ which is the correct answer, this instance is discarded. They also limited the length of the traversal to 3 documents. \citet{welbl2018constructing} only released the supporting documents, questions, and candidate answers for their dataset. They did not release the connecting entities.

We convert this WikiHop dataset into a multi-hop relation extraction dataset. The subject entities and the candidate entities can be easily found in the documents using string matching. We use a named entity recognizer from spaCy\footnote{\url{https://spacy.io/}} to find the other entities in the documents and these entities can link these documents. We find that most of the WikiHop question-answer instances are 2-hop instances. That means for most of the instances of WikiHop dataset, there is at least one document pair in the supporting document set where the first document of the pair contains the subject entity and the second document of the pair contains the correct answer, and these two documents in the pair are directly connected via some third entity. To simplify the multi-hop relation extraction task, we fix the hop count at 2. For every instance of the WikiHop dataset, we can easily find the subject entity $e_s$ and the positive relation $r_p$ from the question. The correct answer $a^*$ is the object entity of a positive tuple. $(e_s, a^*, r_p)$ is the positive tuple for relation extraction. For any other candidate answer $e_w \in C_a-\{a^*\}$, the entity pair $(e_s, e_w)$ is considered as a {\em None} tuple if there exists no relation among the four pairs $(e_s, e_w)$, $(e_w, e_s)$, $(e_w, e_o)$, and $(e_o, e_w)$ in Wikidata. We check for the no relation condition for these four entity pairs involving $e_w$, $e_s$, and $e_o$ to reduce the distant supervision noise in the dataset for {\em None} tuples. We create a {\em None} candidate set $C_n$ with each $e_w \in C_a-\{a^*\}$. We first find all possible pairs of documents from the supporting document set $D_s$ such that the first document of the pair contains the subject entity $e_s$ and the second document of the pair contains either the entity $a^*$ or one of the entities from $C_n$. We discard those pairs of documents that do not contain any common entity. The document pairs where the second document contains the entity $a^*$ are considered as a document chain for the positive tuple $(e_s, a^*, r_p)$ where $r_p \in R$. All other document pairs where the second document contains an entity from the set $C_n$ are considered as a document chain for {\em None} tuple $(e_s, e_w, None)$ where $e_w \in C_n$. In this way, using distant supervision, we can create a dataset for 2-hop relation extraction. Each instance of this dataset has a chain of documents $D=\{D_s \rightarrow D_o\}$ of length 2 that is the textual source of a tuple $(e_s, e_o, r)$. The document $D_s$ contains the subject entity $e_s$ and the document $D_o$ contains the object entity $e_o$. The two documents are connected with at least one common entity $c$. There exists at least one entity chain $e_s \rightarrow c \rightarrow e_o$ in the document chain. The goal is to find the relation $r$ between $e_s$ and $e_o$ from the set $R \cup \{\mathit{None}\}$. This dataset is our multi-hop relation extraction dataset (MHRED), to be used in the remaining sections of this chapter.

\begin{table}[ht]
\small
\centering
\begin{tabular}{|l|l|}
\hline
Question      & located\_in\_the\_administrative\_territorial\_entity Zoo Lake                                                                                                                                                                                                                                                                                                                                                                                                                                                                                                                  \\ \hline
Candidates & Gauteng, Tanzania                                                                                                                                                                                                                                                                                                                                                                                                                                                                               \\ \hline
Answer     & Gauteng                                                                                                                                                                                                                                                                                                                                                                                                                                                                                                                           \\ \hline
Doc1       & \begin{tabular}[c]{@{}l@{}}\textcolor{red}{Zoo Lake} is a popular lake and public park in \textcolor{orange}{Johannesburg} , \\\textcolor{orange}{South Africa} . It is part of the Hermann Eckstein Park and is \\opposite the Johannesburg Zoo . The \textcolor{red}{Zoo Lake} consists of two \\dams , an upper feeder dam , and a larger lower dam , both \\constructed in natural marshland watered by the Parktown \\Spruit .\end{tabular}                                                                                                                                                       \\ \hline
Doc2       & \begin{tabular}[c]{@{}l@{}}\textcolor{orange}{Johannesburg} is the largest city in \textcolor{orange}{South Africa} and is one of \\the 50 largest urban areas in the world . It is the provincial \\capital of \textcolor{blue}{Gauteng} , which is the wealthiest province in \\\textcolor{orange}{South Africa} . \end{tabular} \\ \hline
Doc3       & \begin{tabular}[c]{@{}l@{}}Mozambique is a country in Southeast Africa bordered by the \\Indian Ocean to the east , \textcolor{blue}{Tanzania} to the north , Malawi \\and Zambia to the northwest , Zimbabwe to the west , and \\Swaziland and \textcolor{orange}{South Africa} to the southwest . It is separated \\from Madagascar by the Mozambique Channel to the east . \end{tabular}                                                                                                                                                                                                                                                                                                         \\ \hline
\end{tabular}
\caption{A multi-hop question-answer instance from the WikiHop dataset.}
\label{tab:wikihop}
\end{table}

We include a multi-hop question-answer instance of the WikiHop dataset in Table \ref{tab:wikihop}, which has a question, two candidate answers, and three documents. We can obtain the positive relation \\{\em located\_in\_the\_administrative\_territorial\_entity} and subject entity {\em Zoo Lake} from the question. The subject entity appears in Doc1. Two candidate answers {\em Gauteng} and {\em Tanzania} appear in Doc2 and Doc3 respectively. Doc1 and Doc2 have two common entities {\em Johannesburg} and {\em South Africa}. Doc1 and Doc3 have a common entity {\em South Africa}. Since the correct answer to the question is {\em Gauteng}, the quintuple (Doc1, {\em Zoo Lake}, Doc2, {\em Gauteng}, {\em located\_in\_the\_administrative\_territorial\_entity}) constitutes a positive instance in the MHRED dataset. The quintuple (Doc1, {\em Zoo Lake}, Doc3, {\em Tanzania}, {\em None}) constitutes a negative instance in the MHRED dataset.

\subsection{Dataset Statistics}

The training, validation, and test data of the WikiHop dataset are created using distant supervision, but the validation and test data are manually verified. WikiHop test data is blind and not released. So we use their validation data to create the test data for our task and use their training data for our training and validation purposes. We include the statistics of our multi-hop relation extraction dataset in Table \ref{tab:mhre_dataset_stat}. We include the statistics on the number of common entities present in the two documents of a chain in Table \ref{tab:mhre_common_entity_stat}. We split the training data randomly, with 90\% for training and 10\% for validation. From Table \ref{tab:mhre_dataset_stat}, we see that the dataset contains a much higher number of {\em None} tuples than the positive tuples. So we randomly select {\em None} tuples so that the number of {\em None} tuples is the same as the number of positive tuples for training and validation. For evaluation, we consider the entire test dataset. From Table \ref{tab:mhre_relations}, we see that our MHRED dataset contains more relations than any other distantly supervised relation extraction datasets such as the New York Times \citep{riedel2010modeling,hoffmann2011knowledge} or DocRED \citep{yao2019DocRED}.

\begin{table}[ht]
\centering
\begin{tabular}{l|ccc}
\hline
                     & \multicolumn{1}{c}{Train} & \multicolumn{1}{c}{Test} \\ \hline
\#Positive relations    & 218       & 72       \\ 
\#Document chains    & 143,906     & 5,320          \\ 
\#Positive instances & 40,247      & 1,672          \\
\#Positive entity pairs & 21,490   & 618             \\
\#None instances     & 197,731     & 7,806          \\ \hline
\end{tabular}
\caption{Statistics of the MHRED dataset.}
\label{tab:mhre_dataset_stat}
\end{table}

\begin{table}[ht]
\centering
\begin{tabular}{c|cc}
\hline
\multicolumn{1}{l|}{} & \multicolumn{2}{l}{\#Document chains} \\ \hline
\#Common entities      & Train              & Test              \\ \hline
1                      & 92,140             & 3,615             \\ 
2                      & 36,275             & 1,161             \\ 
3                      & 10,824             & 374               \\ 
4                      & 3,170              & 113               \\ 
$\geq$5                 & 1,497              & 57                \\ \hline
\end{tabular}
\caption{Statistics of the common entities in the MHRED dataset.}
\label{tab:mhre_common_entity_stat}
\end{table}

\begin{table}[ht]
\centering
\begin{tabular}{l|c|l|c}
\hline
Dataset            & $\vert R \vert$ & Dataset & $\vert R \vert$ \\ \hline
ACE04              & 7     & CoNLL04            & 5  \\
SemEval 2010        & 9   & GDS                & 4   \\ 
NYT10              & 53     & NYT11              & 24   \\ 
TACRED             & 41   & DocRED             & 96     \\ 
FewRel 2.0  & 100 & \textbf{MHRED}              & \textbf{218}   \\ \hline
\end{tabular}
\caption{The number of relations in various relation extraction datasets. $R$ is the set of positive relations.}
\label{tab:mhre_relations}
\end{table}

\section{Model Description}

We propose a hierarchical entity graph convolutional network (HEGCN) for the multi-hop relation extraction. We encode the documents in a document chain using a bi-directional long short-term memory (BiLSTM) layer \citep{hochreiter1997long}. On top of the BiLSTM layer, we use two graph convolutional networks (GCN), one after another in a hierarchy. In the first level of the GCN hierarchy, we construct a separate entity mention graph on each document of the chain using all the entities mentioned in that document. Each mention of an entity in a document is considered as a separate node in the graph. We use a graph convolutional network (GCN) on the entity mention graph of each document to capture the relations among the entity mentions in the document. This mention-level graph helps to pass information among the entity mentions in a document. We then construct a unified entity-level graph across all the documents in the chain. Each node of this entity-level graph represents a unique entity in the document chain. Each common entity between two documents in the chain is represented by a single node in the graph. We use a GCN on this unified entity-level graph to capture the relations among the entities across the documents. This unified entity graph helps to pass information among the multiple documents. This graph is a local knowledge graph involving the entities present in the documents. GCN is used to enrich the representations of the entities in the graph. We concatenate the representations of the nodes of the subject entity and object entity and pass it to a feed-forward layer with softmax for relation classification.

\begin{figure*}[ht]
\centering
\includegraphics[scale=0.45]{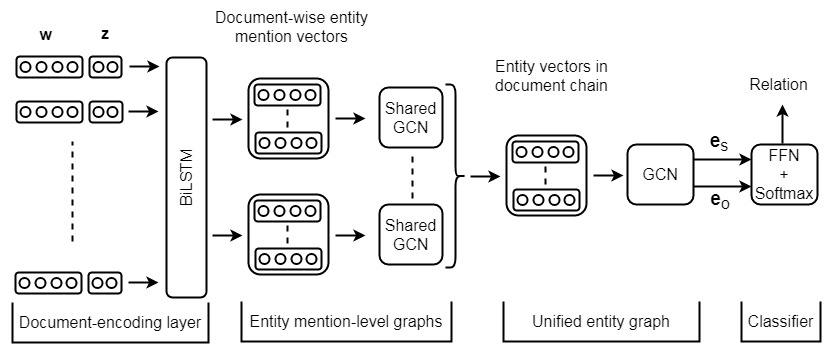}
\caption{The architecture of our proposed HEGCN model. First, a GCN in the hierarchy is shared across the entity mention graphs of the documents in a chain. This diagram is for a document chain of length 2.}
\label{fig:hegcn}
\end{figure*}

\subsection{Document Encoding Layer}

We use two types of embedding vectors: (1) word embedding vector $\mathbf{w} \in \mathbb{R}^{d_w}$ (2) entity token indicator embedding vector $\mathbf{z} \in \mathbb{R}^{d_z}$, which indicates if a word belongs to the subject entity, object entity, or common entities. The subject and object entities are assigned the embedding index of $2$ and $3$, respectively. The common entities in the document chain are assigned embedding index in an increasing order starting from index $4$. The same entities present in two documents in the chain get the same embedding index. Embedding index $0$ is used for padding and $1$ is used for all other tokens in the documents. A document is represented using a sequence of vectors  $\{\mathbf{x}_1, \mathbf{x}_2,....., \mathbf{x}_n\}$ where $\mathbf{x}_t = \mathbf{w}_t \Vert \mathbf{z}_t$. $\Vert$ represents the concatenation of vectors and $n$ is the document length. We concatenate all documents in a chain sequentially by using a document separator token. These token vectors are passed to a BiLSTM layer to capture the interaction among the documents in a chain. 
\begin{align}
&\mathbf{x}_t = \mathbf{w}_t \Vert \mathbf{z}_t \nonumber\\
&\overrightarrow{\mathbf{h}_t} = \overrightarrow{\mathrm{LSTM}}(\mathbf{x}_t, \mathbf{h}_{t-1}) \nonumber\\
&\overleftarrow{\mathbf{h}_t} = \overleftarrow{\mathrm{LSTM}}(\mathbf{x}_t, \mathbf{h}_{t+1}) \nonumber\\
&\mathbf{h}_t = \overrightarrow{\mathbf{h}_t} || \overleftarrow{\mathbf{h}_t}
\end{align}
\noindent $\overrightarrow{\mathbf{h}_t} \in \mathbb{R}^{(d_w+d_z)}$ and $\overleftarrow{\mathbf{h}_t} \in \mathbb{R}^{(d_w+d_z)}$ are the output at the $t$th step of the forward LSTM and backward LSTM respectively. We concatenate them to obtain the $t$th BiLSTM output $\mathbf{h}_t \in \mathbb{R}^{2(d_w+d_z)}$ (Eq. (6.1)).

\subsection{Hierarchical Entity Graph Convolutional Layers}

\citet{Kipf2017SemiSupervisedCW} proposed graph convolutional networks (GCN) which work on graph structures. Here, we describe the GCN which is used in our model. We represent a graph $\mathcal{G}$ with $m$ nodes using an adjacency matrix $\mathbf{A}$ of size $m \times m$. If there is an edge between node $i$ and node $j$, then $A_{ij} = A_{ji} = 1$. Self loops, $A_{ii} = 1$, are added in the graph $\mathcal{G}$ so that the nodes keep their own information too. We normalize the adjacency matrix $\mathbf{A}$ by using symmetric normalization proposed by \citet{Kipf2017SemiSupervisedCW}. A diagonal node degree matrix $\mathbf{D}$ of size $m \times m$ is used in the normalization of $\mathbf{A}$.
\begin{align}
    &D_{ij}=\begin{cases}
    \text{deg}(v_i) & \text{if } i=j\\
    0 & \text{otherwise}
\end{cases}\nonumber\\
    &D^{-\frac{1}{2}}_{ij} = \begin{cases}
    \frac{1}{\sqrt{D}_{ij}} & \text{if } i=j\\
    0 & \text{otherwise}
\end{cases}\nonumber\\
    &\hat{\mathbf{A}} = \mathbf{D}^{-\frac{1}{2}} \mathbf{A} \mathbf{D}^{-\frac{1}{2}} 
\end{align}
\noindent where $\text{deg}(v_i)$ is the number of edges that are connected to the node $v_i$ in $\mathcal{G}$ and $\hat{\mathbf{A}}$ (Eq. (6.2)) is the corresponding normalized adjacency matrix of $\mathcal{G}$.

Each node of the graph receives the hidden representation of its neighboring nodes from the $(l-1)$th layer and uses the following operation (Eq. (6.3)) to update its own hidden representation.
\begin{align}
    &\textbf{g}_i^l = \text{ReLU}(\sum_{j=1}^m \hat{A}_{ij} \textbf{W}^l \textbf{g}_j^{l-1})
\end{align}
\noindent $\textbf{W}^l$ is the trainable weight matrix of the $l$th layer of the GCN, $\textbf{g}_i^l$ is the representation of the $i$th node of the graph at the $l$th layer. If $\textbf{g}_i^l$ has the dimension of $d_g$, then the dimension of the weight matrix $\textbf{W}^l$ is $d_g \times d_g$. $\textbf{g}_i^0$ is the initial input to the GCN.

\subsubsection{Entity Mention Graph Layer}

We construct an entity mention graph (EMG) for each document in the chain on top of the document encoding layer. An entity string may appear at multiple locations in a document and each appearance is considered as an entity mention. We add a node in the graph for each entity mention. We connect two entity mention nodes if they appear in the same sentence (EMG type 1 edge). We assume that since they appear in the same sentence, there may exist some relation between them. We also connect two entity mention nodes if the strings of the two entity mentions are identical (EMG type 2 edge). Let $e_1, \ldots, e_l$ be the sequence of entity mention nodes listed in the order of their appearance in a document. We connect nodes $e_i$ and $e_{i+1}$ ($1 \leq i < l$) with an edge (EMG type 3 edge). EMG type 3 edges create a linear chain of the entity mentions and ensure that the graph is connected. We use a graph convolutional network on this graph topology to capture the relations among the entity mentions in a document.

We obtain the initial representations of the entity mention nodes from the hidden vectors of the document encoding layer. We concatenate the hidden vectors of the first and last token of an entity mention and a context vector to obtain the initial representation of the entity mention node. We derive this context vector using an attention mechanism on the tokens of the sentence in which the entity mention appears. This context vector provides the sentence-specific contextual information regarding the entity mention.
\begin{align}
    &\textbf{p} = \textbf{h}_b ~\Vert~ \textbf{h}_e \nonumber\\
    &s_t = \text{tanh}(\textbf{p}^T \textbf{W}) \textbf{h}_t \nonumber\\
    &\textbf{a} = \text{softmax}({[s_1  s_2  \ldots  s_k]}^T) \nonumber\\
    &\textbf{c} =  \sum_{t=1}^{k} \text{a}_t \textbf{h}_t \nonumber\\
    &\textbf{q} = \textbf{p} ~\Vert~ \textbf{c}
\end{align}
\noindent $\textbf{h}_b \in \mathbb{R}^{2(d_w+d_z)}$ and $\textbf{h}_e \in \mathbb{R}^{2(d_w+d_z)}$ are the hidden vectors from the document encoding layer of the first and last token of an entity mention. $\textbf{W} \in \mathbb{R}^{4(d_w+d_z) \times 2(d_w+d_z)}$ is a trainable weight matrix, $\textbf{h}_t \in \mathbb{R}^{2(d_w+d_z)}$ is the hidden vector of the $t$th token of the sentence in which the entity mention is located, superscript $T$ represents the transpose operation. $\text{a}_t$ is the normalized attention score for the $t$th token with respect to the entity mention, $k$ is length of the sentence in which the entity mention is located, and $\textbf{c} \in \mathbb{R}^{2(d_w+d_z)}$ is the context vector. The entity mention node vector $\textbf{q} \in \mathbb{R}^{6(d_w+d_z)}$ of the $i$th node in the graph is passed to the GCN as $\textbf{g}_i^0$. The parameters of this GCN are shared across the documents in a chain. This layer of the model is referred to as entity mention-level graph convolutional network or EMGCN.

\begin{figure*}[t]
\centering
\includegraphics[scale=0.45]{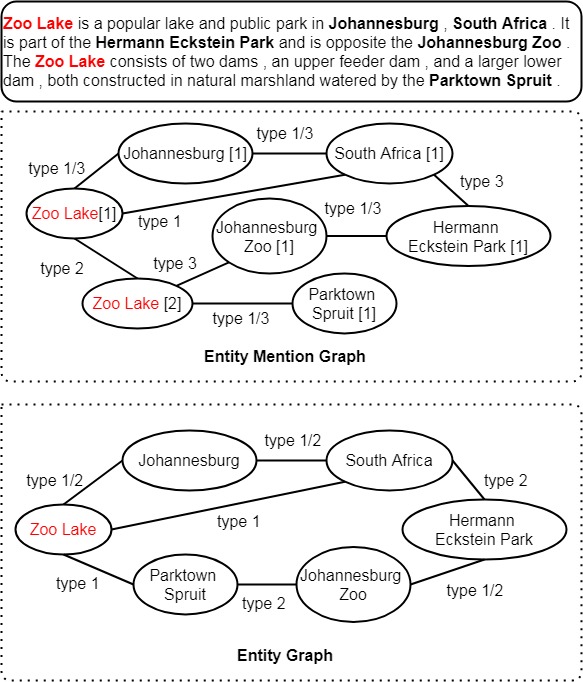}
\caption{Entity mention graph and entity graph construction from Doc1 of Table \ref{tab:wikihop}. Entity mentions are marked in bold font. The numbers in square brackets ([x]) in the entity mention graph are used to distinguish the entity mentions with identical string. Type x/y means this edge can be of both type x and type y. The `EMG' and `EG' prefixes are omitted from the labels of the edges in the entity mention graph and entity graph respectively.}
\label{fig:doc1_hegcn}
\end{figure*}

\subsubsection{Entity Graph Layer}

We construct a unified entity graph (EG) on top of the entity mention graphs. First, we construct an entity graph for each document, where each unique entity string is represented as an entity node in the graph. We add an edge between two entity nodes if the strings of the two entities appear together in at least one sentence in the document (EG type 1 edge). We also form a sequence of entity nodes based on the order of appearance of the entities in a document, where only the first occurrence of multiple occurrences of an entity is kept in the sequence. We connect two consecutive entity nodes in the sequence with an edge (EG type 2 edge). This ensures that the entire entity graph remains connected. 

We construct one entity graph for each document in the document chain. We unify the entity graphs of multiple documents by merging the nodes of common entities between them. The unified entity graph contains all the nodes from the multiple entity graphs, but the common entity nodes which appear in two entity graphs are merged into one node in the unified graph. There is an edge between two entity nodes in the unified entity graph if there exists an edge between them in any of the entity graphs of the documents.

We obtain the initial representations of the entity nodes from the GCN outputs of the entity mention graphs. For the common entities between two documents, we average the GCN outputs of the entity mention nodes that have an identical string as the entity from the entity mention graphs of the two documents. For other entity nodes that appear only in one document, we average the GCN outputs of the entity mention nodes that have an identical string as the entity from the entity mention graph of that document. Each entity vector is passed to another graph convolutional network as $\textbf{g}_i^0$ which represents the initial representation of the $i$th entity node in the unified entity graph. We use a graph convolutional network on this graph topology to capture the relations among the entities across the documents in the document chain. This layer of the model is referred to as entity-level graph convolutional network or EGCN. 

\subsubsection{Example of the Graph Construction}

Here, we describe how the graphs are constructed for the positive MHRED instance mentioned in Table \ref{tab:wikihop}. We show how the entity mention graph and entity graph are constructed from Doc1 of Table \ref{tab:wikihop} in Figure \ref{fig:doc1_hegcn}. Each node in the entity mention graph refers to an entity mention in the document. The entity string {\em Zoo Lake} appears twice in the document, so there are two entity mention nodes {\em Zoo Lake [1]} and {\em Zoo Lake [2]} in the entity mention graph. The two entity mentions {\em Zoo Lake [1]} and {\em Johannesburg [1]} appear in the same sentence, so a EMG type 1 edge is added between them in the graph. All other EMG type 1 edges are added in a similar way. {\em Zoo Lake [1]} and {\em Zoo Lake [2]} are connected with a EMG type 2 edge as they have identical entity mention string. The sequence of entity mentions in the document is follows: {\em Zoo Lake [1]} $\rightarrow$ {\em Johannesburg [1]} $\rightarrow$ {\em South Africa [1]} $\rightarrow$ {\em Hermann Eckstein Park [1]} $\rightarrow$ {\em Johannesburg Zoo [1]} $\rightarrow$ {\em Zoo Lake [2]} $\rightarrow$ {\em Parktown Spruit [1]}. An EMG Type 3 edge is added between two nodes $n_1$ and $n_2$ in the entity mention graph if $n_2$ follows $n_1$ in this sequence.

\begin{figure*}[t]
\centering
\includegraphics[scale=0.48]{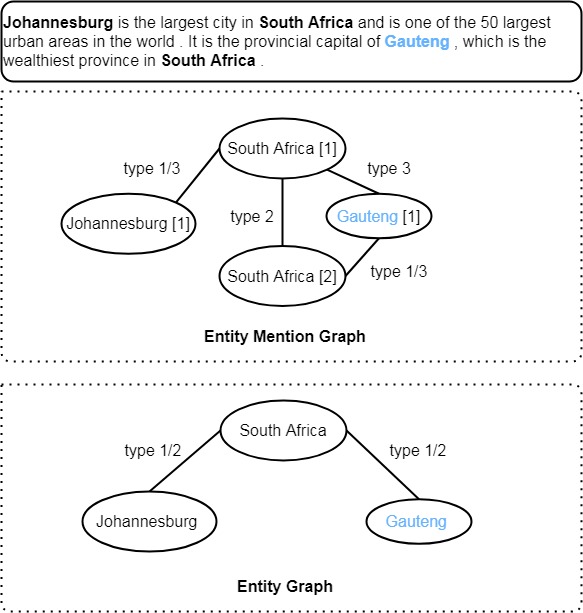}
\caption{Entity mention graph and entity graph construction from Doc2 of Table \ref{tab:wikihop}.}
\label{fig:doc2_hegcn}
\end{figure*}

The entity graph for the document is constructed from the above entity mention graph. Identical entity strings appear as only one node in the entity graph. So there is only one node for {\em Zoo Lake} in the entity graph. Since the strings of the nodes {\em Zoo Lake} and {\em Johannesburg} appear in the same sentence, an EG type 1 edge is added between them. All other EG type 1 edges are similarly added. The sequence of entities based on their appearance in the document is as follows (where only the first occurrence of multiple occurrences of the same entity is kept in the sequence): {\em Zoo Lake} $\rightarrow$ {\em Johannesburg} $\rightarrow$ {\em South Africa} $\rightarrow$ {\em Hermann Eckstein Park} $\rightarrow$ {\em Johannesburg Zoo} $\rightarrow$ {\em Parktown Spruit}. An EG Type 2 edge is added between two nodes $n_1$ and $n_2$ in the entity graph if $n_2$ follows $n_1$ in this sequence. We denote this graph as $\mathcal{G}_s = \{\mathcal{V}_s, \mathcal{E}_s\}$. Similarly, we construct the entity mention graph and entity graph for Doc2 of Table \ref{tab:wikihop} in Figure \ref{fig:doc2_hegcn}. We denote this graph as $\mathcal{G}_o = \{\mathcal{V}_o, \mathcal{E}_o\}$.

\begin{figure*}[t]
\centering
\includegraphics[scale=0.48]{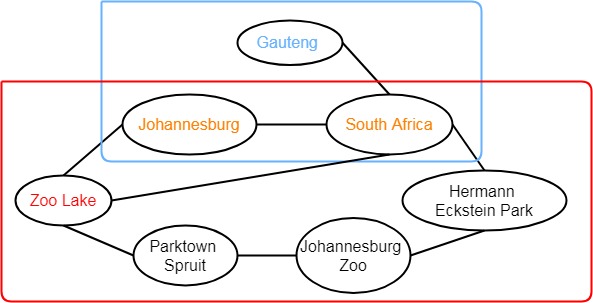}
\caption{A unified entity graph constructed from the entity graphs of Figure \ref{fig:doc1_hegcn} and Figure \ref{fig:doc2_hegcn}. Nodes in the red box are part of the entity graph of the document containing the subject entity \textcolor{red}{Zoo Lake}. Nodes in the blue box are part of the entity graph of the document containing the object entity \textcolor{blue}{Gauteng}. Common entities are marked in orange color.}
\label{fig:unified_hegcn}
\end{figure*}

Now, we need to unify the entity graphs of Figure \ref{fig:doc1_hegcn} and Figure \ref{fig:doc2_hegcn} to create a unified entity graph as shown in Figure \ref{fig:unified_hegcn}. We denote this unified entity graph as $\mathcal{G} = \{\mathcal{V}, \mathcal{E}\}$, where $\mathcal{V}=\mathcal{V}_s\cup\mathcal{V}_o$. The common entity nodes {\em Johannesburg} and {\em South Africa} appear only once in the unified graph $\mathcal{G}$. There is an edge between two nodes in $\mathcal{G}$ if there is an edge between them either in $\mathcal{G}_s$ or in $\mathcal{G}_o$.

\subsection{Relation Classifier}

We concatenate the EGCN outputs of the nodes corresponding to the subject entity $\textbf{e}_s \in \mathbb{R}^{6(d_w+d_z)}$ and object entity $\textbf{e}_o \in \mathbb{R}^{6(d_w+d_z)}$, and pass the concatenated vector to a feed-forward network (FFN) with softmax to predict the normalized probabilities for the relation labels.
\begin{align}
&\mathbf{r} = \mathrm{softmax}(\mathbf{W}_r (\mathbf{e}_s~||~\mathbf{e}_o) + \mathbf{b}_r)
\end{align}
$\mathbf{W}_r \in \mathbb{R}^{ (\vert R \vert+1) \times 12(d_w+d_z)}$ is the weight matrix, $\mathbf{b}_r \in \mathbb{R}^{\vert R \vert+1}$ is the bias vector of the FFN, and $\mathbf{r}$ is the vector of normalized probabilities of relation labels.

\subsection{Loss Function}
We calculate the loss over each mini-batch of size $B$. We use the following negative log-likelihood as our objective function for relation extraction:
\begin{equation}
\mathcal{L} = -\frac{1}{B} \sum_{i=1}^{B} \mathrm{log} (p(r_{i} \vert D_i, e_i^s, e_i^o, \theta))
\end{equation}
where $p(r_{i} \vert D_i, e_i^s, e_i^o, \theta)$ is the conditional probability of the true relation $r_i$ when the document chain $D_i$, the subject entity $e_i^s$, the object entity $e_i^o$, and the model parameters $\theta$ are given.

\section{Experiments}

\subsection{Evaluation Metrics}

We use precision, recall, and F1 score to evaluate the models on our multi-hop dataset. We do not include the {\em None} relation in the evaluation. A confidence threshold is used to decide if the relation of a test instance belongs to the set of relations $R$ or {\em None}. If the network predicts {\em None} for a test instance, then the test instance is classified as {\em None} only. But if the network predicts a relation from the set $R$ and the corresponding softmax score is below the confidence threshold, then the final class is changed to {\em None}. Using the confidence threshold achieves the highest F1 score on the validation dataset. 

\subsection{Parameter Settings}

We use GloVe \citep{pennington2014glove} word embeddings of dimension $d_w$ which is set to 300 in our experiments, and update the embeddings during training. We set the dimension $d_z$ to be 20 for the entity token indicator embedding vectors. The hidden vector dimension of the forward and backward LSTM is set at $320$. The dimension of BiLSTM output is $640$. We use $500$ different convolution filters with kernel width of $3$, $4$, and $5$ for feature extraction. We use one convolutional layer in both entity mention-level GCN and entity-level GCN in our final model. Dropout layers \citep{Srivastava2014DropoutAS} are used in our network with a dropout rate of $0.5$ to avoid overfitting. We train our models with a mini-batch size of $32$ and optimize the network parameters using the Adagrad optimizer \citep{duchi2011adaptive}.

\subsection{Baselines}

We implement four neural baseline models for comparison with our proposed HEGCN model. Similar to our proposed model, we represent the tokens in the documents using pre-trained word embedding vectors and entity token indicator vectors. We use a document separator token when concatenating the vectors of two documents in a chain. 

(1) CNN: We apply the convolution operation on the sequence of token vectors with different kernel sizes. A max-pooling operation is applied to choose the features from the outputs of the convolution operation. This feature vector is passed to a feed-forward layer with softmax to classify the relation.

(2) BiLSTM: The token vectors of the document chain are passed to a BiLSTM layer to encode its meaning. We obtain the entity mention vectors of the subject entity and the object entity by concatenating the hidden vectors of their first and last token. We average the entity mention tokens of the corresponding entity to obtain the representation of the subject entity and the object entity. These two vectors are concatenated and passed to a feed-forward layer with softmax to find the relation between them.

(3) BiLSTM\_CNN: This is a combination of the BiLSTM and CNN model described above. The token vectors of the documents are passed to a BiLSTM layer and then we use the convolution operation with max-pooling with different convolutional kernel sizes on the hidden vectors of the BiLSTM layer. The feature vector obtained from the max-pooling operation is passed to a feed-forward layer with softmax to classify the relation.

(4) LinkPath: This model uses the explicit paths from the subject entity $e_s$ to the object entity $e_o$ via the common entities to find the relation. As we consider only 2-hop relations, each path from $e_s$ to $e_o$ will be of the form $e_s \rightarrow c \rightarrow e_o$, where $c$ is a common entity. Since there can be multiple common entities between two documents and these common entities as well as the subject and object entities can appear multiple times in the two documents, there exist multiple paths from $e_s$ to $e_o$. Each path is formed with four entity mentions: (i) entity mentions of the subject entity and common entity in the first document. (ii) entity mentions of the common entity and object entity in the second document. We concatenate the BiLSTM hidden vectors of the start and end token of an entity mention to obtain its representation. Each path is constructed by concatenating all the four entity mentions of the path. This can be extended from 2-hop to multi-hop relations by using a recurrent neural network that takes the path entity mentions as input, and outputs the hidden representation of the path. We average the vector representations of all the paths and pass it to a feed-forward layer with softmax to find the relation.  

\subsection{Experimental Results}

We include in Table \ref{tab:results} the results of the models on the MHRED dataset. We see that adding a BiLSTM in the document encoding layer improves the performance by more than 6\% in F1 score. This improvement mostly comes from the higher precision score of the BiLSTM models over the CNN model, which classifies a higher number of {\em None} relations as positive relations, leading to a poorer precision score. The BiLSTM layer adequately captures the long term dependencies among the documents and helps to predict the {\em None} samples better. When we add our proposed hierarchical entity graph convolutional layer on top of the BiLSTM layer, we get another 2.2\% F1 score improvement over the next best BiLSTM\_CNN model. Our HEGCN model achieves the highest precision score with a competitive recall score compared to the BiLSTM-based baselines. A higher precision score for relation extraction is very important, as it reduces the number of erroneous tuples to build a cleaner knowledge base. We also perform a statistical significance test using bootstrap resampling to compare each baseline and our HEGCN model, and have ascertained that the higher F1 score achieved by our model is statistically significant ($p < 0.001$).

\begin{table}[ht]
\centering
\begin{tabular}{l|ccc}
\hline
Model       & Prec. & Rec.  & F1    \\ \hline
CNN         & 0.562 & 0.672 & 0.612 \\ 
BiLSTM      & 0.680 & 0.661 & 0.671 \\ 
LinkPath    & 0.665 & 0.684 & 0.674 \\ 
BiLSTM\_CNN & 0.651 & \textbf{0.701} & 0.675 \\ \hline
HEGCN       & \textbf{0.705} & 0.689 & \textbf{0.697} \\ \hline
\end{tabular}
\caption{Performance comparison of the models on the MHRED dataset.}
\label{tab:results}
\end{table}

\section{Analysis \& Discussion}

\subsection{Varying the Number of GCN Layers}

We include in Table \ref{tab:gcn_layers_ablation} the performance of our HEGCN model with different numbers of convolutional layers in the entity mention-level graph convolutional network (EMGCN) and unified entity-level graph convolutional network (EGCN). We see that when we increase the number of layers in either GCN, the performance of the model drops. We finally use only one convolutional layer in both EMGCN and EGCN.

\begin{table}[ht]
\centering
\begin{tabular}{cc|ccc}
\hline
L1 & L2 & Prec. & Rec.  & F1    \\ \hline
1  & 1  & \textbf{0.705} & \textbf{0.689} & \textbf{0.697} \\ 
2  & 1  & 0.656 & 0.666 & 0.661 \\ 
2  & 2  & 0.681 & 0.658 & 0.669 \\ 
3  & 1  & 0.664 & 0.664 & 0.664 \\ 
3  & 2  & 0.618 & 0.669 & 0.642 \\ 
3  & 3  & 0.697 & 0.635 & 0.665 \\ \hline
\end{tabular}
\caption{The ablation study of the HEGCN model with different numbers of convolutional layers (L1 and L2) in EMGCN and EGCN.}
\label{tab:gcn_layers_ablation}
\end{table}

\subsection{Effectiveness of Model Components}

In Table \ref{tab:component_ablation}, we include the ablation results of different components of our HEGCN model. F1 score drops by 1.3\% after removing the entity mention-level graph convolutional network (-- EMGCN). F1 score drops by 0.3\% after removing the unified entity-level graph convolutional network (-- EGCN). When we remove both GCNs (-- Both GCNs) together from the model, the F1 score drops by 2.8\%. Instead of using attention, if we obtain the context vector for each entity mention just by averaging the sentence token vectors, the F1 score drops by 2.6\% (-- Attention).

\begin{table}[ht]
\centering
\begin{tabular}{l|ccc}
\hline
Model  & Prec. & Rec.  & F1    \\ \hline
HEGCN  & \textbf{0.705} & 0.689 & \textbf{0.697} \\ 
\quad -- EMGCN  & 0.695 & 0.673 & 0.684 \\ 
\quad -- EGCN  & 0.682 & \textbf{0.706} & 0.694 \\ 
\quad -- Both GCNs & 0.667 & 0.672 & 0.669 \\
\quad -- Attention & 0.692 & 0.651 & 0.671 \\ \hline
\end{tabular}
\caption{The ablation study of the different components of our HEGCN model.}
\label{tab:component_ablation}
\end{table}

\subsection{Edge Ablation}

In Table \ref{tab:edge_ablation}, we include the ablation study of the different types of edges in the entity mention-level graph convolutional network (EMGCN) and unified entity-level graph convolutional network (EGCN). This study shows that the edges that are added to create a linear chain in the EMGCN (EMG type 3) and EGCN (EG type 2) are the most significant. Removing them separately from the network drops the F1 score by 3.6\% and 1.8\% respectively.

\begin{table}[ht]
\centering
\begin{tabular}{l|ccc}
\hline
Model  & Prec. & Rec.  & F1    \\ \hline
HEGCN  & 0.705 & 0.689 & \textbf{0.697} \\ 
\quad -- EMG type 1  & \textbf{0.717} & 0.654 & 0.684 \\ 
\quad -- EMG type 2  & 0.677 & \textbf{0.696} & 0.686 \\ 
\quad -- EMG type 3 & 0.655 & 0.668 & 0.661 \\ 
\quad -- EG type 1 & 0.694 & 0.685 & 0.690 \\
\quad -- EG type 2 & 0.701 & 0.658 & 0.679 \\ \hline
\end{tabular}
\caption{The ablation study of the different types of edges in our HEGCN model.}
\label{tab:edge_ablation}
\end{table}

\subsection{Error Analysis}

To analyze the prediction errors, we divide them into three categories: (i) positive relations are misclassified as {\em None} relations. 27.2\% of the errors occur in this category. (ii) {\em None} relations are misclassified as any of the positive relations. 5.3\% of the errors occur in this category. (iii) positive relations are misclassified as other positive relations. 5.4\% of the errors occur in this category. This result shows that the majority of misclassification errors come from the first category. We randomly sample negative instances from the entire set of negative instances, such that the number of negative instances equals the total number of positive instances in our training data. Since the positive instances belong to 218 relations, comparatively, there is a higher number of negative instances. This data imbalance may be the reason why our HEGCN model makes more errors in the first category. This imbalance of positive instances and {\em None} instances in distantly supervised data has always been a critical issue, and from Table \ref{tab:mhre_dataset_stat}, we see that our dataset also faces this problem. 

\section{Summary}

In this chapter, we describe how the idea of distant supervision can be extended from sentence-level extraction to multi-hop extraction to cover more relations from the KBs. We propose a general approach to create multi-hop relation extraction datasets. Following this approach, we create a 2-hop relation extraction dataset that covers a higher number of relations from knowledge bases than other distantly supervised relation extraction datasets. We also propose a hierarchical entity graph convolutional network for this task. The two levels of GCN in our model help to capture the relation cues within documents and across documents. Our proposed model improves the F1 score by 2.2\% on our 2-hop dataset, compared to a strong neural baseline, and it can be readily extended to N-hop datasets.

%
\chapter{Conclusion and Future Work}
\label{chapt:conclusion}

In this thesis, we first provide a brief overview of neural networks, knowledge bases, named entity recognition, open information extraction, distantly supervised relation extraction, and multi-hop natural language processing. We have also elucidated our contributions to the development of distantly supervised relation extraction. First, we describe a syntax-focused multi-factor attention model to find the relation between two entities in a sentence in a pipeline fashion. Our model helps to find the relation correctly when sentences are long and entities are located far from each other. 

Second, we address the problem of joint entity and relation extraction using an encoder-decoder approach. Joint extraction is more challenging due to the presence of multiple tuples in a sentence and the sharing of entities among the tuples. We first propose a representation scheme for relation tuples which enables the decoder to generate one word at a time like machine translation models and still finds all the tuples present in a sentence with full entity names of different lengths and with overlapping entities. Next, we propose a pointer network-based decoding approach where an entire tuple is generated at every time step. This joint extraction approach can eliminate the need for a separate named entity recognition system that is required in pipeline approaches. Also, this joint approach can better capture the interaction among multiple relation tuples present in a sentence to achieve improved performance.

Third, we focus on multi-hop relation extraction to extract more relations. We describe a general approach to create multi-hop relation extraction datasets. We adopt this approach to create a 2-hop dataset that includes more relations than other widely used distantly supervised datasets. We also describe a hierarchical entity graph convolutional network for multi-hop relation extraction which improves the performance on our 2-hop dataset compared to some strong neural baselines. This model can be readily extended to N-hop datasets in the future.

With the progress of deep learning algorithms, significant advances have been made in the relation extraction task. However, many challenges remain in this area. In the pipeline approaches, since we need to find relations among all pairs of entities, there can be a very large number of {\em None} instances. This {\em None} class is challenging to identify as it is not a single relation but any relation outside the set of positive relations. Erroneous detection of {\em None} relation reduces the precision of the model and can add many wrong tuples to the KB. To build a cleaner KB, models have to perform very well to detect the {\em None} relation along with classifying the positive relations correctly. Our error analysis in Chapter 4 with varying sentence length and varying distance between the entity pairs shows that the performance of the neural models drops significantly with increasing sentence length and increasing distance between the entities. Future research should focus on this aspect of relation extraction.

Regarding the joint extraction approach, our work and other prior works do not include sentences with zero tuples in training or testing. But many sentences do not contain any relation tuples. So in the future, detecting sentences with no relation tuples must be handled in the joint extraction approaches. The datasets we have used for our experiments in Chapters 4 \& 5 contain a much higher number of sentences with zero tuples than the sentences with valid tuples. So the inclusion of sentences with zero tuples in the experiments makes data imbalance in joint entity and relation extraction more challenging.

Current relation extraction models deal with very few relations whereas existing knowledge bases have thousands of relations. In the future, we need to expand multi-hop relation extraction to cover more relations. We have shown with our 2-hop relation extraction dataset that it covers more relations than any other sentence-level or document-level datasets. We need to extend to N-hop relation extraction to cover more relations from the KB. However, it may not be easy to extend the task as the inclusion of more documents in the chain may make the data more noisy. It will be challenging to create a clean dataset for N-hop relation extraction. Also, we need to explore zero-shot or few-shot relation extraction to cover the relations for which we cannot obtain enough training data using distant supervision. 

%
\singlespacing
\bibliographystyle{plainnat}
\bibliography{thesis}

\end{document}